\RequirePackage{fix-cm}
\documentclass{article}
\usepackage{iclr2027_conference,times}
\iclrfinalcopy
\usepackage{amsmath,amssymb,mathtools,booktabs,graphicx,microtype,xcolor}
\usepackage{needspace,placeins,float,array,multirow,colortbl,longtable}
\usepackage{capt-of}
\usepackage{hyperref,url}
\definecolor{grpocol}{HTML}{C3E5FC}
\definecolor{reinforcecol}{HTML}{E8F6FE}
\definecolor{tableheadcol}{HTML}{FFFFFF}
\definecolor{ralink}{HTML}{31577F}
\definecolor{basecol}{HTML}{EDEDED}

\hypersetup{colorlinks=true,linkcolor=black,citecolor=black,urlcolor=ralink,
pdftitle={Residual Advantage: Student-Relative Teacher Guidance for RL with Verifiable Rewards}}
\newcommand{\RA}{\textsc{RA}}
\newcommand{\CoRA}{Co-\textsc{RA}}
\newcommand{\E}{\mathbb E}
\newcommand{\Var}{\operatorname{Var}}
\newcommand{\Cov}{\operatorname{Cov}}
\newcommand{\KL}{\operatorname{KL}}
\newcommand{\Diag}{\operatorname{Diag}}
\newcommand{\norm}[1]{\lVert#1\rVert}
\newcommand{\sd}[1]{{\scriptsize$\pm$#1}}
\newcommand{\modelindent}{\hspace{0.65em}}

\newcolumntype{R}[1]{>{\raggedleft\arraybackslash}p{#1}}
\newcolumntype{C}[1]{>{\centering\arraybackslash}p{#1}}
\newcommand{\modelgrouprule}{\arrayrulecolor{black!28}\specialrule{0.3pt}{1pt}{1pt}\arrayrulecolor{black}}
\makeatletter
\newcommand{\RA@rowshade}[2]{%
  \begingroup
  \setbox\z@=\hbox{#2}%
  \@tempdima=\ht\@arstrutbox
  \ifdim\ht\z@>\@tempdima \@tempdima=\ht\z@\fi
  \@tempdimb=\dp\@arstrutbox
  \ifdim\dp\z@>\@tempdimb \@tempdimb=\dp\z@\fi
  \advance\@tempdima by \@tempdimb
  \color{#1}\hrule height\@tempdima
  \vskip-\@tempdima
  \endgroup}
\newcommand{\tablerowshade}[1]{\noalign{\RA@rowshade{#1}{}}}
\newcommand{\tablerowshademeasured}[2]{\noalign{\RA@rowshade{#1}{#2}}}
\newsavebox\RAtabbox
\newlength\RAtabwd
\newcommand{\RAshadeindent}{0.65em}
\newcommand{\RA@rowshadeindent}[1]{%
  \begingroup
  \@tempdima=\ht\@arstrutbox
  \advance\@tempdima by \dp\@arstrutbox
  \ifdim\RAtabwd>\z@
    \moveright\RAshadeindent\hbox{\color{#1}\vrule width\dimexpr\RAtabwd-\RAshadeindent\relax height\@tempdima depth\z@}%
    \vskip-\@tempdima
  \fi
  \endgroup}
\newcommand{\tablerowshadeindent}[1]{\noalign{\RA@rowshadeindent{#1}}}
\newcommand{\measuredtable}[1]{%
  \sbox\RAtabbox{\input{#1}}%
  \global\setlength\RAtabwd{\wd\RAtabbox}%
  \input{#1}%
  \global\setlength\RAtabwd{\z@}}
\makeatother
\makeatletter
\newcommand{\tabletwoheadshade}[1]{\noalign{%
  \begingroup
  \@tempdima=\ht\@arstrutbox
  \advance\@tempdima by \dp\@arstrutbox
  \multiply\@tempdima by 2
  \advance\@tempdima by \aboverulesep
  \advance\@tempdima by \cmidrulewidth
  \advance\@tempdima by \belowrulesep
  \color{#1}\hrule height\@tempdima
  \vskip-\@tempdima
  \endgroup}}
\newcommand{\tabletwoheadlabel}[1]{%
  \multirow{2}{*}[-\dimexpr(\aboverulesep+\cmidrulewidth+\belowrulesep)/2\relax]{#1}}
\makeatother
\newtheorem{proposition}{Proposition}

\title{Residual Advantage:\\Student-Relative Teacher Guidance\\for RL with Verifiable Rewards}
\author{Xiaobing Chen$^{1,2,3}$ \quad Zhiqi Pang$^{1}$\thanks{Corresponding author.}\\
$^{1}$Harbin Engineering University \quad $^{2}$Tencent \quad $^{3}$Harbin Institute of Technology\\
\texttt{xbchen@stu.hit.edu.cn}}
\begin{document}
\maketitle
\lhead{Preprint}
\begin{abstract}
Reinforcement learning with verifiable rewards (RLVR) and on-policy distillation (OPD) have become two main paradigms for post-training reasoning models. RLVR gives each response a single outcome label, leaving the steps inside it without separate credit. OPD provides token-level guidance at student-visited prefixes, but its pointwise signal does not directly reflect the pattern of teacher--student disagreement across the vocabulary. Dense, unbounded log-ratio supervision can amplify the teacher's influence, yet a strong solver is not necessarily a suitable guide when the student's solution paths depart from the teacher's. We propose Residual Advantage (\RA{}), which treats the teacher--student probability residual as a bounded one-step reward, subtracts the corresponding state value under the student policy to form a standard advantage, and centers the result within each response before adding it to the verifier advantage. The guidance term has zero mean within each response, so the verifier advantage remains the response's mean label and the teacher only redistributes credit among the steps within it. \CoRA{} further updates a teacher LoRA with verifier advantages on the same scored student batch and uses the updated teacher in the next iteration's residual, adapting guidance to the student's attempts. With Qwen3-1.7B-Base and Qwen3-4B-Base students and a Qwen3-8B teacher, \RA{} combined with GRPO or REINFORCE++ improves the underlying sequence-advantage algorithm in all 24 comparisons on three mathematical benchmarks, raising macro Avg@8 by 1.7--3.6 points and Pass@8 by 3.9--6.3 points. Both combinations surpass teacher-only OPD, and \CoRA{} adds a further 1.0--1.5 Avg@8 points.
\end{abstract}

\section{Introduction}

Reinforcement learning with verifiable rewards (RLVR) trains a language model on its own sampled solutions, scored by a checker of the final answer \citep{deepseekmath,deepseekr1}. GRPO and REINFORCE++ turn this terminal reward into one advantage per response \citep{deepseekmath,hu2025reinforcepp}, so every step receives the same label even when the response mixes useful and mistaken steps. Learning is also confined to the successful paths the student can sample, and under practical budgets much of the gain reflects sharpening of what the model can already do \citep{yue2025rlvr,huang2024sharpening}.

On-policy distillation (OPD) instead supplies a teacher's next-token distribution at every prefix the student visits \citep{gkd,lu2025opd}. Its label, the log ratio $\log(q(a)/p(a))$ or one coordinate of a top-$k$ or full-vocabulary divergence, says how far apart teacher and student are at that token but does not directly reflect where the teacher's extra mass sits; a full-vocabulary objective sees this only implicitly, through the softmax coupling of the coordinates. The log-ratio label is moreover unbounded, growing as $\log(1/q(a))$ at a token the teacher assigns little probability, so on such tokens the teacher's signal can become disproportionately strong. At the same time, a strong solver is not necessarily a good teacher for the current student, least of all on prefixes where the student has left the teacher's solution paths \citep{li2025smallmodels,tiapkin2025teacherhacking}. The question is how teacher guidance should enter RLVR so that the student draws on the teacher's knowledge while the response-level label stays tied to outcome feedback and room for exploration remains.

\begin{figure}[t]
\centering
\includegraphics[width=0.78\linewidth]{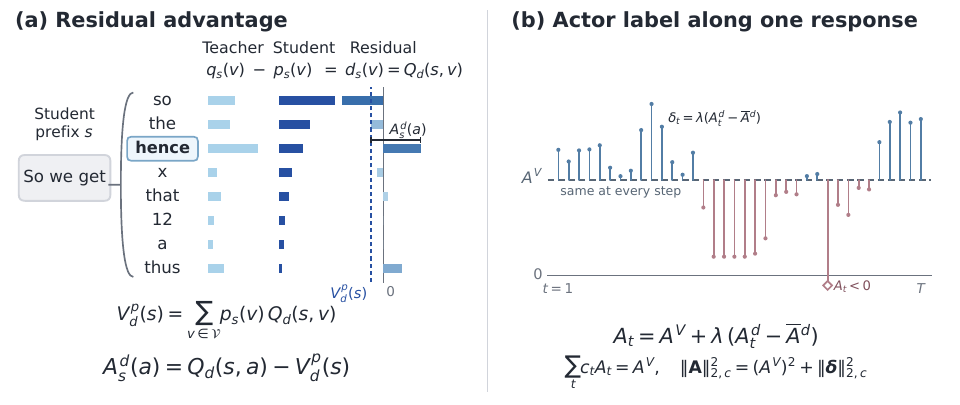}
\caption{\textbf{Local teacher guidance redistributes the verifier's response-level credit.}
\textbf{(a)}~At one prefix, the reward $Q_d(s,v)=q_s(v)-p_s(v)$ over the student's likeliest tokens, its value $V_d^p(s)$ under the student policy, and the advantage $A_s^d(a)$ of the sampled token, here the student's third choice and the teacher's first. \textbf{(b)}~Labels along one response, with blue and coral stems for positive and negative guidance terms $\delta_t$; each label is $A^V$ plus a centered guidance term, so the labels' weighted mean is $A^V$.}
\label{fig:teaser}
\end{figure}

To this end, we propose Residual Advantage (\RA{}). \RA{} treats the probability difference $d=q-p$ as a bounded one-step reward. At each visited prefix it subtracts the reward's expectation under the student's own policy, that is, the policy's value under this reward,
\[
 A^{d}(a)=d(a)-\langle p,d\rangle=\E_{v\sim p}\big[(q(a)-q(v))-(p(a)-p(v))\big],
\]
This design reflects how the teacher's disagreement with the student is distributed. It asks whether, relative to the student, the teacher favors the sampled token over the other candidates the student would consider, each weighted by the student's probability of choosing it. Even with the same residual at the sampled token, teachers that distribute the rest of the vocabulary differently can therefore give local advantages of opposite signs (Appendix~\ref{app:units}). \RA{} then centers the local advantages within the response and adds them to the verifier advantage, $A_t=A^V+\lambda(A_t^d-\overline A^d)$, where $\overline A^d$ is the response mean and $\lambda$ the guidance weight. This gives the guidance term zero mean within every response, so the verifier advantage remains the response's mean label and the teacher only redistributes credit among the steps within it. The guidance is thus student-relative twice over, as an advantage under the student's own next-token distribution and as a deviation within the student's own response.

The appeal of the construction is that it is both elegant and natural. The local advantage is a single inner product, $A^d(a)=\langle e_a-p,\,q-p\rangle$, and three properties follow from it. It is the rate at which the squared distance $\tfrac12\norm{p-q}_2^2$ falls when the student moves probability toward $a$. Its expected update at a prefix is the exact logit gradient of that distance, so the student's own probability of a token scales how strongly the teacher can move it. And it is bounded by the student--teacher distance times the probability the student leaves to other tokens, so guidance fades on its own as the two converge. Scaled in this way by the student's own probability, it also stays small wherever the student is already sure, so the teacher mainly adjusts the choices the student is still weighing instead of imposing its preference on those the student has settled, and it does not inflate small disagreements into large labels. The construction requires no schedule, threshold, or learned component.

\CoRA{} extends the construction to a teacher that learns from the student's attempts. Each iteration scores a student batch, fixes its labels, and updates a teacher LoRA with the base estimator's verifier advantages while the student receives the guided label, much as actor and critic share a rollout batch in PPO \citep{schulman2017ppo}. The updated teacher supplies the distribution in the next iteration's residual, with the same advantage construction and weight as \RA{}. Training on the student's own rollouts with the verifier's advantages lets the teacher adapt to the reasoning along the student's solution paths, with the aim of reducing the bias a frozen teacher carries on trajectories it would not have generated itself. Across two students, two estimators, and three mathematical benchmarks, \RA{} improves all 24 three-seed benchmark means of its base estimator, both combinations surpass teacher-only OPD, and \CoRA{} raises accuracy further with mixed changes in coverage.

\section{Residual Advantage}
\label{sec:method}

\paragraph{Setting.}
A student $p_\theta$ generates a response $a_{i,1:L_i}$ to prompt $x_i$. At the visited prefix $\xi_{it}=(x_i,a_{i,<t})$, let $p_{it}$ be the rollout student's next-token distribution and $q_{it}$ the teacher's distribution at the same prefix over the same aligned vocabulary. We write $\langle p,f\rangle=\sum_v p(v)f(v)$, let $e_a$ be the one-hot vector of token $a$, and drop the index $i$ when a single response is meant. The base estimator converts a terminal verifier reward into one advantage $A_i^V$ per response, by GRPO's within-group normalization \citep{deepseekmath} or by REINFORCE++'s batch statistics \citep{hu2025reinforcepp}, and the actor maximizes the clipped PPO surrogate \citep{schulman2017ppo} with importance ratio $\pi_\theta(a\mid\xi)/p(a\mid\xi)$, the actor's own ratio, distinct from the teacher--student ratio $q/p$. \RA{} changes only the per-token label that enters this surrogate. Every teacher-derived quantity is computed at the rollout policy and held fixed, so no gradient flows through it.

\subsection{Student-relative residual advantage}
\label{sec:coefficient}

At prefix $s$, we define the one-step reward of token $v$ as the teacher--student probability \emph{residual},
\begin{equation}
 R_s(v)=d_s(v)=q(v\mid s)-p(v\mid s).
 \label{eq:residual}
\end{equation}
Under the student policy, the state value of this reward is $V_d^p(s)=\E_{v\sim p}R_s(v)=\langle p_s,d_s\rangle$, its action value is the immediate reward $Q_d(s,a)=d_s(a)$, and the token advantage is
\begin{equation}
 \boxed{\,A_s^{d}(a)=Q_d(s,a)-V_d^p(s)=d_s(a)-\langle p_s,d_s\rangle=\langle e_a-p_s,\,d_s\rangle .\,}
 \label{eq:score}
\end{equation}
The reward is immediate, so nothing is propagated to earlier positions; credit across positions comes from the verifier advantage. Subtracting $V_d^p$ is the REINFORCE baseline for this reward, computed exactly \citep{williams1992simple,sutton2000pg,foerster2018coma}, and we call it the \emph{student value baseline}. The gap $d_s(a)$ at the sampled token alone does not determine the sign (Appendix~\ref{app:units}).

\begin{proposition}[Three properties of the local advantage]
\label{prop:advantage}
At a fixed prefix, whose index we drop, let $L(p,q)=\tfrac12\norm{p-q}_2^2$, let $p=\operatorname{softmax}(z)$, and let $\psi_a=e_a-p=\nabla_z\log p(a)$ be the score of token $a$ in logit coordinates.
\textup{(i)} $A^{d}(a)=\langle\psi_a,d\rangle$ is the first-order decrease of $L$ along the feasible move $p\mapsto(1-\eta)p+\eta e_a$; the exact change is $-\eta A^{d}(a)+\tfrac{\eta^2}{2}\norm{\psi_a}_2^2$.
\textup{(ii)} $\E_{a\sim p}\psi_a=0$ and $\E_{a\sim p}\psi_a\psi_a^\top=J_p=\Diag(p)-pp^\top$, the Jacobian of the softmax. Hence, with $d$ held fixed, $\E_{a\sim p}A^{d}(a)=0$ and
\begin{equation}
 \E_{a\sim p}\big[A^{d}(a)\,\psi_a\big]=J_pd=-\nabla_z L(p,q),\qquad -\frac{\partial L}{\partial z_b}=p(b)\,A^{d}(b).
 \label{eq:expected-update}
\end{equation}
\textup{(iii)} $|A^{d}(a)|\le\norm{\psi_a}_2\norm{d}_2\le\sqrt2\,(1-p(a))\norm{d}_2$, and $-\tfrac12\le A^{d}(a)\le2$.
\end{proposition}

Part (i) says what the local advantage measures. Part (ii) says what it does on average. The update descends the squared distance to the teacher, and logit $b$ moves by $p(b)A^d(b)$, so the student's own probability of a token scales how far the teacher can move it. Since $\E_{a\sim p}\psi_a=0$, the student value baseline is applied automatically in expectation at a single prefix; its role appears only when local advantages are compared across prefixes (Appendix~\ref{app:response-composition}). Part (iii) is where the residual parts from the log ratio, which estimates reverse-KL descent without bias but has no bound. The residual's bound gives the guidance bounds of Section~\ref{sec:centering}.

Separating the sampled token makes this scaling explicit,
\begin{equation}
 A^{d}(a)=\big(1-p(a)\big)\Big[d(a)-\E_{v\sim p(\cdot\mid v\ne a)}d(v)\Big],
 \label{eq:factor}
\end{equation}
the teacher's preference for $a$ over the student's other choices, weighted by the probability the student still gives them. The local advantage vanishes on a token the student has settled, whatever the teacher says, and by part (ii) logit $b$ moves in expectation by $p(b)(1-p(b))$ times a bounded contrast, so the teacher acts most where the student is undecided. None of this is imposed by hand. The relative comparison, the confidence scaling, the zero mean, and the bound are all read off one inner product.

\subsection{Combining with verifier feedback}
\label{sec:centering}

Let $c_t\ge0$ be the actor's relative weights on the response's positions, summing to one ($c_t=1/M$ for token-mean reduction over $M$ positions), and let $\overline{A}^{d}=\sum_tc_tA_t^{d}$ be the response mean of the local advantages. \RA{} adds the centered local advantage, scaled by the guidance weight $\lambda$, to the verifier advantage,
\begin{equation}
 \boxed{\,\delta_t=\lambda\,(A_t^{d}-\overline{A}^{d}),\qquad A_t=A^V+\delta_t,\qquad\lambda\ge0 .\,}
 \label{eq:correction}
\end{equation}
For every response,
\begin{equation}
 \sum_t c_tA_t=A^V,\qquad \delta_t-\delta_u=\lambda\,(A_t^{d}-A_u^{d}),
 \label{eq:mean}
\end{equation}
so the verifier advantage is exactly the response's mean label and every difference between two decisions is preserved. Under the weighted inner product $\langle u,v\rangle_c=\sum_tc_tu_tv_t$, the constant verifier label and the centered guidance term are orthogonal, and the label splits as
\begin{equation}
 \norm{\mathbf A}_{2,c}^2=(A^V)^2+\lambda^2\Var_{t\sim c}(A_t^d).
 \label{eq:orthogonal}
\end{equation}
The verifier decides how good the response is as a whole, and the teacher decides how that credit is shared among its steps. This division concerns the label. Multiplied by different score vectors, a zero-mean guidance term need not give a gradient orthogonal to the verifier's, and on a single token it can outweigh it (Appendices~\ref{app:gate} and~\ref{app:response-composition}). Centering is the orthogonal projection onto zero-mean labels. Without it, a nonzero response mean in the teacher label would be indistinguishable from a change of $A^V$, since both multiply the same mean score direction $\sum_tc_t\psi_t$.

\paragraph{Bounded by the distance, fading on its own.}
By Proposition~\ref{prop:advantage}(iii),
\begin{equation}
 |\delta_t|\le\tfrac52\lambda(1-c_t),\qquad
 \sum_t c_t|\delta_t|\le\norm{\delta}_{2,c}
 \le\sqrt2\lambda\Big(\sum_t c_t\norm{p_t-q_t}_2^2\Big)^{1/2}.
 \label{eq:bound}
\end{equation}
The first bound limits each guidance term. The second bounds the guidance amplitude by the student--teacher distance on the visited prefixes (Appendix~\ref{app:mass-bounds}), so teacher guidance fades as the two converge, at a fixed weight and without a schedule; Appendix~\ref{app:training-diagnostics} shows the mean absolute guidance declining over a training run. The log ratio also vanishes at exact agreement but admits no such bound, since it diverges at a token of vanishing student probability while $\norm{p-q}_2\to0$.

\begin{figure}[t]
\centering
\includegraphics[width=0.78\linewidth]{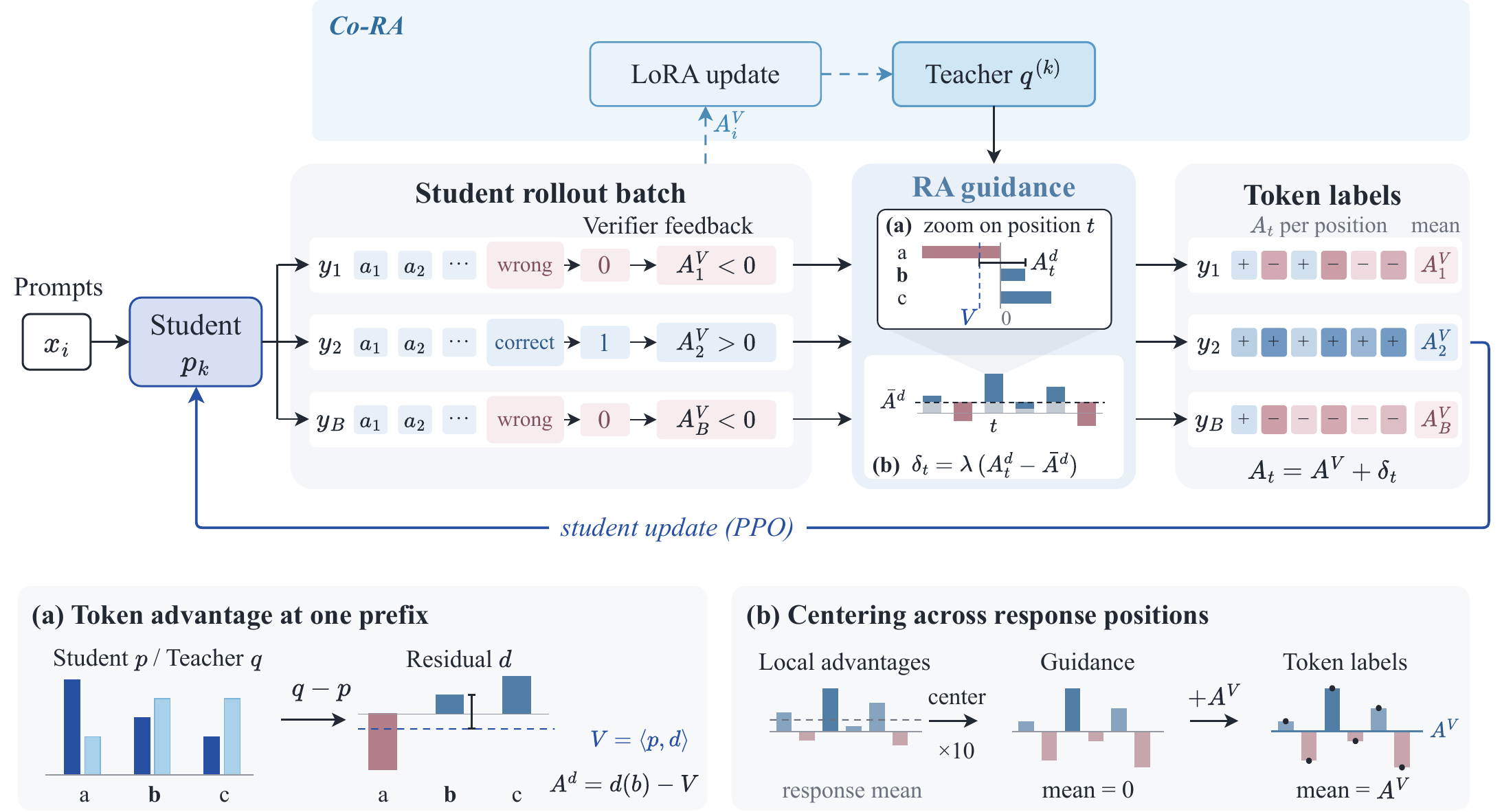}
\caption{\textbf{One iteration of \RA{} and \CoRA{}.}
Three schematic responses, only the middle one correct, receive verifier advantages $A_i^V$. \RA{} adds a centered guidance term to every token label. Token strips show one cell per position, blue for $A_t>0$ and rose for $A_t<0$, darker for larger $|A_t|$; the last cell is the response mean, which centering pins to $A_i^V$. Insets show (a) the student value baseline at one prefix and (b) response centering. In \CoRA{}, a LoRA adapter on the frozen teacher is trained with the verifier advantages of the same scored batch and supplies the next iteration's residual.}
\label{fig:method}
\end{figure}

\subsection{\CoRA{} teaches the teacher on the student's attempts}
\label{sec:cora}

\CoRA{} follows a score-then-update cycle. In every iteration the student generates a batch, the verifier scores it, and the current student and teacher produce the actor labels. These labels are fixed before the student and a LoRA adapter $q_\phi$ \citep{hu2022lora} on the frozen teacher $q_0$ are updated on the same batch. Both updates use the run's own GRPO or REINFORCE++ advantages, the adapter receiving $A_i^V$ and the student the combined label, so the teacher is adapted at the prefixes this student visits according to how its continuations turned out. The aim is for the teacher to become familiar with the reasoning along the student's solution paths and to take in the verifier's signal there, reducing its bias on the student's trajectories so that its guidance becomes more accurate for this student. With batch token-mean reduction (Appendix~\ref{app:adapter}), the adapter's gradient at the start of each update is
\begin{equation}
 \left.\nabla_\phi\mathcal J_T\right|_{\phi_k}
 =\frac1Z\sum_i A_i^V\sum_{t\in\mathcal T_i}\left.\nabla_\phi\log q_{\phi,it}(a_{it})\right|_{\phi_k},
 \qquad Z=\sum_i|\mathcal T_i| ,
 \label{eq:adapter-gradient}
\end{equation}
where $\mathcal T_i$ is the set of token positions of response $i$. This is the base estimator's clipped update applied to the teacher LoRA on student-generated data.

At iteration $k$ the adapted teacher $q_{\phi_k}$ supplies the residual $d_t^{(k)}=q_{\phi_k,t}-p_{k,t}$ against the rollout student $p_k$, and the student uses the same construction and weight $\lambda=10$ as \RA{},
\begin{equation}
 A_t^{\text{\CoRA}}=A^V+\lambda\bigl(A_t^{d,(k)}-\overline A^{d,(k)}\bigr),\qquad
 A_t^{d,(k)}=d_t^{(k)}(a_t)-\langle p_{k,t},d_t^{(k)}\rangle,
 \label{eq:cora}
\end{equation}
where $\overline A^{d,(k)}=\sum_tc_tA_t^{d,(k)}$. After the updates, $q_{\phi_{k+1}}$ supplies the residual in the next iteration. The adapter starts at zero, so the first student update coincides with \RA{}. At each scored batch the response-mean identity and the distance bound hold with the current teacher $q_{\phi_k}$ (Appendix~\ref{app:adaptive-teacher}).

\section{Experiments}
\label{sec:experiments}

\begin{table}[t]
\caption{\textbf{Residual guidance improves both RLVR estimators on every benchmark and metric, and the adapted teacher raises accuracy further.} Three-seed means in \%. Student rows are untrained, rows marked + add \RA{} or \CoRA{} to the estimator above, bold marks the best per column and student, and the last row is the Qwen3-8B teacher; per-seed scores and SD are in Appendix~\ref{app:seeds}.}
\label{tab:main}
\centering
\begingroup
\small
\setlength{\tabcolsep}{3.6pt}
\renewcommand{\arraystretch}{1.0}
\begin{tabular}{@{}l@{\hspace{6pt}}cc@{\hspace{7pt}}cc@{\hspace{7pt}}cc@{\hspace{7pt}}cc@{}}
\toprule
\tabletwoheadshade{tableheadcol}
\tabletwoheadlabel{\textbf{Method}} & \multicolumn{2}{c}{MATH-500} & \multicolumn{2}{c}{AIME24} & \multicolumn{2}{c}{AIME25} & \multicolumn{2}{c}{\textbf{Macro}}\\
\cmidrule(lr){2-3}\cmidrule(lr){4-5}\cmidrule(lr){6-7}\cmidrule(l){8-9}
 & Avg@8 & Pass@8 & Avg@8 & Pass@8 & Avg@8 & Pass@8 & Avg@8 & Pass@8\\
\midrule
\tablerowshade{basecol}\multicolumn{1}{@{}l}{\textit{Qwen3-1.7B-Base}} & 36.83 & 76.67 & 2.08 & 10.00 & 1.53 & 8.89 & 13.48 & 31.85\\
\modelgrouprule
\modelindent OPD (teacher only) & 67.78 & 85.00 & 8.61 & 23.33 & 5.97 & 15.56 & 27.45 & 41.30\\
\modelindent REINFORCE++ & 65.76 & 83.87 & 7.78 & 20.00 & 3.47 & 14.44 & 25.67 & 39.44\\
\tablerowshadeindent{reinforcecol}\modelindent \quad + \RA{} & 69.69 & 85.60 & 13.06 & 30.00 & 5.00 & 20.00 & 29.25 & 45.20\\
\tablerowshadeindent{reinforcecol}\modelindent \quad + \CoRA{} & 71.13 & 85.80 & \textbf{14.17} & \textbf{33.33} & 6.81 & 20.00 & 30.70 & \textbf{46.38}\\
\modelindent GRPO & 68.85 & 84.53 & 9.03 & 20.00 & 4.58 & 16.67 & 27.49 & 40.40\\
\tablerowshadeindent{grpocol}\modelindent \quad + \RA{} & 70.96 & \textbf{86.47} & 11.81 & 24.44 & 6.81 & 23.33 & 29.86 & 44.75\\
\tablerowshadeindent{grpocol}\modelindent \quad + \CoRA{} & \textbf{72.15} & 86.20 & 13.19 & 20.00 & \textbf{7.92} & \textbf{26.67} & \textbf{31.09} & 44.29\\
\addlinespace[3pt]
\midrule
\tablerowshade{basecol}\multicolumn{1}{@{}l}{\textit{Qwen3-4B-Base}} & 51.79 & 89.60 & 7.22 & 22.22 & 4.72 & 20.00 & 21.25 & 43.94\\
\modelgrouprule
\modelindent OPD (teacher only) & 78.83 & 90.87 & 13.19 & 30.00 & 11.81 & 26.67 & 34.61 & 49.18\\
\modelindent REINFORCE++ & 77.42 & 90.07 & 13.19 & 26.67 & 10.97 & 23.33 & 33.86 & 46.69\\
\tablerowshadeindent{reinforcecol}\modelindent \quad + \RA{} & 79.83 & \textbf{92.40} & 15.97 & 33.33 & 15.56 & \textbf{33.33} & 37.12 & \textbf{53.02}\\
\tablerowshadeindent{reinforcecol}\modelindent \quad + \CoRA{} & 81.15 & 92.33 & 17.36 & 30.00 & \textbf{16.67} & 30.00 & 38.39 & 50.78\\
\modelindent GRPO & 80.42 & 90.73 & 14.72 & 26.67 & 12.36 & 26.67 & 35.83 & 48.02\\
\tablerowshadeindent{grpocol}\modelindent \quad + \RA{} & 81.33 & \textbf{92.40} & 17.50 & 33.33 & 13.89 & 30.00 & 37.57 & 51.91\\
\tablerowshadeindent{grpocol}\modelindent \quad + \CoRA{} & \textbf{82.61} & 92.33 & \textbf{18.89} & \textbf{36.67} & 14.31 & 30.00 & \textbf{38.60} & 53.00\\
\addlinespace[3pt]
\midrule
\multicolumn{1}{@{}l}{\textit{Qwen3-8B teacher}} & 84.23 & 95.53 & 26.25 & 43.33 & 21.39 & 38.89 & 43.95 & 59.25\\
\bottomrule
\end{tabular}
\endgroup

\end{table}

\paragraph{Setup.}
Students are Qwen3-1.7B-Base and Qwen3-4B-Base; the teacher is the post-trained Qwen3-8B model in non-thinking mode, which shares their vocabulary \citep{qwen3}. Training uses DAPO-Math-17K \citep{dapo} for two epochs with 128 prompts per rollout batch, eight responses per prompt for GRPO runs and one for OPD and REINFORCE++ runs, temperature $1.0$, and a 12,000-token cap. Verifier runs reward answer correctness, linearly discounted for correct responses beyond 8,000 tokens (Appendix~\ref{app:reward}), and \RA{} is added to the base estimator at the reference weight $\lambda=10$, chosen on a validation split of training prompts (Appendix~\ref{app:validation}). Teacher-only OPD minimizes the reverse KL divergence $\KL(p_t\Vert q_t)$ over the full vocabulary at every visited prefix \citep{gkd,lu2025opd}, the same teacher information \RA{} uses. Evaluation samples eight completions per problem at temperature $0.8$ with the same cap on MATH-500 \citep{lightman2023lets}, AIME24, and AIME25 and reports Avg@8 and Pass@8 (Table~\ref{tab:main}) from unshaped answer correctness; macro scores weight the three benchmarks equally. Each \RA{} run adds one teacher forward pass per rollout, and \CoRA{} also trains the adapter (Appendix~\ref{app:adapter}).

\paragraph{\RA{} improves every matched comparison.}
Table~\ref{tab:main} contains 24 comparisons of \RA{} with its base estimator (two students, two estimators, three benchmarks, two metrics), and \RA{} improves all 24 three-seed means. The macro gains are $1.74$--$3.58$ Avg@8 and $3.89$--$6.33$ Pass@8 points (Table~\ref{tab:gains}).

\paragraph{The guidance transfers across estimators.}
The gains are larger for REINFORCE++, narrowing its macro Avg@8 deficit to GRPO at both sizes. GRPO keeps the higher completion accuracy after guidance, but REINFORCE++ with \RA{} has the higher macro Pass@8 at both sizes, a trade between accuracy and coverage that a single score would hide.

\paragraph{Verifier plus teacher improves on teacher-only distillation.}
Both \RA{} combinations exceed teacher-only OPD in macro Avg@8 and Pass@8 at both sizes, by $1.80$--$2.96$ and $2.73$--$3.90$ points (Table~\ref{tab:gains}). The one exception is AIME25 Avg@8 at 1.7B, where OPD's $5.97$ exceeds the $5.00$ of \RA{}+REINFORCE++, although both \RA{} runs have the higher Pass@8 there. In training, OPD also starts more slowly than REINFORCE++ and stays below \RA{}+REINFORCE++ over most of the shared budget (Appendix~\ref{app:dynamics}). At the same budget, OPD followed by REINFORCE++ reaches $28.39$ and REINFORCE++ with an added reverse-KL loss $26.14$, both below \RA{}+REINFORCE++ (Appendix~\ref{app:sequential}), and sampling the log ratio at the generated token, with the same expected update at each prefix, drops OPD to $24.80$ at 1.7B, below verifier-only training (Appendix~\ref{app:sampled-opd}).

\paragraph{Teaching the teacher on student rollouts raises accuracy in every tested setting.}
\CoRA{} adds $1.03$--$1.45$ macro Avg@8 points to the four \RA{} runs, and Avg@8 rises in all twelve benchmark means. Coverage changes are mixed: Pass@8 rises in four benchmark means, is unchanged in two, and falls in six (Table~\ref{tab:gains}). \CoRA{}+GRPO is the most accurate run for both students. Yet each adapted teacher scores $0.8$--$2.3$ macro Avg@8 and $0.7$--$3.5$ Pass@8 points below the frozen teacher (Appendix~\ref{app:adapted-teacher}), while the students it guides improve: it is a better guide, not a better solver.

\paragraph{The margin over the untrained student persists up to Pass@512.}
From $K=8$ to $K=512$ for the 1.7B REINFORCE++ runs (Appendix~\ref{app:passk}), the macro margin of \RA{} over the untrained student grows from $13.4$ to $16.9$ points, and its margins over REINFORCE++ and OPD from $5.8$ to $10.4$ and from $3.9$ to $6.8$, while that of verifier-only training narrows from $7.6$ to $6.6$, as \citet{yue2025rlvr} report for RLVR. Within the tested budget \RA{} thus raises the coverage the student can reach.

\section{Analysis}
\label{sec:analysis}

\subsection{Ablation Study}
\label{sec:baselines}

We remove the student value baseline and response centering in turn on Qwen3-1.7B-Base with REINFORCE++ at the unchanged weight $\lambda=10$ and compare every control with full \RA{}. Since removing either operation also changes the guidance amplitude, every control is also run at weights $5$ and $20$ (Table~\ref{tab:weight-sweep}, Appendix~\ref{app:lambda-seeds}), so that no conclusion rests on a single amplitude.

\paragraph{Student value baseline.}
Removing the student value baseline lowers both macro scores, by $2.09$ Avg@8 and $3.58$ Pass@8 points at the reference weight and similarly at weights $5$ and $20$ (Table~\ref{tab:weight-sweep}). The label then carries the sampled residual $d_t(a_t)$, which still contains the state value $\langle p_t,d_t\rangle$. This value is fixed before the token is chosen, so it rewards whatever token is sampled at a prefix where the teacher concentrates on the student's likely tokens, and response centering then spreads this prefix-level effect over the other positions (Proposition~\ref{prop:response}). Removing both operations costs $1.73$ Avg@8 and $2.77$ Pass@8 points; the state value then also enters the response mean, shifting it by $-0.32$ on average and beyond $0.5$ on a quarter of the offline responses (Table~\ref{tab:response-mean}).

\begin{table}[!htb]
\caption{\textbf{Every residual variant improves on verifier-only training at every tested weight, and full \RA{} stays above every control at every matched weight.} Qwen3-1.7B-Base with REINFORCE++; three-seed macro means in \%. The weight is $\lambda$, $\gamma_x$, $\gamma_{\rm raw}$, or $\alpha$ as marked, with $10$ the reference weight; per-seed values and sample SD are in Table~\ref{tab:sweep-mass}.}
\label{tab:weight-sweep}
\centering
\begingroup
\small
\setlength{\tabcolsep}{5pt}
\renewcommand{\arraystretch}{1.05}
\begin{tabular}{@{}lC{0.066\linewidth}C{0.066\linewidth}C{0.066\linewidth}@{\hspace{12pt}}C{0.066\linewidth}C{0.066\linewidth}C{0.066\linewidth}@{}}
\toprule
\tabletwoheadshade{tableheadcol}
\tabletwoheadlabel{Variant (weight)} & \multicolumn{3}{c}{Macro Avg@8 at weight} & \multicolumn{3}{c}{Macro Pass@8 at weight}\\
\cmidrule(lr){2-4}\cmidrule(l){5-7}
 & 5 & 10 & 20 & 5 & 10 & 20\\
\midrule
\tablerowshade{reinforcecol}Full \RA{} ($\lambda$) & 28.84 & 29.25 & 29.29 & 44.94 & 45.20 & 45.29\\
w/o student value baseline ($\gamma_x$) & 26.96 & 27.16 & 26.53 & 41.93 & 41.62 & 40.76\\
w/o response centering ($\lambda$) & 28.37 & 28.71 & 28.66 & 44.46 & 44.74 & 44.67\\
Raw residual (w/o both, $\gamma_{\rm raw}$) & 27.48 & 27.52 & 26.96 & 42.63 & 42.43 & 40.96\\
\midrule
Direct $L_2$ loss ($\alpha$) & 27.83 & 28.02 & 27.71 & 43.39 & 43.44 & 42.96\\
\midrule
Verifier only (REINFORCE++) & \multicolumn{3}{c}{25.67} & \multicolumn{3}{c}{39.44}\\
\bottomrule
\end{tabular}
\endgroup

\end{table}

\paragraph{Response centering.}
Restoring response centering at the same weight adds $0.54$ Avg@8 and $0.46$ Pass@8 points, a smaller gain that holds at all three weights (Table~\ref{tab:weight-sweep}). Without centering, the response's mean label is $A^V+\lambda\overline A^{d}$. On the 256 offline responses of Appendix~\ref{app:buckets}, the local advantages average to $3\times10^{-4}$ per token, so the student value baseline leaves no systematic part, and on responses longer than $1{,}000$ tokens the shift has an SD below $0.03$, where centering changes almost nothing (Proposition~\ref{prop:response}). On shorter responses, where few tokens set the average, the SD is $0.32$, and $8\%$ of all responses carry a shift beyond $0.5$, the scale of a verifier advantage (Table~\ref{tab:response-mean}). Without centering these responses receive a response-level label the verifier did not assign, which can even reverse the sign of the verifier's judgment; centering removes this shift deterministically for every response (Equation~\ref{eq:mean}), so the response-level label no longer depends on how the guidance happens to average along a response. Across seeds, full \RA{} also varies less than the uncentered control in macro Avg@8 at every weight (Table~\ref{tab:sweep-mass}).

\paragraph{Guidance weight.}
All tested weights improve both macro scores over verifier-only training (Table~\ref{tab:weight-sweep}). Halving the reference weight lowers both means; doubling it leaves them nearly unchanged ($29.29$ vs.\ $29.25$ Avg@8) but widens the seed SD from $0.25$ to $1.68$, as in every variant (Table~\ref{tab:sweep-mass}).

\subsection{Probability residual versus log ratio}
\label{sec:alternatives}

\begin{table}[!htb]
\caption{\textbf{The probability residual with both baselines is the best construction, and no log-ratio variant reaches it at any swept weight.} Three-seed macro means in \% with SD; weights marked $^{\dagger}$ were chosen on the validation split (Appendix~\ref{app:validation}), and Tables~\ref{tab:sweep-mass} and~\ref{tab:sweep-log} give all swept weights.}
\label{tab:alternatives}
\centering
\begingroup
\small
\setlength{\tabcolsep}{0pt}
\renewcommand{\arraystretch}{1.06}
\begin{tabular}{@{}p{0.30\linewidth}C{0.12\linewidth}C{0.16\linewidth}C{0.14\linewidth}R{0.14\linewidth}R{0.14\linewidth}@{}}
\toprule
Variant & Reward & Baselines & Weight & Avg@8 & Pass@8\\
\midrule
\tablerowshade{reinforcecol}Full \RA{} & $q{-}p$ & both & 10 & 29.25\sd{0.25} & 45.20\sd{1.81}\\
Clipped centered log ratio ($c{=}5$) & $\log(q/p)$ & both & $0.6^{\dagger}$ & 28.04\sd{0.28} & 43.50\sd{0.59}\\
Centered log ratio & $\log(q/p)$ & both & $0.6^{\dagger}$ & 27.85\sd{0.37} & 43.07\sd{0.21}\\
Raw residual (w/o both) & $q{-}p$ & neither & 10 & 27.52\sd{0.43} & 42.43\sd{0.79}\\
Raw log ratio & $\log(q/p)$ & neither & $0.2^{\dagger}$ & 27.28\sd{0.37} & 42.50\sd{0.68}\\
\midrule
Verifier only (REINFORCE++) & \textcolor{black!45}{--} & \textcolor{black!45}{--} & \textcolor{black!45}{--} & 25.67\sd{0.66} & 39.44\sd{1.65}\\
\bottomrule
\end{tabular}
\endgroup

\end{table}

\paragraph{Log ratio.}
Table~\ref{tab:alternatives} replaces the residual by the log ratio at the sampled token, raw or with both operations. The raw log ratio helps only at small weights, peaking at $\beta=0.2$, and falls below verifier-only training from $\beta=1.2$ on; its conditional mean at a prefix is $-\KL(p_t\Vert q_t)$ (Appendix~\ref{app:response-composition}), so it shifts the mean label in proportion to $\beta$. With both operations, clipped to $[-5,5]$ or not, the log ratio stays above verifier-only training at every swept weight, and the probability residual is better still, by $1.21$ Avg@8 and $1.70$ Pass@8 points over the best log variant.

\paragraph{Direct loss.}
A full-vocabulary squared-distance loss, whose gradient at each visited prefix equals the expected update of the uncentered residual label (Appendix~\ref{app:local-objective}), beats verifier-only training but stays below \RA{} and the uncentered control at every weight (Table~\ref{tab:weight-sweep}). The gap lies not in the expected teacher gradient, which the sampled label matches (cosine $0.99$ with its analytic value), but in how the update is shared with the verifier (Appendix~\ref{app:shared-path}). Because \RA{}'s teacher term acts on the tokens the verifier scores, $64\%$ of its gradient mass raises correct or lowers incorrect responses, against $49\%$ for the direct loss, and after clipping and Adam the teacher takes $0.28$ of the update against $0.44$, keeping \RA{}'s update closer to the verifier's direction (cosine $0.93$ against $0.80$).

\subsection{Where the guidance lands}
\label{sec:gate}

By Equation~\ref{eq:factor} and Proposition~\ref{prop:advantage}(ii), the teacher moves logit $b$ in expectation by $p(b)(1-p(b))$ times a bounded contrast. The student-centered log ratio factorizes the same way, but its contrast is unbounded in both directions (Appendix~\ref{app:gate}). Table~\ref{tab:buckets} (Appendix~\ref{app:buckets}) describes the start of training, bucketing the tokens of 256 offline responses of the untrained 1.7B student by the student's probability of the sampled token. Confident tokens, with $p(a_t)\ge0.9$, are 61\% of all tokens and receive under 1\% of either label, as the factor $1-p(a)$ predicts. The residual puts 59\% of its guidance on the 12\% of tokens with $0.1\le p(a_t)<0.5$, five times their share, where the student is choosing among live alternatives. The centered log ratio puts 50\% of its total on the 4\% of tokens the teacher assigns less than $10^{-4}$, where its label diverges. The residual thus weights the forks of the student's own reasoning, consistent with the finding that a few high-entropy tokens carry most of the useful RL signal \citep{wang2025entropy}, whereas the log ratio puts it on tokens that one side considers nearly impossible.

\subsection{More guidance is not always better}
\label{sec:branch}

\RA{} assigns the teacher a restrained role. It labels only the tokens the student samples, so where the teacher clearly disagrees, its preferred token receives no supervision of its own unless the student samples it. A natural alternative is denser, more direct supervision that prescribes the student's actions rather than only evaluating them; we examine it through a minimal extension of the sampled route, a teacher-peak branch. At each position where the teacher's top token $v_t^\star=\arg\max_v q_t(v)$ differs from the student's top token and from the sampled token, which may itself be the student's top token, the branch adds $v_t^\star$ as a pseudo-sample. The added token receives the label a sampled token there would receive, with the verifier part clipped at zero, $\big([A^V]_++\lambda(h_t(v_t^\star)-\bar h)\big)\log\pi_\theta(v_t^\star)$, where $h_t(v)=A_t^d(v)$ and $\bar h$ is the response mean of the sampled route. The sampled route and its centering are unchanged; the branch supervises the teacher's top token as if the student had also sampled it, and its verifier part can only raise a response's total label (Appendix~\ref{app:reference-fork}).

Since the value of additional guidance may depend on the training objective, we add the same branch to three base methods. Sampled OPD (Appendix~\ref{app:sampled-opd}) learns from the teacher alone; there the added token receives the sampled-OPD label $\log q_t(v_t^\star)-\log p_t(v_t^\star)$, which makes sampled OPD a top-$k$ variant. REINFORCE++ learns from the verifier alone, so the branch is its only source of teacher signal. \RA{}+REINFORCE++ already receives a teacher label at every sampled token.

\begin{table}[!htb]
\caption{\textbf{The same teacher-peak branch slightly helps sampled OPD but lowers both students trained with the verifier.} Change in three-seed macro score when the branch is added, Qwen3-1.7B-Base, in points. The weight $\lambda$ multiplies the branch's teacher part and, with \RA{}, also the sampled guidance. Absolute, benchmark, and per-seed scores are in Appendix~\ref{app:branch-results}.}
\label{tab:branch}
\centering
\begingroup
\small
\setlength{\tabcolsep}{4pt}
\renewcommand{\arraystretch}{1.05}
\begin{tabular}{@{}lC{0.13\linewidth}@{\hspace{10pt}}C{0.068\linewidth}C{0.068\linewidth}C{0.068\linewidth}@{\hspace{10pt}}C{0.068\linewidth}C{0.068\linewidth}C{0.068\linewidth}@{}}
\toprule
 & \multicolumn{7}{c}{\textbf{Effect of the teacher-peak branch}}\\
\cmidrule(l){2-8}
 & Sampled OPD & \multicolumn{3}{c}{REINFORCE++} & \multicolumn{3}{c}{\RA{}+REINFORCE++}\\
\cmidrule(lr){2-2}\cmidrule(lr){3-5}\cmidrule(l){6-8}
 & no weight & $\lambda{=}5$ & $\lambda{=}10$ & $\lambda{=}20$ & $\lambda{=}5$ & $\lambda{=}10$ & $\lambda{=}20$\\
\midrule
$\Delta$ macro Avg@8 & $+0.30$ & $-0.38$ & $-0.88$ & $-1.76$ & \multicolumn{1}{>{\columncolor{reinforcecol}[4pt][4pt]}C{0.068\linewidth}}{$-2.05$} & \multicolumn{1}{>{\columncolor{reinforcecol}[4pt][4pt]}C{0.068\linewidth}}{$-2.07$} & \multicolumn{1}{>{\columncolor{reinforcecol}[4pt][0pt]}C{0.068\linewidth}@{}}{$-2.16$}\\
$\Delta$ macro Pass@8 & $+0.64$ & $-0.46$ & $-0.86$ & $-2.11$ & \multicolumn{1}{>{\columncolor{reinforcecol}[4pt][4pt]}C{0.068\linewidth}}{$-3.91$} & \multicolumn{1}{>{\columncolor{reinforcecol}[4pt][4pt]}C{0.068\linewidth}}{$-3.41$} & \multicolumn{1}{>{\columncolor{reinforcecol}[4pt][0pt]}C{0.068\linewidth}@{}}{$-3.89$}\\
\bottomrule
\end{tabular}
\endgroup

\end{table}

Under sampled OPD (Table~\ref{tab:branch}) the branch raises macro Avg@8 from $24.80$ to $25.10$ and the Avg@8 of every seed. Under REINFORCE++, where $\lambda$ scales only the branch, it lowers macro Avg@8 by $0.38$--$1.76$ points and Pass@8 by $0.46$--$2.11$ points, a loss that grows with the branch weight. Added to \RA{}, it lowers macro Avg@8 by $2.05$--$2.16$ points and Pass@8 by $3.41$--$3.91$ points, and at every weight each seed falls below the same seed of \RA{}, whereas doubling \RA{}'s own weight does not (Table~\ref{tab:weight-sweep}).

Because the branch alters several factors at once, it does not isolate a mechanism, and we do not pursue one here. It does answer the question it was designed for: stronger and denser teacher supervision does not by itself yield better performance. A student trained only to imitate the teacher benefits slightly from the teacher's top token, whereas both students trained with the verifier degrade, increasingly so with the branch weight under REINFORCE++. One conjecture is that steering the student directly toward the teacher's choice crowds out the exploration on which RL relies. \RA{} instead scales its expected update on each token by the student's probability of that token and leaves the choice to the student's interaction with the verifier; amplifying this gated signal is harmless. This restraint may be part of what makes its guidance effective, and it leaves an open question for RLVR: how a dense, critic-like signal such as a teacher's should be shaped and enter RL training.

\section{Related Work}

\paragraph{Distilling on the student's own samples.}
GKD studies divergence choices on student-generated sequences \citep{gkd}, MiniLLM optimizes reverse KL with log-ratio returns \citep{minillm}, sampled OPD uses the log ratio at the generated token as a dense reward \citep{lu2025opd}, and OPD+ derives divergence-dependent corrections \citep{opdplus2026}. \RA{} replaces the log ratio by the bounded residual, judges the chosen token against the student's own alternatives, and removes the teacher's response mean. What distinguishes \RA{} is its restraint: the teacher acts not as a distillation target, as in these works, but as a critic-like signal within RL.

\paragraph{Combining distillation with verifiable rewards.}
LUFFY mixes off-policy teacher traces into RLVR \citep{yan2025luffy}. OPDVR gates the sampled log ratio so that token reward signs agree with the verifier \citep{opdvr2026}, whereas \RA{} keeps the verifier advantage as the response mean and allows local sign changes. \citet{li2026sequential} favor OPD followed by RL over joint combinations; at our matched budget both schedules stay below \RA{} (Appendix~\ref{app:sequential}).

\paragraph{Baselines and token-level credit.}
A state-value baseline leaves the expected policy gradient unchanged \citep{williams1992simple,sutton2000pg}, and the variance-optimal baseline also accounts for the score norm \citep{greensmith2004variance}. GRPO, REINFORCE++, RLOO, and ReMax use group, batch, leave-one-out, and greedy baselines \citep{deepseekmath,hu2025reinforcepp,ahmadian2024rloo,li2023remax}, and COMA marginalizes an action under the policy \citep{foerster2018coma}; \RA{} uses an exact policy expectation as the baseline of its local reward. Process supervision uses finer labels \citep{lightman2023lets}, VinePPO extra sampling \citep{kazemnejad2024vineppo}, and implicit PRMs and PRIME the log ratio of an outcome-trained model to a reference \citep{yuan2024implicitprm,cui2025prime}. VPD refines a feedback-conditioned teacher with outcome supervision and distills it on student rollouts \citep{li2026vpd}; \CoRA{} instead trains a teacher LoRA with the base estimator's clipped update on the same scored student batch.

\section{Conclusion}

Residual Advantage judges the teacher--student probability difference against the student's own alternatives and centers it within each response, so the verifier sets a response's mean label and the teacher allocates credit among its steps. The guidance term decreases both as the student approaches the teacher and as the student becomes confident in its own choice, an elegant property that helps \RA{} preserve exploration in RL. Across two students, two estimators, and three benchmarks, \RA{} improves every matched three-seed mean, both combinations surpass teacher-only OPD, and \CoRA{} improves accuracy further while its adapted teacher becomes a worse solver than the frozen one. Up to Pass@512 the margin of \RA{} over the untrained student grows while that of verifier-only training narrows. Removing either baseline, replacing the residual by the log ratio, or adding direct supervision of the teacher's top token lowers performance.

\newpage
\subsection*{AI use statement}
Generative AI tools were used throughout this work. In this work, we used generative AI tools to help develop the theoretical framework, formulate mathematical claims, provide ingredients for and assist in the writing of proofs, propose and refine hypotheses, design and give feedback on the research methodology and experiments, implement methods, assist with translation, and interpret results. We have not used generative AI tools to generate synthetic data sets; cleaning or reformatting datasets and qualitative or thematic data analysis are not applicable to this work, which uses public datasets unmodified and reports quantitative results only. Additionally, we used generative AI tools to brainstorm and identify research gaps, source information and identify, summarize, and analyze related literature, suggest experimental parameters, create and edit software code, create and modify figures and tables, suggest the structure of the paper, draft and edit parts of the text to improve readability, format references, and propose the title. We have reviewed all AI-assisted work: every derivation and proof was rederived and checked by the authors, all cited works were verified against their original sources, all code was reviewed by the authors and every reported experiment was run and evaluated under the authors' control, and every number in the paper was checked against the recorded results. We take responsibility for the final content of this work, including text, claims, and artifacts produced with the aid of generative AI.

\subsection*{Acknowledgments}
Part of this work was done at Tencent. We thank Tencent for providing the computational resources used in this work.

\appendix
\section{Proofs and update identities}
\label{app:proofs}

Throughout, $p,q$ are probability vectors on the same finite support with
at least two coordinates, $d=q-p$, $A^{d}(a)=\langle e_a-p,d\rangle$, and
$\lambda\ge0$. Labels are held fixed during differentiation, expectations
over actions use the same $p$ that defines the label, and scores are
evaluated at the rollout policy. Probability identities also hold on the
boundary of the simplex; logit identities use positive probabilities. The
constructed examples are exact computations on stipulated vectors.

\subsection{Proof of Proposition~\ref{prop:advantage}}
\label{app:probability-path}

\paragraph{Part (i).}
Let $L(p,q)=\frac12\lVert p-q\rVert_2^2$. The path
$p_\eta=(1-\eta)p+\eta e_a$ stays in the simplex for $0\le\eta\le1$, and
$p_\eta-q=-d+\eta(e_a-p)$, so
\begin{equation}
 L(p_\eta,q)-L(p,q)
 =-\eta\langle e_a-p,d\rangle
   +\frac{\eta^2}{2}\lVert e_a-p\rVert_2^2
 =-\eta A^{d}(a)+\frac{\eta^2}{2}\lVert e_a-p\rVert_2^2.
 \label{eq:app-feasible-path}
\end{equation}
Thus $A^{d}(a)$ is the rate of improvement along a probability move toward
$a$, and if $A^{d}(a)>0$ every feasible
$\eta<2A^{d}(a)/\lVert e_a-p\rVert_2^2$ strictly reduces $L$.

\paragraph{Part (ii).}
For $p=\operatorname{softmax}(z)$, write
$J_p=\operatorname{Diag}(p)-pp^\top$. The student mean of the label is
$\sum_a p(a)A^{d}(a)=\langle p,d\rangle-\langle p,d\rangle=0$, and since
$\nabla_z\log p(a)=e_a-p$,
\begin{equation}
 \mathbb E_{A\sim p}[A^{d}(A)(e_A-p)]
 =p\odot A^{d}=J_pd=-\nabla_z L(p,q),
 \qquad -\frac{\partial L}{\partial z_a}=p(a)A^{d}(a).
 \label{eq:app-expected-gradient}
\end{equation}
Here $J_p=\E_{a\sim p}[(e_a-p)(e_a-p)^\top]$ is the Jacobian of the softmax
and the Fisher information of the categorical distribution in logit
coordinates, so the logit gradient is $J_p$ applied to the
probability-space gradient $\nabla_pL=-d$. The same computation shows that
any fixed label $w$ has expected update $J_pw$ and that subtracting an
action-independent baseline $\langle p,w\rangle$ leaves it unchanged,
because $\E_{a\sim p}[e_a-p]=0$; in particular $d$ and $A^{d}$ have the
same expected update at a fixed prefix.

\paragraph{Part (iii).}
Cauchy--Schwarz gives
$|\langle e_a-p,d\rangle|\le\lVert e_a-p\rVert_2\lVert d\rVert_2$ with
$\lVert e_a-p\rVert_2^2=(1-p(a))^2+\sum_{v\ne a}p(v)^2\le2(1-p(a))^2$, since
$\sum_{v\ne a}p(v)^2\le(1-p(a))^2$; equality holds when the student's
remaining mass sits on one token. Separating $a$ in
$A^{d}(a)=d(a)-\sum_vp(v)d(v)$ gives Equation~\ref{eq:factor}. The range
$[-\tfrac12,2]$ is established in Appendix~\ref{app:mass-bounds}.

\paragraph{Other divergences.}
For any $\ell$ differentiable near the interior point $p$, define
$A^{\ell}(a)=-\langle e_a-p,\nabla_p\ell\rangle$, the negative directional
derivative of $\ell$ along $e_a-p$. This label has zero student mean, and
\begin{equation}
 \E_{a\sim p}\big[A^{\ell}(a)(e_a-p)\big]=-J_p\nabla_p\ell=-\nabla_z\,\ell(\operatorname{softmax}(z))
 \label{eq:app-generic-linearization}
\end{equation}
by the chain rule. For $\ell=L$ this recovers $A^{L}=A^{d}$. For
$\ell=\KL(p\Vert q)$, $\nabla_p\ell=\log(p/q)+\mathbf1$, the constant is
annihilated by $\langle e_a-p,\cdot\rangle$, and
$A^{\ell}(a)=\log\frac{q(a)}{p(a)}-\langle p,\log\frac qp\rangle$ is the
student-centered log ratio. The quadratic remainder in
Equation~\ref{eq:app-feasible-path} is specific to $L$.

The identity concerns expectations. A single sampled ascent vector
$g_a=A^{d}(a)(e_a-p)$ can point against $J_pd$, as in
Table~\ref{tab:app-sampled-case}, where a small logit step along $g_a$
increases $L$.

\begin{table}[!htbp]
\caption{\textbf{A sampled update that opposes the expected update.}
The probability vectors and sampled action are inputs; the remaining
entries follow by substitution.}
\label{tab:app-sampled-case}
\centering\small
\begin{tabular}{@{}lll@{}}
\toprule
\tablerowshade{tableheadcol}
Role & Quantity & Value \\
\midrule
Input & Student $p$ & $(0.1,0.3,0.6)$ \\
Input & Teacher $q$ & $(0.2,0.1,0.7)$ \\
Input & Sampled action $a$ & $1$ \\
\midrule
Output & Local advantage $A^{d}(a)$ & $0.09$ \\
Output & Expected update $J_pd$ & $(0.009,-0.063,0.054)$ \\
Output & Sampled update $g_a$ & $(0.081,-0.027,-0.054)$ \\
Output & Inner product $\langle J_pd,g_a\rangle$ & $-0.000486$ \\
\bottomrule
\end{tabular}
\end{table}

\paragraph{The log ratio as an implicit advantage.}
\label{app:implicit-advantage}
Suppose the teacher is a KL-regularized improvement of the student at this
prefix, $q=p\exp(A^\star/\beta)/Z$ with score vector $A^\star$, $\beta>0$,
and $Z=\langle p,\exp(A^\star/\beta)\rangle$. Then
\begin{align}
 \log\frac{q(a)}{p(a)}-\left\langle p,\log\frac qp\right\rangle
 &=\frac{A^\star(a)-\langle p,A^\star\rangle}{\beta},\nonumber\\
 d(a)&=p(a)\left[\frac{e^{A^\star(a)/\beta}}{Z}-1\right]\ge-p(a),
 \label{eq:app-implicit-readouts}
\end{align}
so the student-centered log ratio recovers the score contrast exactly,
while the residual passes it through a student-weighted saturating map.
For fixed $A^\star$, expanding at $\beta^{-1}=0$ gives
\begin{align}
 \frac{\exp(A^\star/\beta)}{Z}
 &=\mathbf1+\frac{A^\star-\langle p,A^\star\rangle\mathbf1}{\beta}
   +O(\beta^{-2}),\nonumber\\
 d&=\frac{J_pA^\star}{\beta}+O(\beta^{-2}).
 \label{eq:app-implicit-linearization}
\end{align}
This is the implicit-advantage model behind on-policy distillation and
preference optimization \citep{rafailov2023dpo,rafailov2024rtoq,peng2019awr},
in which a teacher one ideal KL step ahead of the student carries a score
contrast in its log ratio. A frozen solver need not be such a step for the
current student, and \RA{} builds its local advantage from the probability
residual instead.

\subsection{A local objective and its target}
\label{app:local-objective}

\begin{proposition}[Local quadratic surrogate]
\label{prop:target}
Fix a prefix, the teacher $q$, a finite verifier action-value vector $Q$,
and $\lambda>0$. At the rollout policy, the expected logit update of
the local label $A^V+\lambda A^{d}(a)$, before response centering, is
$\nabla_zF(p)$ for
\begin{equation}
 F(p)=\langle p,Q\rangle-\frac\lambda2\lVert p-q\rVert_2^2.
 \label{eq:local-objective}
\end{equation}
On the closed simplex, this strictly concave function has the unique maximizer
\begin{equation}
 p^\star=\Pi_\Delta(q+Q/\lambda),\qquad
 p^\star(v)=[q(v)+Q(v)/\lambda-\tau]_+,
 \label{eq:projection-target}
\end{equation}
where $\tau$ makes the entries sum to one. At fixed $Q$ and $\lambda$,
the map $q\mapsto p^\star$ is nonexpansive in $\ell_2$.
\end{proposition}

\paragraph{Proof.}
Let $Q(v)=\E[A^V\mid\xi,a_t=v]$. Conditioning on the sampled action gives
$\E[A^V(e_a-p)]=J_pQ$, and with Equation~\ref{eq:app-expected-gradient},
\begin{equation}
 \E\big[(A^V+\lambda A^{d}(a))(e_a-p)\big]=J_p(Q+\lambda d),
 \label{eq:app-local-update}
\end{equation}
which is $\nabla_zF(p)=J_p\nabla_pF$ because $\nabla_pF=Q+\lambda d$.
Completing the square,
\begin{equation}
 F(p)=-\frac\lambda2\Big\lVert p-\big(q+Q/\lambda\big)\Big\rVert_2^2
      +\langle q,Q\rangle+\frac{\lVert Q\rVert_2^2}{2\lambda},
 \label{eq:app-complete-square}
\end{equation}
so for $\lambda>0$ the unique maximizer on the simplex is the Euclidean
projection of $q+Q/\lambda$, which has the threshold form of
Equation~\ref{eq:projection-target} \citep{michelot1986,duchi2008} and
coincides with sparsemax \citep{martins2016sparsemax}. Projection onto a
closed convex set is nonexpansive, which gives the last claim. If
$\lambda=0$ the objective is linear and any distribution on the maximizers
of $Q$ is optimal.

\paragraph{Support.}
$p^\star(v)=0$ if and only if $q(v)+Q(v)/\lambda\le\tau$. A token outside
the teacher's support is retained exactly when $Q(v)>\lambda\tau$, a
teacher-favored token is dropped exactly when $Q(v)\le-\lambda(q(v)-\tau)$,
and $p^\star=q$ when $Q\equiv0$.

\paragraph{Log ratio.}
For positive $p,q$ and $\beta>0$, the label $A^V+\beta\log(q/p)$ has
expected update $\nabla_z[\langle p,Q\rangle-\beta\KL(p\Vert q)]$, because
$J_p\mathbf1=0$. With $p_\beta^\star=q\exp(Q/\beta)/Z$ and
$Z=\langle q,\exp(Q/\beta)\rangle$,
\begin{equation}
 \langle p,Q\rangle-\beta\KL(p\Vert q)
 =\beta\log Z-\beta\KL(p\Vert p_\beta^\star),
 \label{eq:app-tilt-optimum}
\end{equation}
so $p_\beta^\star$ is the unique maximizer. If $q$ has zero coordinates the
statement holds on its support, and finite KL excludes mass outside it.
Both maximizers describe a local optimum at one prefix; which tokens a
finite training run discovers is a separate question.

\paragraph{Scope.}
The vector $Q$ is frozen at the rollout policy; the surrogate does not differentiate its dependence on the policy or on the visited prefixes. Under the assumptions of Proposition~\ref{prop:response}, each position's expected teacher term carries the factor $1-c_t$ of Equation~\ref{eq:app-fixed-response}, so the sum of local objectives uses the effective weight $\lambda(1-c_t)$ at position $t$. The random lengths and batch normalizer $Z$ of the actual protocol fall outside these hypotheses (Appendix~\ref{app:response-composition}). The direct $L_2$ control applies $\alpha J_pd$ at each visited prefix through a differentiable loss and matches the uncentered local expectation at $\alpha=\lambda$; response centering, gradient noise, and optimizer state still separate the two training updates.

\subsection{Local probability scaling}
\label{app:gate}

\begin{proposition}[Local logit influence]
\label{prop:gate}
Fix a prefix, positive distributions $p,q$, a finite verifier action-value
vector $Q$, and nonnegative weights $\lambda,\beta$, before response
centering.
\textup{(i)} On logit coordinate $b$, the expected residual update is
$\lambda p(b)A^{d}(b)$, with
\[
 -\tfrac12\lambda p(b)\le\lambda p(b)A^{d}(b)
 \le2\lambda p(b)(1-p(b))^2\le\tfrac{8}{27}\lambda,
\]
and the verifier contribution is $p(b)C_Q(b)$, where
$C_Q(b)=Q(b)-\langle p,Q\rangle$.
\textup{(ii)} The raw or student-centered log-ratio label gives
$\beta p(b)[\log(q(b)/p(b))+\KL(p\Vert q)]$ on that coordinate. For fixed
positive $q$ and $p(b)\to0$ with the other probabilities proportionally
rescaled, this is $\beta p(b)[\log(1/p(b))+O(1)]$.
\end{proposition}

\paragraph{Proof.}
Coordinate $b$ of Equation~\ref{eq:app-expected-gradient} is
$p(b)A^{d}(b)$, and Equation~\ref{eq:app-range} bounds the label; the
function $2x(1-x)^2$ on $[0,1]$ has derivative $2(1-x)(1-3x)$ and maximum
$8/27$ at $x=1/3$. Coordinate $b$ of $J_pQ$ is $p(b)C_Q(b)$. For the log
ratio, coordinate $b$ of $J_p\ell$ is $p(b)(\ell(b)-\langle p,\ell\rangle)$
with $\langle p,\ell\rangle=-\KL(p\Vert q)$, which stays finite in the
stated limit, so the bracket equals $\log(1/p(b))+O(1)$. Relative to a
nonzero verifier coordinate its magnitude is
$\beta|\log(1/p(b))+O(1)|/|C_Q(b)|$, unbounded for fixed $Q$, whereas
$|A^{d}(b)|\le2$.

\paragraph{Relative magnitude.}
Both the verifier and residual terms are $O(p(b))$, but equal order bounds
neither their ratio nor the sign of their sum. At $p=(.99,.01)$,
$q=(.01,.99)$, $Q=(0,-.01)$, and $\lambda=10$, the second coordinate
receives $+0.19404$ from the residual and $-0.000099$ from the verifier,
so the teacher reverses the net direction. A sufficient condition for a
positive combined update on coordinate $b$ is $C_Q(b)>\lambda/2$, since
$C_Q(b)+\lambda A^{d}(b)\ge C_Q(b)-\lambda/2$; at $\lambda=10$ the
threshold is $5$, and under the fixed-weight composition of
Proposition~\ref{prop:response} the effective weight is $\lambda(1-c_t)$.
These are statements about the expected logit update at the rollout
policy, and they say nothing about entropy, exploration, or a finite
parameter step.

Response centering is a different operation from the student value
baseline. Applied to the student-centered log label it gives the same
$1-c_t$ factor as Equation~\ref{eq:app-fixed-response}; applied to the raw
log label it can introduce cross-position terms, because the raw
conditional mean $-\KL(p_t\Vert q_t)$ depends on earlier actions
(Appendix~\ref{app:response-composition}).

\subsection{The residual and the log ratio as ends of one family}
\label{app:units}

For positive distributions and $1<\alpha\le2$, define
\begin{equation}
 D_\alpha(p,q)=\frac{\sum_v[p_v^\alpha-q_v^\alpha-\alpha q_v^{\alpha-1}(p_v-q_v)]}
 {\alpha(\alpha-1)},
\end{equation}
the separable Bregman divergence generated by $x^\alpha/[\alpha(\alpha-1)]$.
Its negative student-probability derivative is the label
\begin{equation}
 w_\alpha(v)=\frac{q(v)^{\alpha-1}-p(v)^{\alpha-1}}{\alpha-1},\qquad 1<\alpha\le2,
 \label{eq:power-unit}
\end{equation}
whose limit $w_1=\log(q/p)$ is the OPD label and whose endpoint $w_2=q-p$
is the residual, with $D_1=\KL(p\Vert q)$ and
$D_2=\frac12\lVert p-q\rVert^2$. By Equation~\ref{eq:app-generic-linearization},
\begin{align}
 -\left.\frac{d}{d\eta}D_\alpha(p+\eta(e_a-p),q)\right|_{\eta=0}
 &=w_\alpha(a)-\langle p,w_\alpha\rangle,\nonumber\\
 \E_p[(w_\alpha(A)-\E_pw_\alpha)(e_A-p)]&=-\nabla_zD_\alpha,
\end{align}
so the path and expected-gradient readings of
Proposition~\ref{prop:advantage} hold throughout the family. The teacher
sensitivity $\partial w_\alpha(v)/\partial q(v)=q(v)^{\alpha-2}$ is
unbounded at a teacher-rare token for every $\alpha<2$ and constant at the
quadratic endpoint, the only member of the family with bounded teacher
influence.

At $p=(1-\epsilon,\epsilon)$ and $q=(\epsilon,1-\epsilon)$, set
$D=1-2\epsilon$ and $L=\log[(1-\epsilon)/\epsilon]$. The student-centered
labels of the confidently wrong mode are $-2\epsilon D$ for the residual
and $-2\epsilon L$ for the log ratio; the sampled gradients have norms
$2\sqrt2\epsilon^2D$ and $2\sqrt2\epsilon^2L$, and the expected gradients
$2\sqrt2\epsilon(1-\epsilon)D$ and $2\sqrt2\epsilon(1-\epsilon)L$. Since
$D/L\to0$, the residual pushes a confidently wrong mode more gently than
the log ratio. These statements precede response centering, which can
transfer a label from another position into a confident one.

\begin{figure}[!htbp]
\centering
\includegraphics[width=0.82\linewidth]{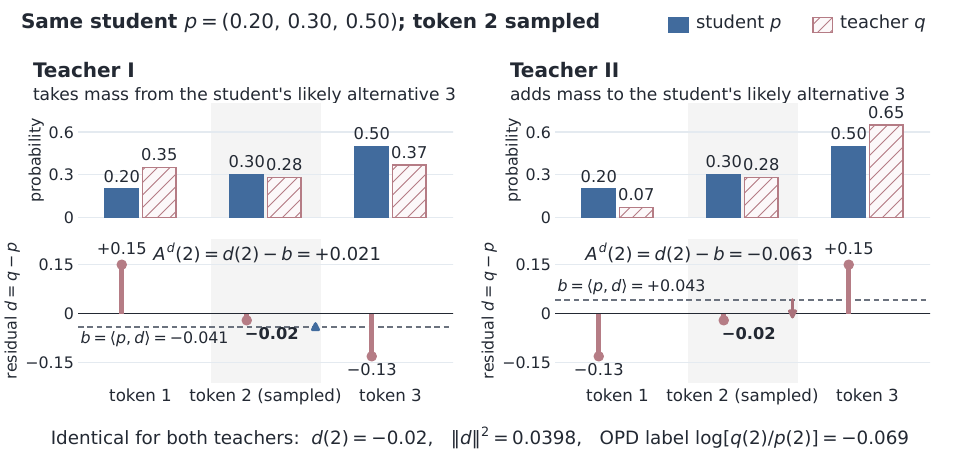}
\caption{\textbf{The same sampled reward can yield local advantages of opposite signs.}
Both teachers assign the sampled token probability $0.28$ against the student's $0.30$, giving the same log ratio ($-0.069$) and residual ($-0.02$). Their residual vectors have the same squared norm, $0.0398$, but the student values differ. The resulting $A^d(2)$ is $+0.021$ for teacher I and $-0.063$ for teacher II.}
\label{fig:configuration}
\end{figure}

\paragraph{A sampled reward and its local advantage.}
Figure~\ref{fig:configuration} holds the student's distribution and the teacher's probability of the sampled token fixed while changing the teacher's alternatives. Both teachers give the same sampled log ratio and the same residual, but different student values yield local advantages of opposite signs. This is the usual distinction between $Q$ and $Q-V$; in expectation, sampled OPD still follows the negative full-vocabulary reverse-KL gradient.

Table~\ref{tab:configuration} lists the inputs and derived quantities of
the two teachers in Figure~\ref{fig:configuration}.

\begin{table}[!htbp]
\caption{\textbf{The two teachers of Figure~\ref{fig:configuration}.}
The student distribution and sampled action are shared. The last row pairs
the two teachers as positions of one response with equal position weights
and unit guidance weight.}
\label{tab:configuration}
\centering\small
\begin{tabular}{@{}lcc@{}}
\toprule
\tablerowshade{tableheadcol}
Quantity & Teacher I & Teacher II\\
\midrule
Student $p$ & $(0.20,0.30,0.50)$ & $(0.20,0.30,0.50)$\\
Teacher $q$ & $(0.35,0.28,0.37)$ & $(0.07,0.28,0.65)$\\
Residual $d=q-p$ & $(+0.150,-0.020,-0.130)$ & $(-0.130,-0.020,+0.150)$\\
Sampled action $a$ & 2 & 2\\
Sampled residual $d(a)$ & -0.020 & -0.020\\
Squared residual norm $\|d\|_2^2$ & 0.0398 & 0.0398\\
OPD label $\log[q(a)/p(a)]$ & -0.0690 & -0.0690\\
Student value baseline $b$ & -0.041 & +0.043\\
Local advantage $A^{d}(a)$ & +0.021 & -0.063\\
Position weight / guidance weight & $1/2$ / $1$ & $1/2$ / $1$\\
Paired-response $\delta$ & +0.042 & -0.042\\
\bottomrule
\end{tabular}

\end{table}

\subsection{Label centering and gradient variance}
\label{app:variance}

For $\psi(A)=\nabla_\theta\log p(A)$ and an action-independent scalar
$\nu$, let $g_\nu=(d(A)-\nu)\psi(A)$. Its mean does not depend on $\nu$,
and completing the square gives
\begin{equation}
 \nu_* =\frac{\E_p[d(A)\|\psi(A)\|^2]}{\E_p\|\psi(A)\|^2},\qquad
 \operatorname{tr}\Var(g_\nu)=\operatorname{tr}\Var(g_{\nu_*})
   +\E_p\|\psi\|^2(\nu-\nu_*)^2.
\end{equation}
Thus $b=\E_p d$ minimizes $\E_p(d-\nu)^2$, whereas the gradient-variance
optimum weights by score norm and may differ. For binary logits, write
$p=(x,1-x)$ and $d=(u,-u)$. Then $b=(2x-1)u$, $\nu_*=-b$, and
\begin{equation}
 \operatorname{tr}\Var(g_0)=2x(1-x)b^2,\qquad
 \operatorname{tr}\Var(g_b)=8x(1-x)b^2.
\end{equation}
At a single prefix the student value baseline therefore has four times
the raw trace variance whenever the latter is nonzero. It is chosen for
the conditional zero mean that Proposition~\ref{prop:response} requires,
and that property has this variance cost.

\begin{table}[!htbp]
\caption{\textbf{A confident student.}
The sampled and expected student-centered logit updates are computed
from the stipulated probability vectors.}
\label{tab:app-variance-case}
\centering\small
\begin{tabular}{@{}lll@{}}
\toprule
\tablerowshade{tableheadcol}
Role & Quantity & Value \\
\midrule
Input & Student $p$ & $(0.99,0.01)$ \\
Input & Teacher $q$ & $(0.01,0.99)$ \\
\midrule
Output & Common-action update & $(-0.000196,0.000196)$ \\
Output & Expected update & $(-0.019404,0.019404)$ \\
\bottomrule
\end{tabular}
\end{table}

In Table~\ref{tab:app-variance-case}, the rare action supplies most of
the expected update despite its low sampling probability.

\subsection{Sensitivity, range, and the guidance bounds}
\label{app:mass-bounds}

The label is linear in $q$, so at fixed $p,a$ replacing $q$ by $q+e$
changes it by $\Delta A^{d}(a)=\langle e_a-p,e\rangle$ and
\begin{equation}
 |\Delta A^{d}(a)|\le\lVert e_a-p\rVert_2\lVert e\rVert_2
 =\sqrt{1-2p(a)+\lVert p\rVert_2^2}\,\lVert e\rVert_2
 \le\sqrt2\lVert e\rVert_2.
 \label{eq:app-mass-sensitivity}
\end{equation}
Moving teacher mass $\epsilon$ from an off-action token $u$ to another
off-action token $v$ changes the label by $\epsilon[p(u)-p(v)]$, a
distinction the sampled log ratio cannot make, since its teacher
sensitivity is $\partial\log(q/p)/\partial q=1/q$. Bounded sensitivity is
the bounded-influence property of the quadratic comparison
\citep{hampel1974influence,basu1998robust}, and Appendix~\ref{app:units}
places it among the alternatives.

For the range, set $x=p(a)$ and $y=\max_{v\ne a}p(v)$. The affine
function $A^{d}(a)=q(a)-\langle p,q\rangle-x+\lVert p\rVert_2^2$ is
maximized over $q$ at $q=e_a$ and minimized at an off-action coordinate
with student mass $y$, so
\begin{align}
 A^{d}(a)&\le (1-x)^2+\sum_{v\ne a}p(v)^2\le2(1-x)^2,\\
 A^{d}(a)&\ge\lVert p\rVert_2^2-x-y
 \ge(x-\tfrac12)^2+(y-\tfrac12)^2-\tfrac12\ge-\tfrac12.
 \label{eq:app-range}
\end{align}
The lower endpoint is attained by $p=(1/2,1/2)$, $q=(0,1)$, $a=1$, and the
supremum $2$ is approached by $p=(\epsilon,1-\epsilon)$, $q=(1,0)$, $a=1$
as $\epsilon\downarrow0$. For any realized nonnegative weights with
$\sum_t c_t=1$, the guidance term
$\delta_t=\lambda\sum_{u\ne t}c_u(A_t^{d}-A_u^{d})$ therefore satisfies
\begin{equation}
 \sum_t c_t\delta_t=0,\qquad
 |\delta_t|\le\frac52\lambda(1-c_t)\le\frac52\lambda,
 \label{eq:app-correction-bound}
\end{equation}
for every realized response, whatever determined its weights. With fixed
weights and a perturbation only at position $j$,
$\lVert\Delta\delta\rVert_{2,c}
=\lambda\sqrt{c_j(1-c_j)}|\Delta A_j^{d}|$, where
$\lVert v\rVert_{2,c}^2=\sum_t c_tv_t^2$.

\paragraph{Vanishing with distributional agreement.}
Cauchy--Schwarz also bounds each local advantage by the full residual,
$|A_t^d|\le\sqrt2\lVert d_t\rVert_2$, so for any realized response and its
normalized weights,
\begin{align}
 |\delta_t|
 &\le\sqrt2\lambda\left[(1-c_t)\lVert d_t\rVert_2
              +\sum_{u\ne t}c_u\lVert d_u\rVert_2\right]\nonumber\\
 &\le2\sqrt2\lambda(1-c_t)\max_u\lVert d_u\rVert_2.
 \label{eq:app-residual-vanishing}
\end{align}
For the mean absolute amplitude, weighted Cauchy--Schwarz and
$\Var_c(A^d)\le\sum_t c_t(A_t^d)^2$ give the sharper aggregate bound
\begin{equation}
 \sum_t c_t|\delta_t|
 \le\lVert\delta\rVert_{2,c}
 =\lambda\sqrt{\Var_c(A^d)}
 \le\sqrt2\lambda
       \left(\sum_t c_t\lVert d_t\rVert_2^2\right)^{1/2}.
 \label{eq:app-residual-rms}
\end{equation}
At a fixed guidance weight, the guidance term therefore vanishes
uniformly when student and teacher agree uniformly at the scored prefixes.
The implication runs one way. Small guidance can also arise when centering
cancels large, nearly equal local advantages, and neither bound says how
agreement or guidance evolve during training.

The log ratio also vanishes at exact agreement, but it has no uniform
bound of this kind in probability distance. For $p=(1-\epsilon,\epsilon)$
and $q=(1-\epsilon^2,\epsilon^2)$, $\lVert p-q\rVert_2\to0$ while the
rare-action student-centered log label
$(1-\epsilon)\log[\epsilon/(1+\epsilon)]\to-\infty$, and pairing this
position with an agreement position at equal weights gives an unbounded
response-centered log label as well.

\paragraph{Outcome and guidance components.}
With $P_c=I-\mathbf1c^\top$, the actor label has the vector form
\begin{equation}
 \mathbf A=A^V\mathbf1+\lambda P_c\mathbf A^d,\qquad c^\top P_c=0.
 \label{eq:app-centering-projection}
\end{equation}
Under $\langle u,v\rangle_c=\sum_t c_tu_tv_t$, the two components are
orthogonal because
$\langle\mathbf1,P_c\mathbf A^d\rangle_c=c^\top P_c\mathbf A^d=0$, and
expanding the squared weighted norm gives
\begin{align}
 \norm{\mathbf A}_{2,c}^2
 &=(A^V)^2+\lambda^2\Var_{t\sim c}(A_t^d),\nonumber\\
 \Var_{t\sim c}(A_t^d)&=\sum_t c_t(A_t^d-\overline A^d)^2.
 \label{eq:app-centering-decomposition}
\end{align}
Teacher guidance adds variation to the actor label while preserving its
response mean. Orthogonality of the label components does not extend to
their parameter gradients (Appendix~\ref{app:response-composition}).

\begin{table}[!htbp]
\caption{\textbf{Further quantities for the two teachers.}
The teachers are those of Figure~\ref{fig:configuration}. The final column
pairs each teacher's position with an agreement position at equal
weights $c=(1/2,1/2)$, keeping the symbolic weight $\lambda$.}
\label{tab:app-configuration-outputs}
\centering\small
\begin{tabular}{@{}lll@{}}
\toprule
\tablerowshademeasured{tableheadcol}{\shortstack[l]{First-position guidance term\\with an agreement position}}
Teacher & Expected logit update $J_pd$ &
\shortstack[l]{First-position guidance term\\with an agreement position} \\
\midrule
I & $(0.0382,0.0063,-0.0445)$ & $0.0105\lambda$ \\
II & $(-0.0346,-0.0189,0.0535)$ & $-0.0315\lambda$ \\
\bottomrule
\end{tabular}
\end{table}

The two teachers differ in both the expected logit update and the
response-centered guidance term (Table~\ref{tab:app-configuration-outputs}).
Pairing each teacher's position with an agreement position isolates its
own contribution.

\subsection{Comparing local advantages across a response}
\label{app:response-composition}

At one prefix the student value baseline leaves the expected update
unchanged (Proposition~\ref{prop:advantage}(ii)). Its role appears when
local advantages from different prefixes are combined by centering. With
$\mathcal F_t$ the prompt and actions before position $t$ and
$\psi_t=\nabla_\theta\log p_t(a_t)$, the local advantage has conditional
zero mean, $\E[A_t^d\mid\mathcal F_t]=0$, whereas the raw residual has
conditional mean $\langle p_t,d_t\rangle$, the level of disagreement at
that prefix.

\begin{proposition}[Response composition]
\label{prop:response}
Under on-policy sampling with a fixed finite horizon and trajectory-independent token weights,
\begin{equation}
 \E\Big[\sum_t c_t\delta_t\psi_t\Big]=\lambda\sum_t c_t(1-c_t)\,\E\big[A_t^d\psi_t\big].
 \label{eq:response-expectation}
\end{equation}
\end{proposition}

Local advantages and scores are both martingale differences under the
student's sampling, so every cross-position term vanishes. Without the
student value baseline this cancellation fails, and the disagreement
levels of other prefixes enter each token's update through centering. The
identity needs the stated assumptions, and the response-mean identity of
Equation~\ref{eq:mean} needs none of them.

\paragraph{What the identity says about centering.}
With token-mean weights $c_t=1/M$, the factor $1-c_t$ is within $10^{-3}$
of one for a response of a thousand tokens or more. Under the hypotheses
of Proposition~\ref{prop:response}, centering therefore leaves the expected
update almost unchanged, and the response mean it removes, $\overline A^d$,
has zero expectation and fluctuation of order $M^{-1/2}$.
Appendix~\ref{app:buckets} measures $\overline A^d$ on the offline responses.
The local advantages average to $3\times10^{-4}$ per token, so the
student value baseline holds in practice and the response mean carries no
systematic part, and on responses longer than $1{,}000$ tokens
$\lambda\overline A^d$ has an SD below $0.03$, as the hypotheses predict.
The fluctuation is not uniformly small, however. On shorter responses,
where few tokens set the average, its SD is $0.32$, and $8\%$ of all
responses carry $|\lambda\overline A^d|>0.5$, the scale of a verifier
advantage. Without centering these responses receive a response-level
label that the verifier did not assign, and Equation~\ref{eq:mean} removes
it for every response. The hypotheses of Proposition~\ref{prop:response}
thus describe the long responses, which carry most of the tokens, and the
small gain of centering in Section~\ref{sec:baselines}, about half a point
at every weight, is what acting on the rest is worth.

For a realized response, the teacher gradient at the rollout policy is
$G_d=\lambda\sum_t c_t(A_t^d-\overline A^d)\psi_t=\lambda\,\Cov_{t\sim c}(A_t^d,\psi_t)$
and the verifier's is $G_V=A^V\sum_tc_t\psi_t$. The verifier moves the
response along its mean score direction and the teacher along the
covariance between local advantages and scores. The two label components
are orthogonal, but these two parameter directions need not be, so the
mean identity fixes the teacher's share of the label and not of the update.

\paragraph{Proof of Proposition~\ref{prop:response}.}
Let $\mathcal F_t$ contain the prompt and actions preceding position $t$,
and let $\psi_t=\nabla_\theta\log p_\theta(A_t\mid\mathcal F_t)$ at the
rollout parameters. The score identity and the student value baseline
give $\mathbb E[\psi_t\mid\mathcal F_t]=0$ and
$\mathbb E[A_t^{d}\mid\mathcal F_t]=0$. Expanding the centered estimator,
\begin{equation}
 \sum_t c_t\delta_t\psi_t
 =\lambda\sum_t c_t(1-c_t)A_t^{d}\psi_t
  -\lambda\sum_{t}\sum_{u\ne t}c_tc_uA_u^{d}\psi_t.
 \label{eq:app-cross-expansion}
\end{equation}
For $u<t$, $A_u^{d}$ is $\mathcal F_t$-measurable, so its product with
$\psi_t$ has zero expectation. For $u>t$, $\psi_t$ is
$\mathcal F_u$-measurable, and conditioning on $\mathcal F_u$
annihilates $A_u^{d}$. Thus
\begin{equation}
 \mathbb E\!\left[\sum_t c_t\delta_t\psi_t\right]
 =\lambda\sum_t c_t(1-c_t)\mathbb E[A_t^{d}\psi_t],
 \label{eq:app-fixed-response}
\end{equation}
which is Equation~\ref{eq:response-expectation}. Each conditional term is
the negative gradient of the local squared distance through that prefix's
logits with the prefix held fixed; the identity does not differentiate
the distribution of visited prefixes. Without the student value baseline
the case $u>t$ fails, because a later raw residual has conditional mean
$\langle p_u,d_u\rangle$, and centering then couples position $t$ to the
disagreement levels of prefixes not yet visited.

The same proof applies to the student-centered log label
$\ell_t^p=\ell_t(A_t)-\langle p_t,\ell_t\rangle$,
\begin{equation}
 \E\left[\sum_t c_t\beta(\ell_t^p-\bar\ell^p)\psi_t\right]
 =\beta\sum_t c_t(1-c_t)\E[\ell_t^p\psi_t],
 \label{eq:app-centered-log-response}
\end{equation}
so response centering changes the expected update even though the
student value baseline alone does not. For the raw $\ell_t$ the cross
terms can survive because $\E[\ell_t\mid\mathcal F_t]=-\KL(p_t\Vert q_t)$.
For example, let $p_1=q_1=(1/2,1/2)$ and $p_2=(1/2,1/2)$ after either
first action, with $q_2=(.9,.1)$ after the first action, $q_2=p_2$ after
the second, and $c_1=c_2=1/2$. The independent first-action logit then
receives the expected update $-\beta\log(.6)/16\ne0$ from response-centered
raw log guidance although its own raw log label is zero, whereas the
centered log label gives it zero expected update.

\subsection{The teacher-peak branch}
\label{app:reference-fork}

Let $\mathcal T_i$ contain the positions of response $i$; every one of them
receives the guidance term. At each visited prefix, both routes use
\[
 h_{it}(v)=d_{it}(v)-\langle p_{it},d_{it}\rangle=A_{it}^d(v),
 \qquad d_{it}=q_{it}-p_{it}.
\]
The sampled label remains
$A_{it}=A_i^V+\lambda(h_{it}(a_{it})-\bar h_i)$, where
$\bar h_i=\sum_{t\in\mathcal T_i}c_{it}h_{it}(a_{it})$ uses exactly the
sampled \RA{} weights. With $v_{it}^\star=\arg\max_v q_{it}(v)$,
$s_{it}=\arg\max_v p_{it}(v)$, and
$\mathcal B_i=\{t\in\mathcal T_i:v_{it}^\star\ne s_{it},\ v_{it}^\star\ne a_{it}\}$,
the branch adds the teacher's peak as a pseudo-sample carrying the label a
sampled token at that position would receive, with the verifier part
clipped at zero:
\begin{align}
 \mathcal J_{\rm branch}(\theta)
 &=\frac1Z\sum_i\sum_{t\in\mathcal T_i}A_{it}\log\pi_\theta(a_{it}\mid\xi_{it})\nonumber\\
 &\quad+\frac1Z\sum_i\sum_{t\in\mathcal B_i}
       \big([A_i^V]_++\lambda(h_{it}(v_{it}^\star)-\bar h_i)\big)\log\pi_\theta(v_{it}^\star\mid\xi_{it}),
 \label{eq:app-fork-objective}
\end{align}
with $[x]_+=\max(x,0)$ and $Z=\sum_i|\mathcal T_i|$. The sampled route is
computed first, and the added entries enter neither $\bar h_i$ nor the
normalizer, so the sampled labels are those of \RA{} and the sampled mean
is unchanged for every response,
\begin{equation}
 \sum_{t\in\mathcal T_i}c_{it}A_{it}=A_i^V,
 \qquad
 \sum_{t\in\mathcal T_i}c_{it}\lambda
       (h_{it}(a_{it})-\bar h_i)=0.
 \label{eq:app-fork-sampled-mean}
\end{equation}
The added labels have no zero-mean constraint. They sum over response $i$
to $|\mathcal B_i|[A_i^V]_++\lambda\sum_{t\in\mathcal B_i}(h_{it}(v_{it}^\star)-\bar h_i)$,
whose verifier part is never negative, so the branch raises the total
label of every response with a positive verifier advantage and leaves the
verifier part of the others unchanged. The teacher part pushes toward the
peak apart from the shift by $\bar h_i$, as the next proposition shows.

\begin{proposition}[The teacher's peak under the residual advantage]
\label{prop:branch}
Fix a prefix with positive $p,q$ on a common coordinate support and a
teacher peak $b\in\arg\max_vq(v)$ over that support. Then
$\norm{p}_2^2-p(b)\le A^{d}(b)\le2(1-p(b))^2$. Hence
$A^{d}(b)\ge-p(b)(1-p(b))\ge-\tfrac14$, the constant $-\tfrac14$ is sharp,
and $A^{d}(b)\ge0$ whenever $p(b)\le\norm{p}_2^2$.
\end{proposition}

\paragraph{Proof.}
Write $A^{d}(b)=\langle e_b-p,q-p\rangle
=[q(b)-\langle p,q\rangle]-[p(b)-\norm{p}_2^2]$. Since $q(b)\ge q(u)$ for
every $u$, $\langle p,q\rangle\le q(b)$, which gives the lower bound; with
$\norm{p}_2^2\ge p(b)^2$ it yields $A^{d}(b)\ge-p(b)(1-p(b))\ge-\tfrac14$,
and $p(b)\le\norm{p}_2^2$ makes the lower bound nonnegative. For the upper
bound, $q(b)-\langle p,q\rangle=\sum_{u\ne b}p(u)(q(b)-q(u))\le(1-p(b))q(b)$
and $\norm{p}_2^2-p(b)\le p(b)^2+(1-p(b))^2-p(b)=(1-p(b))(1-2p(b))$, so
$A^{d}(b)\le(1-p(b))(q(b)+1-2p(b))\le2(1-p(b))^2$. For sharpness put
$p(b)=\tfrac12$, spread the remaining mass uniformly over $n$ other
tokens, and let $q$ be uniform over these $n+1$ tokens; then
$q(b)=\langle p,q\rangle$ and $A^{d}(b)=-\tfrac14+\tfrac1{4n}$.

The peak's local advantage is therefore negative only when the student
already gives the peak more probability than its collision probability
$\norm{p}_2^2=\E_{u\sim p}[p(u)]$, and never by more than $\tfrac14$. With
$\bar h_i\in[-\tfrac12,2]$, a convex combination of local advantages
(Proposition~\ref{prop:advantage}(iii)), each added label lies in
$\big[[A_i^V]_+-\tfrac94\lambda,\ [A_i^V]_++\tfrac52\lambda\big]$.

\subsection{An adapted teacher in the residual advantage}
\label{app:adaptive-teacher}

At iteration $k$, \CoRA{} scores the current student batch with the
pre-update teacher $q_{\phi_k}$ and holds these scores fixed during the
student update. Its residual is $d_t^{(k)}=q_{\phi_k,t}-p_{k,t}$, its
local advantage is
$A_t^{d,(k)}=\langle e_{a_t}-p_{k,t},d_t^{(k)}\rangle$, and with the same
normalized response weights $c_t$ its guidance term is
$\delta_t^{(k)}=\lambda(A_t^{d,(k)}-\sum_uc_uA_u^{d,(k)})$. The
per-response results of Appendix~\ref{app:mass-bounds} apply directly:
\begin{align}
 \sum_tc_t\bigl(A^V+\delta_t^{(k)}\bigr)&=A^V,\nonumber\\
 |\delta_t^{(k)}|&\le\tfrac52\lambda(1-c_t),\nonumber\\
 |\delta_t^{(k)}|&\le2\sqrt2\lambda(1-c_t)
       \max_u\lVert q_{\phi_k,u}-p_{k,u}\rVert_2.
 \label{eq:app-adaptive-bounds}
\end{align}
Teacher adaptation thus preserves the mean identity, bounded guidance,
and uniform vanishing with distributional agreement at each scoring
step, and the fixed-prefix expected-gradient identity holds with
$q=q_{\phi_k}$.

The teacher LoRA is trained on the same student-generated trajectories
with their verifier advantages (Appendix~\ref{app:adapter}). At the start
of its clipped update the snapshot ratio is one, so
\begin{equation}
 \left.\nabla_\phi\mathcal J_T\right|_{\phi_k}
 =\frac1Z\sum_i A_i^V\sum_{t\in\mathcal T_i}
   \left.\nabla_\phi\log q_{\phi,it}(a_{it})\right|_{\phi_k},
 \qquad Z=\sum_i|\mathcal T_i|,
 \label{eq:app-adapter-gradient}
\end{equation}
with $A_i^V$ held fixed. At a fixed prefix a sampled term has logit
gradient $A^V(e_a-q_{\phi_k})$, and if
$\widehat A(v)=\E[A^V\mid\xi,a=v]$ under the student's continuations, its
student-policy expectation is
$p_k\odot\widehat A-q_{\phi_k}\langle p_k,\widehat A\rangle$. The
trajectories come from $p_k$ and no importance correction to $q_{\phi_k}$
is applied, so this is a student-on-policy gradient for the teacher.
Shared parameters, the LoRA parameterization, and clipping determine the
actual teacher update.

The updated $q_{\phi_{k+1}}$ enters the next iteration through the same
residual. At a fixed student distribution, replacing a teacher by
$q+\Delta q$ changes the local advantage by $\langle e_a-p,\Delta q\rangle$,
as in Equation~\ref{eq:app-mass-sensitivity}, so teacher adaptation
changes the existing guidance without adding another student signal. Both
distributions and the visited prefixes move across iterations, and the
identities above do not by themselves give monotone distance reduction,
policy improvement, or declining guidance during training.

\FloatBarrier
\section{Experimental Protocol and Comparison Definitions}
\label{app:protocol}

\subsection{Training and evaluation configuration}

\begingroup
\small
\setlength{\LTcapwidth}{\linewidth}
\renewcommand{\tablename}{\normalsize Table}
\begin{longtable}{@{}p{0.48\linewidth}p{0.46\linewidth}@{}}
\caption{\textbf{Training and evaluation configuration.} Lengths count response tokens. The correct-only linear length penalty applies to verifier runs; all runs share the response cap. Evaluation uses unshaped answer correctness.}
\label{tab:setup}\\
\toprule
\rowcolor{tableheadcol}
\multicolumn{1}{@{}>{\columncolor{tableheadcol}[0pt][\tabcolsep]}p{0.48\linewidth}}{Setting} & \multicolumn{1}{>{\columncolor{tableheadcol}[\tabcolsep][0pt]}p{0.46\linewidth}@{}}{Configuration}\\
\midrule
\endfirsthead
\multicolumn{2}{@{}l}{\normalsize Table~\thetable{} (continued)}\\[3pt]
\toprule
\rowcolor{tableheadcol}
\multicolumn{1}{@{}>{\columncolor{tableheadcol}[0pt][\tabcolsep]}p{0.48\linewidth}}{Setting} & \multicolumn{1}{>{\columncolor{tableheadcol}[\tabcolsep][0pt]}p{0.46\linewidth}@{}}{Configuration}\\
\midrule
\endhead
\midrule
\multicolumn{2}{r@{}}{\footnotesize Continued on next page}\\
\endfoot
\bottomrule
\endlastfoot
Students & Qwen3-1.7B-Base and Qwen3-4B-Base\\
Teacher backbone / inference mode & Qwen3-8B (post-trained) / non-thinking\\
Training data / prompt count & DAPO-Math-17K / 17,398\\
Validation prompts drawn from the training file / problems used & 256 / 64\\
Validation samples per problem / temperature & 8 / 1.0\\
Training epochs / prompts per rollout batch & 2 / 128\\
Incomplete final batch / rollout batches per epoch & Dropped / 135\\
Responses per prompt & 8 (GRPO runs) / 1 (OPD and REINFORCE++ runs)\\
Training temperature / response cap & 1.0 / 12,000\\
Verifier full-reward threshold / penalty window & 8,000 / 4,000\\
\RA{} weight $\lambda$ (reference configuration) & 10\\
\RA{} weight sweep (1.7B, REINFORCE++) & $5,\ 10,\ 20$\\
\CoRA{} residual weight $\lambda$ & 10\\
No-baseline / raw residual / no-centering weights (1.7B) & $\gamma_x=10$ / $\gamma_{\rm raw}=10$ / $\lambda=10$\\
Residual-control and direct $L_2$ weight sweeps (1.7B) & $5,\ 10,\ 20$ each\\
Raw log $\beta$ and centered log $\gamma_{\log}$ sweeps (1.7B) & $0.05,\ 0.2,\ 0.6,\ 1.2,\ 2$\\
Clipped centered log ratio: clip $c$ / weight sweep & $5$ / $0.2,\ 0.6,\ 1.2$\\
Reported $\beta$ / $\gamma_{\log}$ / clipped $\gamma_{\log}$ / $\alpha$ & $0.2$ / $0.6$ / $0.6$ / $10$\\
Weight selection & Validation split (Appendix~\ref{app:validation})\\
Teacher-only OPD objective / weight & Full-vocabulary reverse KL / $1$\\
Reverse-KL loss on REINFORCE++: sweep / reported (Appendix~\ref{app:sequential}) & $0.05,\ 0.2,\ 0.6$ / $0.2$\\
Sampled OPD (Appendix~\ref{app:sampled-opd}) label / weight & $\log[q_t(a_t)/p_t(a_t)]$ / $1$\\
Evaluation temperature / response cap & 0.8 / 12,000\\
Evaluation completions per problem & 8\\
MATH-500 / AIME24 / AIME25 problems & 500 / 30 / 30\\
MATH-500 Avg@8 / Pass@8 increments (pp) & 0.025 / 0.2\\
AIME Avg@8 / Pass@8 increments (pp) & $100/240$ / $100/30$\\
\end{longtable}
\endgroup

\begin{table}[!htbp]
\caption{\textbf{Macro differences among runs.} Differences of three-seed macro means from Tables~\ref{tab:main}, \ref{tab:benchmark-controls}, and~\ref{tab:branch}, in percentage points. Each mean and difference is computed from raw counts before rounding; subtracting the displayed macro columns can differ by $0.01$.}
\label{tab:gains}
\centering
\begingroup
\small
\setlength{\tabcolsep}{5pt}
\begin{tabular}{@{}llcc@{}}
\toprule
\tablerowshade{tableheadcol}
Student & Comparison (macro, pp) & $\Delta$ Avg@8 & $\Delta$ Pass@8\\
\midrule
1.7B & \RA{}+REINFORCE++ $-$ REINFORCE++ & +3.58 & +5.76\\
1.7B & \CoRA{}+REINFORCE++ $-$ \RA{}+REINFORCE++ & +1.45 & +1.18\\
1.7B & \RA{}+REINFORCE++ $-$ OPD & +1.80 & +3.90\\
1.7B & \CoRA{}+REINFORCE++ $-$ OPD & +3.25 & +5.08\\
1.7B & \RA{}+REINFORCE++ with branch $-$ \RA{}+REINFORCE++ & -2.07 & -3.41\\
1.7B & \RA{}+GRPO $-$ GRPO & +2.37 & +4.35\\
1.7B & \CoRA{}+GRPO $-$ \RA{}+GRPO & +1.23 & -0.46\\
1.7B & \RA{}+GRPO $-$ OPD & +2.40 & +3.45\\
1.7B & \CoRA{}+GRPO $-$ OPD & +3.63 & +2.99\\
\midrule
4B & \RA{}+REINFORCE++ $-$ REINFORCE++ & +3.26 & +6.33\\
4B & \CoRA{}+REINFORCE++ $-$ \RA{}+REINFORCE++ & +1.27 & -2.24\\
4B & \RA{}+REINFORCE++ $-$ OPD & +2.51 & +3.84\\
4B & \CoRA{}+REINFORCE++ $-$ OPD & +3.78 & +1.60\\
4B & \RA{}+GRPO $-$ GRPO & +1.74 & +3.89\\
4B & \CoRA{}+GRPO $-$ \RA{}+GRPO & +1.03 & +1.09\\
4B & \RA{}+GRPO $-$ OPD & +2.96 & +2.73\\
4B & \CoRA{}+GRPO $-$ OPD & +3.99 & +3.82\\
\bottomrule
\end{tabular}
\endgroup

\end{table}

\subsection{Probability and actor conventions}

\begin{table}[!htbp]
\caption{\textbf{Probability, actor, and reporting settings.} These settings supplement Table~\ref{tab:setup}.}
\label{tab:protocol-actor}
\centering\small
\begin{tabular}{@{}p{0.48\linewidth}p{0.46\linewidth}@{}}
\toprule
\tablerowshade{tableheadcol}
Setting & Configuration\\
\midrule
Aligned model-output width / tokenizer nominal size & $151936$ / $151669$\\
Probability, baseline, and label accumulation & At least float32\\
Actor reduction / additional label whitening & Token mean / none\\
PPO clipping width / epochs per rollout batch & $0.2$ / $1$\\
REINFORCE++ optimizer steps per rollout batch & $1$\\
Student minibatch size & $256$ responses\\
Student optimizer / learning rate & AdamW / $10^{-6}$\\
Student optimizer betas & $(0.9,0.999)$\\
Student weight decay / gradient norm limit & $0$ / $1$\\
Score display precision & $2$ decimal places\\
\bottomrule
\end{tabular}
\end{table}

Student and teacher are evaluated at the same visited prefix on the same aligned output coordinates. The aligned width is the width of the probability tensor, which exceeds the tokenizer's nominal size (Table~\ref{tab:protocol-actor}); the local advantage sums over all aligned coordinates.

Rollouts use temperature one and full probability support. Labels are computed at the rollout distribution and held fixed through the student PPO update. The actor loss covers every token of each response, and the guidance term is applied at every one of these positions, with no token excluded; the centering weights follow the actor's relative token measure.

Probabilities, the student value baseline, and actor labels are accumulated at the precision in Table~\ref{tab:protocol-actor}, and no further whitening is applied to $A^V+\delta$ after the base estimator has normalized its advantages. The sampled routes use token-mean reduction. With $Z$ the batch's total token count, response $i$'s sampled teacher gradient is $(M_i/Z)\sum_t c_{it}\delta_{it}\psi_{it}$, so the normalized response gradient analyzed in the text carries the trajectory-dependent batch weight $M_i/Z$, which lies outside the fixed-weight expectation of Proposition~\ref{prop:response}. The teacher-peak branch adds its term with the same $Z$. The PPO update and student optimizer follow Table~\ref{tab:protocol-actor}.

\subsection{Training reward and advantage construction}
\label{app:reward}

\begin{table}[!htbp]
\caption{\textbf{Verifier advantage normalization settings.} Normalization follows reward shaping, with no reward-side KL.}
\label{tab:protocol-normalization}
\centering\small
\begin{tabular}{@{}p{0.48\linewidth}p{0.46\linewidth}@{}}
\toprule
\tablerowshade{tableheadcol}
Setting & Configuration\\
\midrule
GRPO group size / standard deviation & $8$ responses / sample standard deviation\\
GRPO denominator stabilizer & $10^{-6}$\\
REINFORCE++ discount / variance floor & $1$ / $10^{-8}$\\
REINFORCE++ normalization measure & Response tokens\\
\bottomrule
\end{tabular}
\end{table}

Training uses the batch, sampling, and length settings of Table~\ref{tab:setup}. For runs with verifier feedback, let $y_i\in\{0,1\}$ be answer correctness and $L_i$ the number of response tokens. The reward passed to the base estimator is
\begin{equation}
 R_i=y_i\,\operatorname{clip}\!\left(\frac{12000-L_i}{4000},0,1\right).
 \label{eq:shaped-reward}
\end{equation}
A correct response receives full reward up to the threshold in Table~\ref{tab:setup}, then loses reward linearly until the cap; incorrect responses receive zero. Shaping occurs before advantage normalization, so a zero raw reward can yield a negative response advantage. Teacher-only OPD uses neither this reward nor its soft length penalty and keeps the same 12,000-token hard cap.

GRPO normalizes the shaped returns within each prompt group by their mean and sample standard deviation, with the group size and denominator stabilizer of Table~\ref{tab:protocol-normalization}. Groups with identical returns stay in the batch with zero advantages, and a group of correct responses can still receive different advantages through the length shaping.

REINFORCE++ uses terminal returns and the discount and token measure in Table~\ref{tab:protocol-normalization}. With no reward-side KL, the response advantage is
\begin{align}
 \bar R_{\rm tok}&=\frac{\sum_i |\mathcal T_i| R_i}{\sum_i |\mathcal T_i|},&
 v_{\rm tok}&=\frac{\sum_i|\mathcal T_i|(R_i-\bar R_{\rm tok})^2}{\sum_i|\mathcal T_i|},\nonumber\\
 A_i^{V,\mathrm R}&=\frac{R_i-\bar R_{\rm tok}}
 {\sqrt{\max\{v_{\rm tok},10^{-8}\}}}.
 \label{eq:rpp-return}
\end{align}
Here $|\mathcal T_i|$ counts the tokens of response $i$, so the normalization is token-weighted when response lengths differ, and its variance is computed from the shaped returns.

\subsection{Evaluation measures}

Evaluation uses the checkpoint at the end of the second epoch. MATH-500 has 500 problems and each AIME set 30. For every problem, eight completions are sampled at temperature $0.8$ with the 12,000-token cap. With $y_{jk}$ the correctness of completion $k$ of problem $j$, each seed's scores are
\begin{align}
 \mathrm{Avg@8}&=\frac{100}{8N}\sum_{j=1}^{N}\sum_{k=1}^{8}y_{jk},\nonumber\\
 \mathrm{Pass@8}&=\frac{100}{N}\sum_{j=1}^{N}
       \mathbf1\!\left[\sum_{k=1}^{8}y_{jk}>0\right].
 \label{eq:eval-eight}
\end{align}
Both metrics use unshaped correctness. Avg@8 is completion accuracy, and Pass@8 is the fraction of problems solved at least once in the eight sampled completions, with no extrapolation to a larger budget. Macro scores weight the three benchmark percentages equally within each seed and metric, and the main tables report the mean over three seeds. Appendix~\ref{app:seeds} gives per-seed macro scores and their sample SD, which uses denominator $3-1$ and is neither a standard error nor a confidence interval.

Single-seed score increments follow from these counts (Table~\ref{tab:setup}), and three-seed averages permit finer values. Means and SD are computed from per-seed scores and shown to two decimals, so recomputing them from displayed values can differ in the last decimal.

\subsection{Validation split and weight selection}
\label{app:validation}

Before training, 256 prompts are drawn at random from the DAPO-Math-17K file with a fixed seed, and 64 of them form the validation set. Validation samples eight responses per problem at the training temperature of $1.0$ and scores mean accuracy over the samples. The guidance weight $\lambda$ and the control weights of Table~\ref{tab:setup} were chosen on this split, and the evaluation benchmarks of Section~\ref{sec:experiments} played no part in any selection. The validation prompts are not removed from the training file, so they select weights among training-distribution prompts and never among evaluation problems. The same 256 prompts supply the offline responses of Appendix~\ref{app:buckets}. The sweeps of Tables~\ref{tab:sweep-mass} and~\ref{tab:sweep-log} evaluate every swept weight on the benchmarks after the fact, so they show how the selected weights compare with their neighbors and not how they were chosen.

\subsection{Guidance construction}
\label{app:guidance}

\begin{table}[!htbp]
\caption{\textbf{Residual guidance with a fixed or adapted teacher.} Each method uses one teacher distribution, the student value baseline, and response centering. The weight is $\lambda=10$ in both reference configurations.}
\label{tab:protocol-guidance}
\centering\small
\begin{tabular}{@{}p{0.32\linewidth}p{0.29\linewidth}p{0.31\linewidth}@{}}
\toprule
\tablerowshade{tableheadcol}
Operation at iteration $k$ & \RA{} & \CoRA{}\\
\midrule
Teacher distribution $q_t$ & Fixed $q_{0,t}$ & Current $q_{\phi_k,t}$\\
Local reward field & $d_t=q_{0,t}-p_{k,t}$ & $d_t=q_{\phi_k,t}-p_{k,t}$\\
Student value baseline & $\langle p_{k,t},d_t\rangle$ & $\langle p_{k,t},d_t\rangle$\\
Sampled local advantage & \multicolumn{2}{c}{$A_t^d=d_t(a_t)-\langle p_{k,t},d_t\rangle$}\\
Centered guidance term & $\lambda(A_t^d-\overline A^d)$ & $\lambda(A_t^d-\overline A^d)$\\
\bottomrule
\end{tabular}
\end{table}

Table~\ref{tab:protocol-guidance} compares the teacher distributions of \RA{} and \CoRA{}. \RA{} keeps $q_0$ fixed; \CoRA{} uses the current adapter snapshot $q_{\phi_k}$ in the same residual, with the same weight $\lambda=10$ and centering weights. The combined label is fixed during each scored batch's update, and the updated adapter is used in the next iteration. Teacher-only OPD minimizes the exact per-token reverse KL divergence $\KL(p_t\Vert q_t)$, summed over the full aligned vocabulary at every visited prefix of the student's responses, with the teacher held fixed and gradients through all of the student's probabilities. It uses no verifier feedback or length shaping. The sampled variant of Appendix~\ref{app:sampled-opd} replaces this loss by the policy-gradient label $\log[q_t(a_t)/p_t(a_t)]$ at the sampled token \citep{lu2025opd} and keeps every other setting.

\paragraph{Control weights.}
The controls are not amplitude-calibrated. The residual controls (without the student value baseline, without response centering, and the raw residual) use the reference weight $10$, and each of them and the direct-$L_2$ loss is also run at $5$ and $20$, the weights of the sweep of full \RA{}. The log-ratio controls are swept over the weights in Table~\ref{tab:setup}, and Table~\ref{tab:alternatives} reports each at the weight selected on the validation split (Appendix~\ref{app:validation}); Tables~\ref{tab:sweep-mass} and~\ref{tab:sweep-log} list every swept weight on the benchmarks. No weight, including $\lambda$ of full \RA{}, was selected on the evaluation benchmarks. Each weight multiplies its signal directly and stays fixed during training.

The raw log ratio adds $\beta\ell_{it}$, with $\ell_{it}=\log[q_{it}(a_{it})/p_{it}(a_{it})]$, to the REINFORCE++ verifier advantage without either operation. The raw residual adds $\gamma_{\rm raw}d_{it}(a_{it})$ without either operation, and the uncentered control retains $A_t^d$ (Equation~\ref{eq:centering-controls}). Direct $L_2$ is a loss control; at $\alpha=\lambda$ and a fixed prefix its gradient at the rollout policy matches the expected residual-advantage gradient before response centering (Appendix~\ref{app:local-objective}), while response centering and batch weights lie outside this identity.

\paragraph{Teacher-peak branch.}
At each visited prefix let $v_t^\star=\arg\max_v q_t(v)$ and $s_t=\arg\max_v p_t(v)$, with both maxima taken over the aligned support. The branch adds $v_t^\star$ exactly when $v_t^\star\ne s_t$ and $v_t^\star\ne a_t$; the sampled token may equal $s_t$. With \RA{}, the sampled label remains $A_t=A^V+\lambda(h_t(a_t)-\bar h)$, with $h_t(v)=d_t(v)-\langle p_t,d_t\rangle$ and $\bar h$ computed first from the sampled entries, and the added token receives $[A^V]_++\lambda(h_t(v_t^\star)-\bar h)$ with $[x]_+=\max(x,0)$, the label a sampled token at that position would receive with its verifier part clipped at zero. Both routes use the same residual $d_t=q_t-p_t$ and weight $\lambda$, and $Z$ counts only sampled tokens, so the added entries neither dilute the sampled route nor introduce a separate weight. Equation~\ref{eq:app-fork-objective} gives the batch objective. With REINFORCE++ the sampled label is $A^V$, and the added token receives the same $[A^V]_++\lambda(h_t(v_t^\star)-\bar h)$, with $\bar h$ computed from the sampled tokens, so $\lambda$ multiplies only the branch's teacher part. Sampled OPD has no verifier term; there the added token receives the sampled-OPD label $\log q_t(v_t^\star)-\log p_t(v_t^\star)$, so each triggered position is supervised at two tokens, a top-$k$ variant of the sampled estimate.

The REINFORCE++ runs take one student optimizer step per rollout batch. At the rollout policy the ratio is one and clipping is inactive, so the log-probability objective of Equation~\ref{eq:app-fork-objective} and the ratio surrogate have equal gradients and give the same single update; they differ once the parameters move.

\subsection{Label and direct-loss controls}
\label{app:controls}

Every control below keeps the REINFORCE++ verifier advantage and changes only how the teacher enters, through the actor label or, for the direct loss, through the objective; Table~\ref{tab:setup} lists the weights.

\paragraph{Raw log ratio.}
To test the role of \RA{}'s advantage construction, we replace its guidance term with the raw sampled log ratio while retaining the REINFORCE++ sequence advantage:
\[
 A_t^{\mathrm{add}}=A^V+\beta\,\ell_t,\qquad
 \ell_t=\log\frac{q_t(a_t)}{p_t(a_t)}.
\]
The weight $\beta$ is swept over five values and held fixed within each run of the 1.7B student (Table~\ref{tab:sweep-log}; Appendix~\ref{app:guidance}).

Adding the raw log ratio improves both macro scores over plain REINFORCE++ at $\beta\le0.6$ and lowers both at $\beta\ge1.2$, whereas the complete \RA{} guidance term improves both at every tested weight (Tables~\ref{tab:sweep-mass} and~\ref{tab:sweep-log}). At its best weight, $\beta=0.2$, it remains $1.97$ Avg@8 and $2.70$ Pass@8 points below full \RA{}. With verifier feedback present in both guided runs, this comparison favors the complete \RA{} construction over the raw log ratio. The following controls examine its components and a direct-loss alternative.

\paragraph{Alternative advantage constructions.}
The centered log ratio uses the conventional action-minus-value construction with the log-ratio reward, followed by the same response centering. The control without the student value baseline retains the sampled residual and response centering but drops the expectation over the student's alternatives. Writing $L_t(v)=\log[q_t(v)/p_t(v)]$ and $P_c=I-\mathbf1c^\top$ for response centering, their actor labels are
\begin{equation}
\begin{aligned}
 \mathbf A^{\rm clog}
 &=A^V\mathbf1+\gamma_{\log}P_c
   [L_t(a_t)-\langle p_t,L_t\rangle]_t,\\
 \mathbf A^{\rm noV}
 &=A^V\mathbf1+\gamma_xP_c[d_t(a_t)]_t.
\end{aligned}
\label{eq:component-labels}
\end{equation}
The clipped centered log ratio replaces $L_t$ in $\mathbf A^{\rm clog}$ by $\operatorname{clip}(L_t,-c,c)$ with $c=5$. These guidance terms are held fixed, as in \RA{}, and their weights stay fixed during training: $\gamma_x=10$, and $\gamma_{\log}$ is swept for both log-ratio controls (Appendix~\ref{app:guidance}). These controls ask how the local reward and its student value baseline contribute once teacher guidance is centered across a response.

\paragraph{Raw residual and removal of response centering.}
Two controls on the 1.7B student with REINFORCE++ use the same probability residual as \RA{}:
\begin{equation}
 A_t^{\rm raw}=A^V+\gamma_{\rm raw}d_t(a_t),\qquad
 A_t^{\rm noRC}=A^V+\lambda A_t^d.
 \label{eq:centering-controls}
\end{equation}
The raw residual removes both the student value baseline and response centering. Its weight $\gamma_{\rm raw}=10$ equals the reference $\lambda$ and multiplies $d_t(a_t)$ directly. The second control retains the student value baseline and removes only response centering, at the unchanged weight $\lambda=10$. All three residual controls are also run at weights $5$ and $20$ (Table~\ref{tab:sweep-mass}). The comparison fixes the weight while changing the centering operation rather than matching the resulting signal amplitude. All teacher terms are held fixed and use the same actor reduction as \RA{}. The raw residual differs from the no-student-value-baseline control in Equation~\ref{eq:component-labels}, which retains response centering.

Without response centering, the weighted response mean of the actor label is $A^V+\gamma_{\rm raw}\sum_tc_td_t(a_t)$ for the raw residual and $A^V+\lambda\overline A^d$ for the uncentered control, whereas \RA{} preserves $A^V$ exactly. The student value baseline compares actions at a prefix; response centering removes the realized teacher mean across positions. Table~\ref{tab:weight-sweep} summarizes these two controls across the weight sweep, and Table~\ref{tab:seed-reporting} gives the individual runs.

\paragraph{The same local geometry through a direct loss.}
The direct-$L_2$ control adds a differentiable penalty to the verifier objective:
\begin{equation}
\begin{aligned}
 \mathcal J_{\rm direct}(\theta)
 &=\mathcal J_{\rm PPO}(A^V)
 -\frac{\alpha}{2Z}\sum_i\sum_{t\in\mathcal T_i^{\rm teach}}\sum_{v\in\mathcal V}
 \bigl(\pi_\theta(v\mid\xi_{it})-q_{it}(v)\bigr)^2 .
\end{aligned}
\label{eq:component-direct-l2}
\end{equation}
Here $Z$ is the token-reduction denominator. The teacher distributions and visited prefixes are fixed, and gradients pass through the current student probabilities. The vocabulary sum is not divided by $K$. We run $\alpha\in\{5,10,20\}$, the weights of the $\lambda$ sweep, without amplitude calibration; $\alpha=10$ gives the highest macro Avg@8 and is reported in Table~\ref{tab:alternatives}. At a fixed prefix, the logit gradient at the rollout policy is $\alpha J_pd$, matching the expected residual-advantage gradient at this common weight before response centering (Proposition~\ref{prop:advantage}(ii); Appendix~\ref{app:local-objective}). Response centering, trajectory-dependent reduction weights, and sampled gradient noise still distinguish the two complete training updates. The REINFORCE++ comparison uses one optimizer step per rollout batch, for which surrogate clipping is inactive at the rollout policy.

\subsection{Adapter objective and update order}
\label{app:adapter}

\begin{table}[!htbp]
\caption{\textbf{Adapter configuration and update schedule.}}
\label{tab:protocol-adapter}
\centering\small
\begin{tabular}{@{}p{0.48\linewidth}p{0.46\linewidth}@{}}
\toprule
\tablerowshade{tableheadcol}
Setting & Configuration\\
\midrule
Adapter rank / scaling & $16$ / $32$\\
Dropout / initial effective update & $0$ / $0$\\
Target matrices & Attention and MLP projections\\
Separate optimizer / learning rate & AdamW / $10^{-5}$\\
Optimizer betas & $(0.9,0.999)$\\
Weight decay / gradient norm limit & $0$ / $1$\\
Update interval & Every student rollout iteration\\
Update data / epochs & Current scored student batch / $1$\\
Minibatch size & Up to $256$ responses (capped by batch size)\\
Clipping width / reduction & $\epsilon_{\rm clip}=0.2$ / batch token mean\\
\bottomrule
\end{tabular}
\end{table}

The teacher backbone $q_0$ is frozen. The adapter and its separate optimizer follow Table~\ref{tab:protocol-adapter}; the adapter targets the attention and MLP projections, and its effective update starts at zero.

At iteration $k$, the student $p_k$ generates a batch and the current adapter snapshot $q_{\phi_k}$ scores it before either model is updated. The verifier advantage $A_i^{V,\mathcal B}$ comes from the run's base estimator $\mathcal B\in\{\mathrm{GRPO},\mathrm{REINFORCE++}\}$, and the adapter applies that estimator's clipped surrogate to the same scored batch:
\begin{align*}
\mathcal J_T(\phi)&=\frac1Z\sum_i\sum_{t\in\mathcal T_i}
 \min\!\big(\rho^T_{it}A_i^{V,\mathcal B},\ \operatorname{clip}(\rho^T_{it},1-\epsilon_{\rm clip},1+\epsilon_{\rm clip})A_i^{V,\mathcal B}\big),\\
\rho^T_{it}&=\frac{q_{\phi,it}(a_{it})}{q_{\phi_k,it}(a_{it})},
\qquad Z=\sum_i|\mathcal T_i|,
\end{align*}
with clipping width $\epsilon_{\rm clip}$ from Table~\ref{tab:protocol-adapter}, the student's response tokens and batch token-mean reduction, and $q_{\phi_k}$ the adapter snapshot that scored this iteration's responses. The snapshot ratio limits how far $q_\phi$ moves during its update and does not reweight the student's samples toward the teacher's distribution. At the start of the update every ratio is one and the gradient is Equation~\ref{eq:adapter-gradient}. The adapter receives the verifier advantage without the guidance term, optimizer and adapter states persist across iterations, and no adapter KL regularizer or label clipping is used.

Every iteration follows the same order. The student generates a batch; the current snapshots and the verifier score it; the labels and rollout probabilities are frozen; the student and the teacher LoRA are updated on that batch, in either order, since their targets are fixed and their parameters are separate. The updated $q_{\phi_{k+1}}$ scores the next iteration, whose residual is $q_{\phi_{k+1}}-p_{k+1}$ with the same weight, student value baseline, and response centering (Table~\ref{tab:protocol-guidance}). The backbone $q_0$ stays frozen and the adapter state is carried forward.

\subsection{Computation and evidence scope}

All runs train for two epochs on the 17,398-prompt DAPO-Math-17K file. With 128 prompts per rollout batch and the final incomplete batch dropped, each epoch processes 135 batches, or 17,280 prompt occurrences, so the two epochs comprise 270 rollout batches and 34,560 prompt occurrences. The scheduled generation budgets are therefore 276,480 responses for GRPO runs and 34,560 for OPD and REINFORCE++ runs, cumulative counts of student generations on which Appendix~\ref{app:dynamics} reports training dynamics. \RA{} adds a teacher forward pass and the full-vocabulary inner product; \CoRA{} scores with the current adapted teacher and additionally trains its LoRA on every rollout batch (Table~\ref{tab:protocol-adapter}). The method needs neither counterfactual continuations nor a separately fitted value head. Wall-clock and peak-memory measurements are not reported.

\FloatBarrier
\section{Training dynamics}
\label{app:dynamics}

\begin{figure}[!htbp]
\centering
\includegraphics[width=\linewidth]{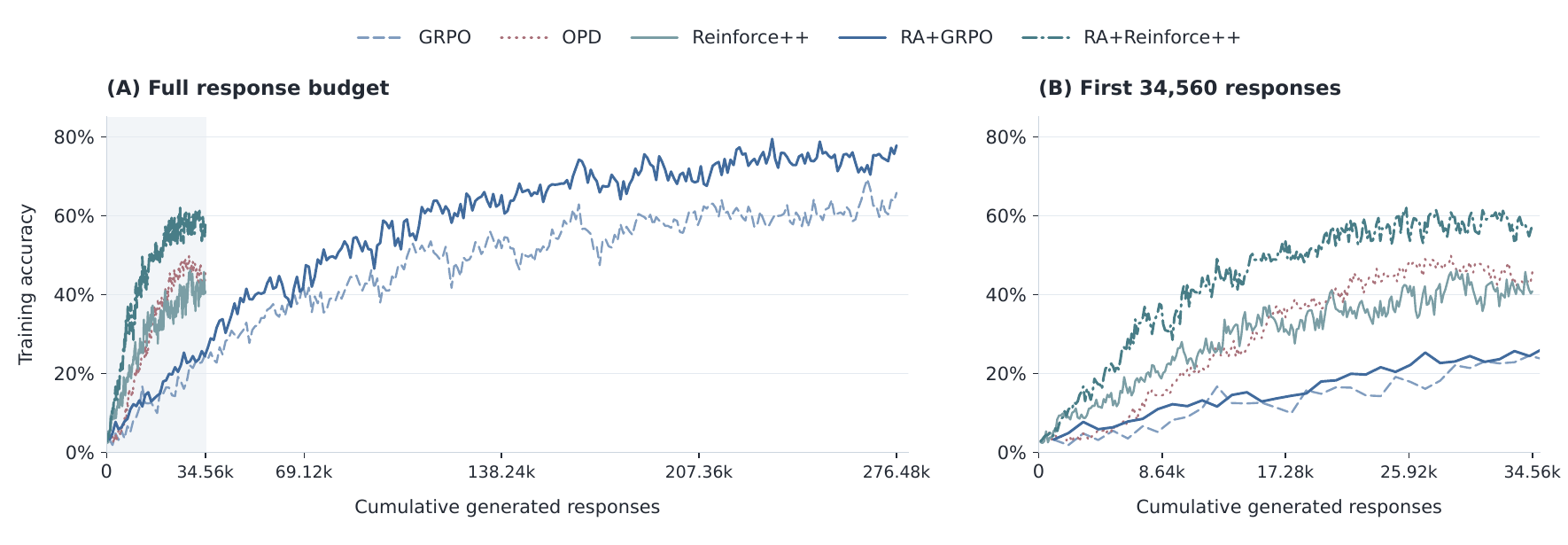}
\caption{\textbf{Training accuracy of Qwen3-4B-Base on DAPO-Math-17K.} (A) Full two-epoch response budgets: 276,480 generations for GRPO runs and 34,560 for the one-response runs. (B) The first 34,560 generations, the shaded region in (A), show the early trajectories at the same response budget. Curves show logged training accuracy; \CoRA{} is omitted.}
\label{fig:dynamics}
\end{figure}

Figure~\ref{fig:dynamics} reports logged training accuracy for the Qwen3-4B-Base student across the five frozen-teacher and verifier runs under the two-epoch DAPO-Math-17K protocol of Table~\ref{tab:setup}. The horizontal coordinate counts scheduled student generations: each processed prompt contributes eight responses in GRPO runs and one in OPD and REINFORCE++ runs. Teacher scoring and updates, retries, filtering, and evaluation lie outside this count.

\RA{} maintains higher training accuracy than its verifier-only estimator over most of training. Among the one-response runs, OPD rises more slowly than REINFORCE++ at first and later exceeds it, while staying below \RA{}+REINFORCE++ over most of the shared response budget, as the expanded early window shows. Teacher-only OPD trains without verifier feedback and without the length shaping of the verifier runs, so its trajectory reflects a different recipe and not only a different label.

Training accuracy measures correctness on the training prompts under the rollout policy, whereas Table~\ref{tab:main} evaluates held-out benchmarks at a different sampling temperature, so the sizes of the two gains need not agree.

\FloatBarrier
\subsection{Guidance magnitude and response length}
\label{app:training-diagnostics}
\label{sec:training-diagnostics}

\begin{figure}[!htbp]
\centering
\begin{minipage}[t]{0.49\linewidth}
\centering
\includegraphics[width=\linewidth]{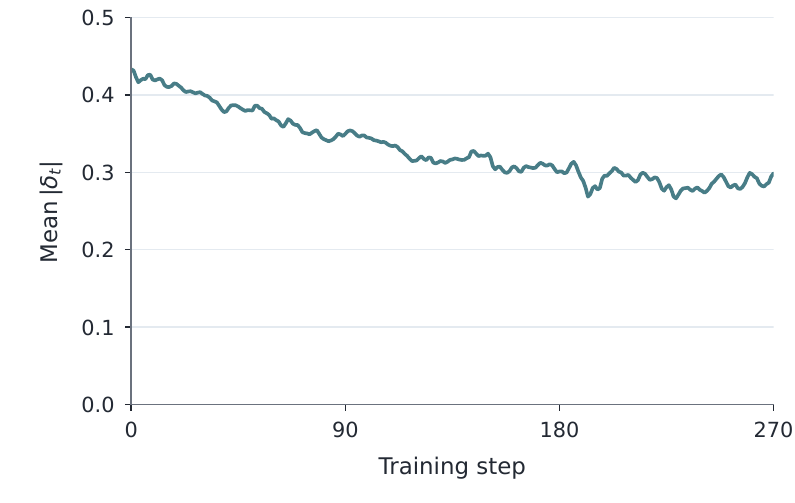}\\[-2pt]
{\small (a) Teacher guidance}
\end{minipage}\hfill
\begin{minipage}[t]{0.49\linewidth}
\centering
\includegraphics[width=\linewidth]{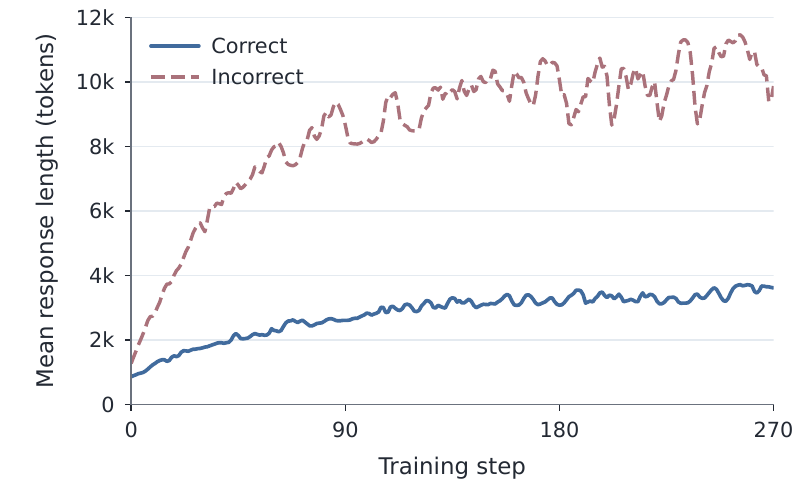}\\[-2pt]
{\small (b) Response length by correctness}
\end{minipage}
\caption{\textbf{Training diagnostics for the 4B student with \RA{}+REINFORCE++.}
The guidance weight is fixed at $\lambda=10$, and the horizontal axis counts rollout batches. (a) Recorded mean absolute guidance term $|\delta_t|$, averaged over the tokens of each batch. (b) Mean response length separated by verifier correctness. The curves are from a single training seed.}
\label{fig:signal-length}
\end{figure}

At the fixed weight $\lambda=10$, the recorded mean absolute guidance term falls from about $0.43$ to about $0.30$ over training (Figure~\ref{fig:signal-length}a). The teacher signal weakens in this run without any weight reduction, as the distance bound of Equation~\ref{eq:bound} allows, while still contributing to the actor labels at the end of training. The quantity is a token average over responses whose lengths change during training, so it is read together with panel (b).

Correct-response mean length rises but stays below about 4,000 tokens. Incorrect-response mean length rises more sharply and later fluctuates mostly between 9,000 and 12,000 tokens, close to the cap; the shaped reward penalizes length only for correct responses, so nothing in the reward opposes the growth of incorrect ones. Correct responses in this run therefore stay short relative to the cap, and failed responses dominate the token count, which is where a per-token modulation of a shared negative label has the most to act on. \RA{} supplies local guidance at student-visited prefixes without complete teacher trajectories; length alone does not measure exploration.

\FloatBarrier
\subsection{Where the guidance lands}
\label{app:buckets}

Table~\ref{tab:buckets} is computed offline rather than logged during training. The untrained Qwen3-1.7B-Base student generates one response for each of the 256 validation prompts of Appendix~\ref{app:validation}, under the training sampling settings of Table~\ref{tab:setup} (vLLM, temperature $1.0$, full support, 12,000-token cap), and the student and teacher distributions at every visited prefix are then recomputed in float32 over the full aligned vocabulary, giving 176,628 tokens. Two labels are computed on these same responses, full \RA{} ($\lambda=10$) and the centered log-ratio control ($\gamma_{\log}=0.6$, Table~\ref{tab:alternatives}), and the tokens are bucketed by the student's probability of the sampled token. For each bucket the table reports the share of tokens and the share of the summed absolute guidance term, $|\delta_t|$ for \RA{} and the centered log-ratio term of Equation~\ref{eq:component-labels} for the control; the ratio of the two shares is the bucket's enrichment. Because both labels are evaluated on the same tokens, the two columns share one set of token shares; the statistic depends only on the two distributions and the sampled tokens, not on correctness. The last row is the overlapping set of tokens with teacher probability below $10^{-4}$. Both labels apply the student value baseline and response centering, so the two columns differ only in the reward. The residual concentrates on tokens where the student is uncertain, whereas the log ratio concentrates on tokens that one side considers nearly impossible, above all on tokens the teacher assigns less than $10^{-4}$, where its label diverges (Proposition~\ref{prop:gate}); Section~\ref{sec:gate} discusses the pattern.

\begin{table}[!htbp]
\caption{\textbf{Where the guidance lands.} Share of tokens and of the summed absolute guidance term per bucket of the student's probability of the sampled token, for full \RA{} and the centered log-ratio control, both computed offline on the same 256 responses of the untrained 1.7B student, with the ratio of the two shares in parentheses. The last row overlaps the others.}
\label{tab:buckets}
\centering
\begingroup
\small
\setlength{\tabcolsep}{5pt}
\renewcommand{\arraystretch}{1.0}
\begin{tabular}{@{}lccc@{}}
\toprule
\tablerowshade{tableheadcol}
Tokens with & Share of tokens & Share of \RA{} guidance & Share of centered log-ratio guidance\\
\midrule
$p(a_t)\ge0.9$ & 60.8\% & 0.4\% ($0.01\times$) & 0.6\% ($0.01\times$)\\
$0.5\le p(a_t)<0.9$ & 11.9\% & 17.8\% ($1.5\times$) & 7.1\% ($0.6\times$)\\
$0.1\le p(a_t)<0.5$ & 11.8\% & \textbf{58.9\%} ($5.0\times$) & 31.6\% ($2.7\times$)\\
$p(a_t)<0.1$ & 15.5\% & 22.9\% ($1.5\times$) & \textbf{60.7\%} ($3.9\times$)\\
\midrule
$q(a_t)<10^{-4}$ & 3.8\% & 4.1\% ($1.1\times$) & \textbf{50.1\%} ($13\times$)\\
\bottomrule
\end{tabular}
\endgroup

\end{table}

\paragraph{The response mean of the local advantages.}
Table~\ref{tab:response-mean} reports, on the same 256 responses, the mean over a response's tokens of the local advantage, $\overline A^d$, and of the raw residual $d_t(a_t)$, both scaled by $\lambda=10$. Without response centering, the first is what the teacher adds to the response's mean label (Equation~\ref{eq:correction}) and the second is the corresponding shift of the raw residual control (Equation~\ref{eq:centering-controls}). Across tokens the local advantage averages to $3\times10^{-4}$, so the student value baseline holds in practice; its response mean is a fluctuation whose spread falls with length, from an SD of $0.32$ below $1{,}000$ tokens to $0.03$ between $1{,}000$ and $4{,}000$, and $8\%$ of all responses lie beyond $0.5$ in absolute value, most of them positive (the 95th percentile is $+0.54$ and the 5th $-0.29$). The raw residual shifts every length group by a negative amount, $-0.32$ on average, with $26\%$ of responses beyond $0.5$. Only three of the 256 responses are correct, so no split by correctness is reported. Section~\ref{sec:baselines} uses both facts.

\begin{table}[!htbp]
\caption{\textbf{Response mean of the guidance without centering.} Mean and SD across responses of $\lambda\overline A^d$ (local advantage, with the student value baseline) and of $\lambda\,\overline{d_t(a_t)}$ (raw residual), on the 256 offline responses of Table~\ref{tab:buckets}, with $\lambda=10$; $M$ is the number of tokens and $n$ the number of responses in each group.}
\label{tab:response-mean}
\centering
\begingroup
\small
\setlength{\tabcolsep}{5pt}
\renewcommand{\arraystretch}{1.0}
\begin{tabular}{@{}lccccc@{}}
\toprule
\tablerowshade{tableheadcol}
Responses & $n$ & \multicolumn{2}{c}{$\lambda\overline A^d$} & \multicolumn{2}{c}{Raw residual}\\
\cmidrule(lr){3-4}\cmidrule(l){5-6}
 & & Mean & SD & Mean & SD\\
\midrule
All & 256 & $+0.03$ & $0.30$ & $-0.32$ & $0.73$\\
$M<1{,}000$ & 216 & $+0.04$ & $0.32$ & $-0.31$ & $0.78$\\
$1{,}000\le M\le4{,}000$ & 34 & $+0.00$ & $0.03$ & $-0.38$ & $0.29$\\
$M>4{,}000$ & 6 & $+0.00$ & $0.01$ & $-0.05$ & $0.24$\\
\bottomrule
\end{tabular}
\endgroup

\end{table}

\FloatBarrier
\subsection{How the sampled label and the direct loss share the update}
\label{app:shared-path}

The direct $L_2$ loss of Section~\ref{sec:alternatives} and the uncentered residual label have the same expected teacher gradient at each visited prefix (Appendix~\ref{app:local-objective}), yet the loss trails both residual labels at every swept weight (Table~\ref{tab:weight-sweep}). Table~\ref{tab:shared-path} compares the three runs at the reference weight $10$ on Qwen3-1.7B-Base with REINFORCE++, over their three seeds. On each logged batch, $g_T$ and $g_V$ are the gradients of the teacher term and of the verifier term. The conditional noise is $\operatorname{tr}\operatorname{Cov}(g_T)/\norm{\E g_T}^2$ over the sampled tokens at the visited prefixes. The shared-path term compares $\cos(g_T,g_V)$ on the sampled tokens with the same cosine when the teacher term is evaluated on independently resampled tokens. A unit of teacher-gradient mass is outcome-compatible when it raises a correct response or lowers an incorrect one. The update $u$ is the parameter step after global gradient clipping and Adam, $u_V$ the step the verifier term alone would take, $s_T$ the teacher's share of $u$, and the next-batch probe evaluates, on the following batch, the gain in the verifier objective produced by each step.

\begin{table}[!htbp]
\caption{\textbf{The sampled label reproduces the direct loss's expected teacher gradient but shares the update with the verifier.} Qwen3-1.7B-Base with REINFORCE++ at weight $10$; aggregates over three seeds.}
\label{tab:shared-path}
\centering
\begingroup
\small
\setlength{\tabcolsep}{5pt}
\renewcommand{\arraystretch}{1.0}
\begin{tabular}{@{}lccc@{}}
\toprule
\tablerowshade{tableheadcol}
Statistic & Full \RA{} & Uncentered & Direct $L_2$\\
\midrule
Cosine of sampled and analytic $g_T$ & 0.994 & 0.992 & 1.000\\
Conditional noise of $g_T$ & 0.85 & 1.08 & 0\\
$\cos(g_T,g_V)$, sampled tokens & $+0.18$ & $+0.13$ & $0.00$\\
$\cos(g_T,g_V)$, resampled tokens & $+0.02$ & $0.00$ & --\\
Outcome-compatible teacher mass & 64\% & 60\% & 49\%\\
Batches with $\cos(g_T,g_V)<0$ & 26\% & 33\% & 50\%\\
RMS $\norm{g_T}/\norm{g_V}$ & 0.58 & 0.62 & 0.43\\
Global clip rate & 39\% & 43\% & 25\%\\
Teacher share of the update $s_T$ & 0.28 & 0.31 & 0.44\\
$\cos(u,u_V)$ & 0.93 & 0.91 & 0.80\\
Verifier retention & 0.94 & 0.91 & 0.80\\
Next-batch verifier gain & 1.13$\times$ & 1.08$\times$ & 1.00$\times$\\
Paired win rate against $L_2$ & 70\% & 63\% & --\\
\bottomrule
\end{tabular}
\endgroup

\end{table}

Averaged over 64 resampled tokens, the sampled teacher gradient matches its analytic expectation (cosine $0.994$ for \RA{}, norm ratio $1.001$), so the gap does not lie in the expected teacher gradient. The sampled gradient is noisier; for the uncentered label, most of its extra second moment relative to the direct loss is the estimator's own noise ($75$--$90\%$), and $15$--$25\%$ comes from sharing the verifier's sampled path: $\cos(g_T,g_V)$ is $+0.18$ for \RA{} on the sampled tokens and $+0.02$ on resampled ones, whereas the direct loss, which acts on every token, is uncorrelated with the verifier. As a result, $64\%$ of \RA{}'s teacher-gradient mass is outcome-compatible against $49\%$ for the direct loss, mainly because more of it lowers incorrect responses ($46\%$ against $35\%$). The direct loss has the smaller gradient ($0.43$ of the verifier's, against $0.58$), but it is deterministic given the prefixes, and after clipping and Adam it takes the larger share of the update, $0.44$ against $0.28$. Over training this share rises from $0.36$ to $0.48$ for the direct loss and falls from $0.30$ to $0.25$ for \RA{}, and the cosine of the update with the verifier's step falls from $0.88$ to $0.74$ for the direct loss while \RA{} stays at $0.92$--$0.94$. On the next batch, \RA{}'s step raises the verifier objective $1.10$--$1.16$ times as much as the direct loss's step and wins $66$--$74\%$ of paired comparisons. The ordering \RA{}, uncentered label, direct loss holds on every seed for the shared-path term, the outcome-compatible mass, the teacher share, and the cosine with the verifier's step. The divergence between the uncentered label and the direct loss grows with the weight (update cosine $0.97$, $0.91$, and $0.83$ at $5$, $10$, and $20$), as does the margin of \RA{} over the direct loss in Table~\ref{tab:weight-sweep}, and it lies mostly on tokens with $0.1\le p(a_t)<0.5$ ($55\%$), the forks of Section~\ref{sec:gate}. Response centering corrects the uncentered label mainly on responses shorter than $1{,}000$ tokens ($79\%$ of the correction), in line with Section~\ref{sec:baselines}.

\FloatBarrier
\section{Three-seed results}
\label{app:seeds}

Each method is evaluated over three training seeds under the protocol of Appendix~\ref{app:protocol}. Table~\ref{tab:seed-reporting} first averages the three benchmarks equally within each seed, then reports the mean and sample standard deviation (SD) of the three macro scores; SD uses denominator $3-1$ and describes variation across runs rather than a standard error or confidence interval. S1--S3 identify the three runs.

\begin{table}[!htbp]
\caption{\textbf{Macro performance across three training seeds.} Scores in \%; $\pm$ denotes sample SD across the three run-level macro scores. The middle block lists the 1.7B REINFORCE++ controls of Section~\ref{sec:analysis}, whose reference is the \RA{}+REINFORCE++ run of the first block; log-ratio weights are in parentheses.}
\label{tab:seed-reporting}
\centering
\begingroup
\footnotesize
\setlength{\tabcolsep}{1.5pt}
\renewcommand{\arraystretch}{1.0}
\begin{tabular}{p{\dimexpr0.3861\linewidth-2\tabcolsep\relax}R{\dimexpr0.0604\linewidth-2\tabcolsep\relax}R{\dimexpr0.0604\linewidth-2\tabcolsep\relax}R{\dimexpr0.0604\linewidth-2\tabcolsep\relax}R{\dimexpr0.1182\linewidth-2\tabcolsep\relax}p{\dimexpr0.0151\linewidth-2\tabcolsep\relax}R{\dimexpr0.0604\linewidth-2\tabcolsep\relax}R{\dimexpr0.0604\linewidth-2\tabcolsep\relax}R{\dimexpr0.0604\linewidth-2\tabcolsep\relax}R{\dimexpr0.1182\linewidth-2\tabcolsep\relax}}
\toprule
\tabletwoheadshade{tableheadcol}
\tabletwoheadlabel{Method} & \multicolumn{4}{c}{Avg@8} & & \multicolumn{4}{c}{Pass@8}\\
\cmidrule(lr){2-5}\cmidrule(l){7-10}
 & S1 & S2 & S3 & Mean $\pm$ SD & & S1 & S2 & S3 & Mean $\pm$ SD\\
\midrule
\multicolumn{10}{l}{\textit{Qwen3-1.7B-Base}}\\
\modelgrouprule
\modelindent OPD & 27.61 & 27.17 & 27.58 & 27.45\sd{0.25} & & 41.60 & 40.69 & 41.60 & 41.30\sd{0.53}\\
\modelindent REINFORCE++ & 26.12 & 24.91 & 25.97 & 25.67\sd{0.66} & & 41.20 & 37.93 & 39.18 & 39.44\sd{1.65}\\
\tablerowshade{reinforcecol}\modelindent \RA{}+REINFORCE++ & 29.36 & 28.96 & 29.43 & 29.25\sd{0.25} & & 46.11 & 43.11 & 46.38 & 45.20\sd{1.81}\\
\tablerowshade{reinforcecol}\modelindent \CoRA{}+REINFORCE++ & 30.13 & 30.65 & 31.33 & 30.70\sd{0.61} & & 44.29 & 47.36 & 47.49 & 46.38\sd{1.81}\\
\modelindent GRPO & 27.90 & 27.38 & 27.19 & 27.49\sd{0.37} & & 40.56 & 40.22 & 40.42 & 40.40\sd{0.17}\\
\tablerowshade{grpocol}\modelindent \RA{}+GRPO & 29.94 & 29.49 & 30.13 & 29.86\sd{0.33} & & 44.42 & 44.22 & 45.60 & 44.75\sd{0.74}\\
\tablerowshade{grpocol}\modelindent \CoRA{}+GRPO & 30.28 & 31.90 & 31.09 & 31.09\sd{0.81} & & 43.04 & 45.53 & 44.29 & 44.29\sd{1.24}\\
\midrule
\multicolumn{10}{l}{\textit{Qwen3-1.7B-Base, controls with REINFORCE++ (Section~\ref{sec:analysis})}}\\
\modelgrouprule
\modelindent w/o student value baseline & 27.22 & 27.44 & 26.82 & 27.16\sd{0.31} & & 41.60 & 43.11 & 40.16 & 41.62\sd{1.48}\\
\modelindent w/o response centering & 28.80 & 28.31 & 29.01 & 28.71\sd{0.36} & & 42.78 & 45.20 & 46.24 & 44.74\sd{1.78}\\
\modelindent Raw residual (w/o both) & 27.89 & 27.63 & 27.04 & 27.52\sd{0.43} & & 42.71 & 43.04 & 41.53 & 42.43\sd{0.79}\\
\modelindent Clipped centered log ratio ($c{=}5$, $0.6$) & 28.28 & 27.73 & 28.12 & 28.04\sd{0.28} & & 44.09 & 42.91 & 43.51 & 43.50\sd{0.59}\\
\modelindent Centered log ratio ($0.6$) & 28.18 & 27.45 & 27.93 & 27.85\sd{0.37} & & 42.91 & 42.98 & 43.31 & 43.07\sd{0.21}\\
\modelindent Raw log ratio ($0.2$) & 27.51 & 26.85 & 27.47 & 27.28\sd{0.37} & & 42.71 & 41.73 & 43.04 & 42.50\sd{0.68}\\
\modelindent Direct $L_2$ loss ($\alpha{=}10$) & 28.26 & 28.13 & 27.67 & 28.02\sd{0.31} & & 44.02 & 43.31 & 42.98 & 43.44\sd{0.53}\\
\modelindent \RA{} with $\lambda{=}5$ & 28.62 & 29.00 & 28.89 & 28.84\sd{0.20} & & 45.27 & 44.42 & 45.13 & 44.94\sd{0.45}\\
\modelindent \RA{} with $\lambda{=}20$ & 29.67 & 30.75 & 27.45 & 29.29\sd{1.68} & & 45.33 & 48.09 & 42.44 & 45.29\sd{2.82}\\
\midrule
\multicolumn{10}{l}{\textit{Qwen3-4B-Base}}\\
\modelgrouprule
\modelindent OPD & 34.60 & 34.25 & 34.97 & 34.61\sd{0.36} & & 48.11 & 47.91 & 51.51 & 49.18\sd{2.02}\\
\modelindent REINFORCE++ & 34.19 & 33.73 & 33.67 & 33.86\sd{0.28} & & 47.91 & 47.64 & 44.51 & 46.69\sd{1.89}\\
\tablerowshade{reinforcecol}\modelindent \RA{}+REINFORCE++ & 37.31 & 36.81 & 37.25 & 37.12\sd{0.27} & & 53.16 & 52.89 & 53.02 & 53.02\sd{0.13}\\
\tablerowshade{reinforcecol}\modelindent \CoRA{}+REINFORCE++ & 38.53 & 38.14 & 38.51 & 38.39\sd{0.22} & & 50.93 & 50.60 & 50.80 & 50.78\sd{0.17}\\
\modelindent GRPO & 35.55 & 35.90 & 36.06 & 35.83\sd{0.26} & & 47.91 & 48.11 & 48.04 & 48.02\sd{0.10}\\
\tablerowshade{grpocol}\modelindent \RA{}+GRPO & 37.29 & 37.56 & 37.87 & 37.57\sd{0.29} & & 50.73 & 50.93 & 54.07 & 51.91\sd{1.87}\\
\tablerowshade{grpocol}\modelindent \CoRA{}+GRPO & 38.30 & 38.78 & 38.73 & 38.60\sd{0.26} & & 52.89 & 53.09 & 53.02 & 53.00\sd{0.10}\\
\bottomrule
\end{tabular}
\endgroup

\end{table}

\subsection{Benchmark-level results for the controls}
\label{app:component-seeds}
\label{app:benchmark-controls}

Table~\ref{tab:benchmark-controls} lists benchmark-level three-seed means for the control runs of Section~\ref{sec:analysis}, with the labels of Tables~\ref{tab:weight-sweep} and~\ref{tab:alternatives}. The reference runs repeat Table~\ref{tab:main}; the teacher-peak branch runs of Section~\ref{sec:branch} are in Appendix~\ref{app:branch-results}. Macro means and sample SD use per-seed macro scores, so recomputing them from the displayed benchmark values can differ in the last decimal.

Each control uses the same weight across its three seeds. The log-ratio and no-baseline weights multiply the complete response-centered guidance term in Equation~\ref{eq:component-labels}; the direct-$L_2$ weight scales the quadratic auxiliary loss in Equation~\ref{eq:component-direct-l2}; the raw residual and the raw log ratio add the uncentered sampled reward at the weights $\gamma_{\rm raw}=10$ and $\beta=0.2$ (Equation~\ref{eq:centering-controls}; Appendix~\ref{app:controls}). Log-ratio and direct-$L_2$ controls appear at the weights reported in Table~\ref{tab:alternatives}.

\begin{table}[!htbp]
\caption{\textbf{Benchmark-level results for the control runs of Section~\ref{sec:analysis}.} Qwen3-1.7B-Base with REINFORCE++; three-seed means in \%. The macro columns repeat Tables~\ref{tab:weight-sweep} and~\ref{tab:alternatives}; sample SD are reported in Table~\ref{tab:seed-reporting}.}
\label{tab:benchmark-controls}
\centering
\begingroup
\footnotesize
\setlength{\tabcolsep}{1pt}
\renewcommand{\arraystretch}{1.0}
\begin{tabular}{p{\dimexpr0.3760\linewidth-2\tabcolsep\relax}R{\dimexpr0.0780\linewidth-2\tabcolsep\relax}R{\dimexpr0.0780\linewidth-2\tabcolsep\relax}R{\dimexpr0.0780\linewidth-2\tabcolsep\relax}R{\dimexpr0.0780\linewidth-2\tabcolsep\relax}R{\dimexpr0.0780\linewidth-2\tabcolsep\relax}R{\dimexpr0.0780\linewidth-2\tabcolsep\relax}R{\dimexpr0.0780\linewidth-2\tabcolsep\relax}R{\dimexpr0.0780\linewidth-2\tabcolsep\relax}}
\toprule
\tabletwoheadshade{tableheadcol}
\tabletwoheadlabel{Variant} & \multicolumn{2}{c}{MATH-500} & \multicolumn{2}{c}{AIME24} & \multicolumn{2}{c}{AIME25} & \multicolumn{2}{c}{Macro}\\
\cmidrule(lr){2-3}\cmidrule(lr){4-5}\cmidrule(lr){6-7}\cmidrule(l){8-9}
 & Avg@8 & Pass@8 & Avg@8 & Pass@8 & Avg@8 & Pass@8 & Avg@8 & Pass@8\\
\midrule
Verifier only (REINFORCE++) & 65.76 & 83.87 & 7.78 & 20.00 & 3.47 & 14.44 & 25.67 & 39.44\\
Teacher only (OPD) & 67.78 & 85.00 & 8.61 & 23.33 & 5.97 & 15.56 & 27.45 & 41.30\\
\tablerowshade{reinforcecol}Full \RA{} ($\lambda{=}10$) & 69.69 & 85.60 & 13.06 & 30.00 & 5.00 & 20.00 & 29.25 & 45.20\\
\addlinespace[2pt]
\modelgrouprule
\multicolumn{9}{l}{\textit{Probability residual, baselines removed}}\\
w/o student value baseline & 67.17 & 84.87 & 9.72 & 24.44 & 4.58 & 15.56 & 27.16 & 41.62\\
w/o response centering & 69.03 & 85.33 & 12.36 & 28.89 & 4.72 & 20.00 & 28.71 & 44.74\\
Raw residual (w/o both) & 67.83 & 85.07 & 10.42 & 25.56 & 4.31 & 16.67 & 27.52 & 42.43\\
\addlinespace[2pt]
\modelgrouprule
\multicolumn{9}{l}{\textit{Log ratio in place of the probability residual}}\\
Clipped centered log ratio ($c{=}5$) & 68.43 & 86.07 & 10.97 & 25.56 & 4.72 & 18.89 & 28.04 & 43.50\\
Centered log ratio & 68.13 & 85.87 & 10.69 & 25.56 & 4.72 & 17.78 & 27.85 & 43.07\\
Raw log ratio & 67.53 & 85.27 & 10.00 & 24.44 & 4.31 & 17.78 & 27.28 & 42.50\\
\addlinespace[2pt]
\modelgrouprule
\multicolumn{9}{l}{\textit{Direct loss}}\\
Direct $L_2$ loss & 68.50 & 85.87 & 10.83 & 26.67 & 4.72 & 17.78 & 28.02 & 43.44\\
\addlinespace[2pt]
\modelgrouprule
\multicolumn{9}{l}{\textit{Guidance weight}}\\
\RA{} with $\lambda{=}5$ & 68.73 & 85.93 & 12.36 & 28.89 & 5.42 & 20.00 & 28.84 & 44.94\\
\RA{} with $\lambda{=}20$ & 68.98 & 85.87 & 13.06 & 31.11 & 5.83 & 18.89 & 29.29 & 45.29\\
\bottomrule
\end{tabular}
\endgroup

\end{table}

\subsection{Per-seed benchmark results for the controls}
\label{app:centering-seeds}

Table~\ref{tab:component-seed-results} gives per-seed benchmark results for seven controls: no student value baseline, centered log ratio, and clipped centered log ratio (Equation~\ref{eq:component-labels}), no response centering and the raw residual (Equation~\ref{eq:centering-controls}), the raw log ratio (Appendix~\ref{app:controls}), and direct $L_2$ (Equation~\ref{eq:component-direct-l2}). Each cell gives Avg@8 / Pass@8 in percent; macro scores weight the three benchmarks equally. The weight sweeps follow in Tables~\ref{tab:sweep-mass} and~\ref{tab:sweep-log}.

\begin{table}[!htbp]
\caption{\textbf{Per-seed benchmark results for seven controls.} Qwen3-1.7B-Base with REINFORCE++; entries are Avg@8 / Pass@8 in \%. Weights: no student value baseline $\gamma_x=10$, no response centering $\lambda=10$, raw residual $\gamma_{\rm raw}=10$, clipped ($c=5$) and centered log ratio $\gamma_{\log}=0.6$, raw log ratio $\beta=0.2$, and direct $L_2$ $\alpha=10$.}
\label{tab:component-seed-results}
\centering
\begingroup
\small
\setlength{\tabcolsep}{5pt}
\renewcommand{\arraystretch}{1.0}
\begin{tabular*}{\linewidth}{@{\extracolsep{\fill}}llcccc@{}}
\toprule
\tablerowshade{tableheadcol}
Control & Seed & MATH-500 & AIME24 & AIME25 & Macro\\
\midrule
 & S1 & 66.65 / 84.80 & 10.83 / 23.33 & 4.17 / 16.67 & 27.22 / 41.60\\
 & S2 & 67.73 / 86.00 & 10.42 / 23.33 & 4.17 / 20.00 & 27.44 / 43.11\\
\multirow{-3}{*}{w/o student value baseline} & S3 & 67.13 / 83.80 & 7.92 / 26.67 & 5.42 / 10.00 & 26.82 / 40.16\\
\midrule
 & S1 & 68.48 / 85.00 & 13.33 / 26.67 & 4.58 / 16.67 & 28.80 / 42.78\\
 & S2 & 69.50 / 85.60 & 11.25 / 30.00 & 4.17 / 20.00 & 28.31 / 45.20\\
\multirow{-3}{*}{w/o response centering} & S3 & 69.13 / 85.40 & 12.50 / 30.00 & 5.42 / 23.33 & 29.01 / 46.24\\
\midrule
 & S1 & 67.83 / 84.80 & 11.67 / 26.67 & 4.17 / 16.67 & 27.89 / 42.71\\
 & S2 & 67.48 / 85.80 & 11.25 / 26.67 & 4.17 / 16.67 & 27.63 / 43.04\\
\multirow{-3}{*}{Raw residual (w/o both)} & S3 & 68.20 / 84.60 & 8.33 / 23.33 & 4.58 / 16.67 & 27.04 / 41.53\\
\midrule
 & S1 & 68.58 / 85.60 & 11.25 / 26.67 & 5.00 / 20.00 & 28.28 / 44.09\\
 & S2 & 69.03 / 85.40 & 9.17 / 26.67 & 5.00 / 16.67 & 27.73 / 42.91\\
\multirow{-3}{*}{Clipped centered log ratio ($c{=}5$)} & S3 & 67.70 / 87.20 & 12.50 / 23.33 & 4.17 / 20.00 & 28.12 / 43.51\\
\midrule
 & S1 & 68.70 / 85.40 & 10.42 / 23.33 & 5.42 / 20.00 & 28.18 / 42.91\\
 & S2 & 68.18 / 85.60 & 9.17 / 30.00 & 5.00 / 13.33 & 27.45 / 42.98\\
\multirow{-3}{*}{Centered log ratio} & S3 & 67.53 / 86.60 & 12.50 / 23.33 & 3.75 / 20.00 & 27.93 / 43.31\\
\midrule
 & S1 & 67.13 / 84.80 & 9.17 / 23.33 & 6.25 / 20.00 & 27.51 / 42.71\\
 & S2 & 67.23 / 85.20 & 10.00 / 23.33 & 3.33 / 16.67 & 26.85 / 41.73\\
\multirow{-3}{*}{Raw log ratio} & S3 & 68.25 / 85.80 & 10.83 / 26.67 & 3.33 / 16.67 & 27.47 / 43.04\\
\midrule
 & S1 & 68.53 / 85.40 & 12.08 / 30.00 & 4.17 / 16.67 & 28.26 / 44.02\\
 & S2 & 68.13 / 86.60 & 11.25 / 26.67 & 5.00 / 16.67 & 28.13 / 43.31\\
\multirow{-3}{*}{Direct $L_2$ loss} & S3 & 68.85 / 85.60 & 9.17 / 23.33 & 5.00 / 20.00 & 27.67 / 42.98\\
\bottomrule
\end{tabular*}
\endgroup

\end{table}

\subsection{Teacher-peak branch results}
\label{app:branch-results}

Tables~\ref{tab:branch-benchmarks} and~\ref{tab:branch-seeds} list benchmark-level means and per-seed macro scores for the branch runs of Section~\ref{sec:branch}; the base runs are in Tables~\ref{tab:sampled-opd}, \ref{tab:seed-reporting}, and~\ref{tab:sweep-mass}. At every weight, each seed of \RA{} with the branch falls below the same seed of \RA{}. With REINFORCE++, each seed with the branch falls below the same verifier-only seed at the weights $5$ and $10$, and two of the three seeds do at $20$.

\begin{table}[!htbp]
\caption{\textbf{Benchmark-level results for the teacher-peak branch.} Qwen3-1.7B-Base; three-seed means in \%, with sample SD for the macro scores.}
\label{tab:branch-benchmarks}
\centering
\begingroup
\footnotesize
\setlength{\tabcolsep}{3pt}
\renewcommand{\arraystretch}{1.0}
\begin{tabular}{@{}lccccccccc@{}}
\toprule
\tabletwoheadshade{tableheadcol}
\tabletwoheadlabel{Base method + branch} & & \multicolumn{2}{c}{MATH-500} & \multicolumn{2}{c}{AIME24} & \multicolumn{2}{c}{AIME25} & \multicolumn{2}{c}{Macro}\\
\cmidrule(lr){3-4}\cmidrule(lr){5-6}\cmidrule(lr){7-8}\cmidrule(l){9-10}
 & $\lambda$ & Avg@8 & Pass@8 & Avg@8 & Pass@8 & Avg@8 & Pass@8 & Avg@8 & Pass@8\\
\midrule
Sampled OPD & -- & 63.23 & 83.27 & 7.50 & 20.00 & 4.58 & 14.44 & 25.10\sd{1.00} & 39.24\sd{2.53}\\
\addlinespace[2pt]
REINFORCE++ & 5 & 65.58 & 83.60 & 6.94 & 18.89 & 3.33 & 14.44 & 25.29\sd{1.10} & 38.98\sd{2.22}\\
REINFORCE++ & 10 & 65.06 & 82.40 & 6.81 & 21.11 & 2.50 & 12.22 & 24.79\sd{1.17} & 38.58\sd{3.02}\\
REINFORCE++ & 20 & 64.36 & 82.00 & 4.86 & 16.67 & 2.50 & 13.33 & 23.91\sd{1.91} & 37.33\sd{3.63}\\
\addlinespace[2pt]
\RA{}+REINFORCE++ & 5 & 66.35 & 84.20 & 9.31 & 24.44 & 4.72 & 14.44 & 26.79\sd{1.22} & 41.03\sd{2.07}\\
\RA{}+REINFORCE++ & 10 & 67.09 & 84.27 & 9.86 & 23.33 & 4.58 & 17.78 & 27.18\sd{1.00} & 41.79\sd{2.70}\\
\RA{}+REINFORCE++ & 20 & 66.81 & 84.20 & 9.72 & 22.22 & 4.86 & 17.78 & 27.13\sd{2.06} & 41.40\sd{2.83}\\
\bottomrule
\end{tabular}
\endgroup

\end{table}

\begin{table}[!htbp]
\caption{\textbf{Per-seed macro scores for the teacher-peak branch.} Entries are macro Avg@8 / Pass@8 in \%.}
\label{tab:branch-seeds}
\centering
\begingroup
\footnotesize
\setlength{\tabcolsep}{4pt}
\renewcommand{\arraystretch}{1.0}
\begin{tabular}{@{}lcccc@{}}
\toprule
\tablerowshade{tableheadcol}
Base method + branch & $\lambda$ & S1 & S2 & S3\\
\midrule
Sampled OPD & -- & 25.63 / 41.07 & 23.94 / 36.36 & 25.74 / 40.29\\
\addlinespace[2pt]
REINFORCE++ & 5 & 25.93 / 40.16 & 24.02 / 36.42 & 25.91 / 40.36\\
REINFORCE++ & 10 & 24.78 / 39.36 & 23.62 / 35.24 & 25.96 / 41.13\\
REINFORCE++ & 20 & 24.39 / 37.40 & 25.53 / 40.93 & 21.80 / 33.67\\
\addlinespace[2pt]
\RA{}+REINFORCE++ & 5 & 27.43 / 42.71 & 25.39 / 38.71 & 27.56 / 41.67\\
\RA{}+REINFORCE++ & 10 & 28.19 / 44.16 & 26.20 / 38.84 & 27.14 / 42.38\\
\RA{}+REINFORCE++ & 20 & 27.64 / 41.67 & 28.89 / 44.09 & 24.86 / 38.44\\
\bottomrule
\end{tabular}
\endgroup

\end{table}

\subsection{Weight sweeps}
\label{app:lambda-seeds}

Tables~\ref{tab:sweep-mass} and~\ref{tab:sweep-log} give three runs at each tested weight for Qwen3-1.7B-Base with REINFORCE++. The $\lambda=10$ row repeats the reference runs of Table~\ref{tab:main}, and the other rows come from the control runs of Tables~\ref{tab:weight-sweep} and~\ref{tab:alternatives}. Columns S1--S3 give each run's macro Avg@8 / Pass@8 in percent, and the last two columns give three-seed means with sample SD. The guidance weight $\lambda$ multiplies the complete guidance term in Equation~\ref{eq:correction}; the control weights are defined in Appendix~\ref{app:controls}. Table~\ref{tab:weight-sweep} repeats the three-seed means of Table~\ref{tab:sweep-mass}, and the verifier-only reference is in Table~\ref{tab:seed-reporting}. Table~\ref{tab:lambda-seed-results} breaks every residual run down by benchmark and seed.

\begin{table}[!htbp]
\caption{\textbf{Weight sweeps with the probability residual.} Qwen3-1.7B-Base with REINFORCE++; S1--S3 entries are macro Avg@8 / Pass@8 in \%. The shaded row is the reference configuration of Table~\ref{tab:main}, and $^{\dagger}$ marks the direct-$L_2$ weight selected on the validation split and reported in Table~\ref{tab:alternatives}. The weight is $\lambda$ for full \RA{} and the uncentered control, $\gamma_x$ and $\gamma_{\rm raw}$ for the two controls without the student value baseline, and $\alpha$ for direct $L_2$.}
\label{tab:sweep-mass}
\centering
\begingroup
\small
\setlength{\tabcolsep}{4pt}
\renewcommand{\arraystretch}{1.0}
\begin{tabular*}{\linewidth}{@{\extracolsep{\fill}}llccccc@{}}
\toprule
\tablerowshade{tableheadcol}
Variant & Weight & S1 & S2 & S3 & Avg@8 & Pass@8\\
\midrule
 & 5 & 28.62 / 45.27 & 29.00 / 44.42 & 28.89 / 45.13 & 28.84\sd{0.20} & 44.94\sd{0.45}\\
\tablerowshade{reinforcecol} & 10 & 29.36 / 46.11 & 28.96 / 43.11 & 29.43 / 46.38 & 29.25\sd{0.25} & 45.20\sd{1.81}\\
\multirow{-3}{*}{Full \RA{}} & 20 & 29.67 / 45.33 & 30.75 / 48.09 & 27.45 / 42.44 & 29.29\sd{1.68} & 45.29\sd{2.82}\\
\midrule
 & 5 & 27.70 / 42.84 & 29.19 / 45.40 & 28.22 / 45.13 & 28.37\sd{0.76} & 44.46\sd{1.40}\\
 & 10 & 28.80 / 42.78 & 28.31 / 45.20 & 29.01 / 46.24 & 28.71\sd{0.36} & 44.74\sd{1.78}\\
\multirow{-3}{*}{w/o response centering} & 20 & 29.46 / 46.38 & 30.33 / 46.58 & 26.18 / 41.07 & 28.66\sd{2.19} & 44.67\sd{3.13}\\
\midrule
 & 5 & 26.83 / 42.71 & 27.56 / 41.80 & 26.47 / 41.27 & 26.96\sd{0.56} & 41.93\sd{0.73}\\
 & 10 & 27.22 / 41.60 & 27.44 / 43.11 & 26.82 / 40.16 & 27.16\sd{0.31} & 41.62\sd{1.48}\\
\multirow{-3}{*}{w/o student value baseline} & 20 & 26.56 / 42.58 & 27.51 / 41.40 & 25.51 / 38.31 & 26.53\sd{1.00} & 40.76\sd{2.20}\\
\midrule
 & 5 & 27.54 / 44.22 & 28.25 / 43.24 & 26.64 / 40.42 & 27.48\sd{0.81} & 42.63\sd{1.97}\\
 & 10 & 27.89 / 42.71 & 27.63 / 43.04 & 27.04 / 41.53 & 27.52\sd{0.43} & 42.43\sd{0.79}\\
\multirow{-3}{*}{Raw residual (w/o both)} & 20 & 27.15 / 42.84 & 27.81 / 42.44 & 25.92 / 37.60 & 26.96\sd{0.96} & 40.96\sd{2.92}\\
\midrule
 & 5 & 28.18 / 44.02 & 28.00 / 43.31 & 27.31 / 42.84 & 27.83\sd{0.46} & 43.39\sd{0.59}\\
 & $10^{\dagger}$ & 28.26 / 44.02 & 28.13 / 43.31 & 27.67 / 42.98 & 28.02\sd{0.31} & 43.44\sd{0.53}\\
\multirow{-3}{*}{Direct $L_2$ loss ($\alpha$)} & 20 & 28.45 / 44.42 & 28.11 / 43.44 & 26.57 / 41.02 & 27.71\sd{1.00} & 42.96\sd{1.75}\\
\bottomrule
\end{tabular*}
\endgroup

\end{table}

\begin{table}[!htbp]
\caption{\textbf{Weight sweeps with the log ratio.} Qwen3-1.7B-Base with REINFORCE++; S1--S3 entries are macro Avg@8 / Pass@8 in \%. For each control, $^{\dagger}$ marks the weight selected on the validation split (Appendix~\ref{app:validation}) and reported in Table~\ref{tab:alternatives}.}
\label{tab:sweep-log}
\centering
\begingroup
\small
\setlength{\tabcolsep}{4pt}
\renewcommand{\arraystretch}{1.0}
\begin{tabular*}{\linewidth}{@{\extracolsep{\fill}}llccccc@{}}
\toprule
\tablerowshade{tableheadcol}
Variant & Weight & S1 & S2 & S3 & Avg@8 & Pass@8\\
\midrule
 & 0.05 & 26.70 / 41.40 & 26.36 / 41.73 & 26.71 / 41.67 & 26.59\sd{0.20} & 41.60\sd{0.18}\\
 & $0.2^{\dagger}$ & 27.51 / 42.71 & 26.85 / 41.73 & 27.47 / 43.04 & 27.28\sd{0.37} & 42.50\sd{0.68}\\
 & 0.6 & 27.02 / 42.04 & 26.53 / 41.53 & 27.04 / 41.33 & 26.86\sd{0.29} & 41.64\sd{0.37}\\
 & 1.2 & 25.27 / 38.78 & 25.05 / 39.18 & 25.25 / 38.98 & 25.19\sd{0.12} & 38.98\sd{0.20}\\
\multirow{-5}{*}{Raw log ratio ($\beta$)} & 2 & 24.79 / 37.40 & 23.71 / 37.80 & 24.55 / 37.67 & 24.35\sd{0.57} & 37.62\sd{0.20}\\
\midrule
 & 0.05 & 26.61 / 41.47 & 26.30 / 40.29 & 26.59 / 40.96 & 26.50\sd{0.17} & 40.90\sd{0.59}\\
 & 0.2 & 27.78 / 42.91 & 26.97 / 41.67 & 27.69 / 43.24 & 27.48\sd{0.44} & 42.61\sd{0.83}\\
 & $0.6^{\dagger}$ & 28.18 / 42.91 & 27.45 / 42.98 & 27.93 / 43.31 & 27.85\sd{0.37} & 43.07\sd{0.21}\\
 & 1.2 & 27.73 / 42.84 & 27.49 / 41.67 & 27.68 / 42.27 & 27.63\sd{0.13} & 42.26\sd{0.59}\\
\multirow{-5}{*}{Fully centered log ($\gamma_{\log}$)} & 2 & 27.08 / 42.00 & 26.79 / 40.62 & 27.06 / 41.29 & 26.98\sd{0.16} & 41.30\sd{0.69}\\
\midrule
 & 0.2 & 27.92 / 43.69 & 27.05 / 42.44 & 27.64 / 44.09 & 27.54\sd{0.44} & 43.41\sd{0.86}\\
 & $0.6^{\dagger}$ & 28.28 / 44.09 & 27.73 / 42.91 & 28.12 / 43.51 & 28.04\sd{0.28} & 43.50\sd{0.59}\\
\multirow{-3}{*}{Clipped centered log, $c{=}5$} & 1.2 & 27.95 / 43.04 & 27.66 / 41.73 & 27.91 / 43.38 & 27.84\sd{0.15} & 42.72\sd{0.87}\\
\bottomrule
\end{tabular*}
\endgroup

\end{table}

\begin{table}[!htbp]
\caption{\textbf{Per-seed benchmark results for the residual weight sweeps.} Qwen3-1.7B-Base with REINFORCE++; entries are Avg@8 / Pass@8 in \%. The shaded rows are the reference runs of Table~\ref{tab:main}. The weight is $\lambda$ for full \RA{} and the uncentered control, $\gamma_x$ for the control without the student value baseline, and $\gamma_{\rm raw}$ for the raw residual. Macro values average the three benchmark percentages within each run.}
\label{tab:lambda-seed-results}
\centering
\begingroup
\footnotesize
\setlength{\tabcolsep}{3pt}
\renewcommand{\arraystretch}{1.0}
\begin{tabular*}{\linewidth}{@{\extracolsep{\fill}}llccccc@{}}
\toprule
\tablerowshade{tableheadcol}
Variant & Weight & Seed & MATH-500 & AIME24 & AIME25 & Macro\\
\midrule
 &  & S1 & 65.45 / 83.60 & 8.75 / 20.00 & 4.17 / 20.00 & 26.12 / 41.20\\
 &  & S2 & 65.58 / 83.80 & 7.92 / 20.00 & 1.25 / 10.00 & 24.91 / 37.93\\
\multirow{-3}{*}{Verifier only} & \multirow{-3}{*}{\textcolor{black!45}{--}} & S3 & 66.25 / 84.20 & 6.67 / 20.00 & 5.00 / 13.33 & 25.97 / 39.18\\
\midrule
 &  & S1 & 68.35 / 85.80 & 12.50 / 30.00 & 5.00 / 20.00 & 28.62 / 45.27\\
 &  & S2 & 69.50 / 86.60 & 11.67 / 26.67 & 5.83 / 20.00 & 29.00 / 44.42\\
\multirow{-3}{*}{Full \RA{}} & \multirow{-3}{*}{5} & S3 & 68.35 / 85.40 & 12.92 / 30.00 & 5.42 / 20.00 & 28.89 / 45.13\\
\tablerowshade{reinforcecol} &  & S1 & 68.90 / 85.00 & 13.75 / 33.33 & 5.42 / 20.00 & 29.36 / 46.11\\
\tablerowshade{reinforcecol} &  & S2 & 70.23 / 86.00 & 12.08 / 26.67 & 4.58 / 16.67 & 28.96 / 43.11\\
\tablerowshade{reinforcecol}\multirow{-3}{*}{Full \RA{}} & \multirow{-3}{*}{10} & S3 & 69.95 / 85.80 & 13.33 / 30.00 & 5.00 / 23.33 & 29.43 / 46.38\\
 &  & S1 & 69.85 / 86.00 & 13.33 / 33.33 & 5.83 / 16.67 & 29.67 / 45.33\\
 &  & S2 & 71.00 / 87.60 & 14.17 / 33.33 & 7.08 / 23.33 & 30.75 / 48.09\\
\multirow{-3}{*}{Full \RA{}} & \multirow{-3}{*}{20} & S3 & 66.10 / 84.00 & 11.67 / 26.67 & 4.58 / 16.67 & 27.45 / 42.44\\
\midrule
 &  & S1 & 67.68 / 85.20 & 10.42 / 26.67 & 5.00 / 16.67 & 27.70 / 42.84\\
 &  & S2 & 69.25 / 86.20 & 12.92 / 30.00 & 5.42 / 20.00 & 29.19 / 45.40\\
\multirow{-3}{*}{w/o response centering} & \multirow{-3}{*}{5} & S3 & 68.00 / 85.40 & 12.08 / 26.67 & 4.58 / 23.33 & 28.22 / 45.13\\
 &  & S1 & 68.48 / 85.00 & 13.33 / 26.67 & 4.58 / 16.67 & 28.80 / 42.78\\
 &  & S2 & 69.50 / 85.60 & 11.25 / 30.00 & 4.17 / 20.00 & 28.31 / 45.20\\
\multirow{-3}{*}{w/o response centering} & \multirow{-3}{*}{10} & S3 & 69.13 / 85.40 & 12.50 / 30.00 & 5.42 / 23.33 & 29.01 / 46.24\\
 &  & S1 & 69.63 / 85.80 & 12.92 / 33.33 & 5.83 / 20.00 & 29.46 / 46.38\\
 &  & S2 & 70.58 / 86.40 & 14.17 / 33.33 & 6.25 / 20.00 & 30.33 / 46.58\\
\multirow{-3}{*}{w/o response centering} & \multirow{-3}{*}{20} & S3 & 65.20 / 83.20 & 10.00 / 23.33 & 3.33 / 16.67 & 26.18 / 41.07\\
\midrule
 &  & S1 & 66.75 / 84.80 & 10.83 / 26.67 & 2.92 / 16.67 & 26.83 / 42.71\\
 &  & S2 & 67.28 / 85.40 & 9.58 / 23.33 & 5.83 / 16.67 & 27.56 / 41.80\\
\multirow{-3}{*}{w/o student value baseline} & \multirow{-3}{*}{5} & S3 & 66.50 / 83.80 & 8.33 / 23.33 & 4.58 / 16.67 & 26.47 / 41.27\\
 &  & S1 & 66.65 / 84.80 & 10.83 / 23.33 & 4.17 / 16.67 & 27.22 / 41.60\\
 &  & S2 & 67.73 / 86.00 & 10.42 / 23.33 & 4.17 / 20.00 & 27.44 / 43.11\\
\multirow{-3}{*}{w/o student value baseline} & \multirow{-3}{*}{10} & S3 & 67.13 / 83.80 & 7.92 / 26.67 & 5.42 / 10.00 & 26.82 / 40.16\\
 &  & S1 & 67.18 / 84.40 & 10.00 / 26.67 & 2.50 / 16.67 & 26.56 / 42.58\\
 &  & S2 & 66.70 / 84.20 & 9.58 / 23.33 & 6.25 / 16.67 & 27.51 / 41.40\\
\multirow{-3}{*}{w/o student value baseline} & \multirow{-3}{*}{20} & S3 & 65.70 / 81.60 & 7.50 / 20.00 & 3.33 / 13.33 & 25.51 / 38.31\\
\midrule
 &  & S1 & 68.05 / 86.00 & 11.25 / 26.67 & 3.33 / 20.00 & 27.54 / 44.22\\
 &  & S2 & 67.68 / 86.40 & 10.42 / 26.67 & 6.67 / 16.67 & 28.25 / 43.24\\
\multirow{-3}{*}{Raw residual (w/o both)} & \multirow{-3}{*}{5} & S3 & 67.00 / 84.60 & 8.75 / 23.33 & 4.17 / 13.33 & 26.64 / 40.42\\
 &  & S1 & 67.83 / 84.80 & 11.67 / 26.67 & 4.17 / 16.67 & 27.89 / 42.71\\
 &  & S2 & 67.48 / 85.80 & 11.25 / 26.67 & 4.17 / 16.67 & 27.63 / 43.04\\
\multirow{-3}{*}{Raw residual (w/o both)} & \multirow{-3}{*}{10} & S3 & 68.20 / 84.60 & 8.33 / 23.33 & 4.58 / 16.67 & 27.04 / 41.53\\
 &  & S1 & 67.70 / 85.20 & 11.25 / 26.67 & 2.50 / 16.67 & 27.15 / 42.84\\
 &  & S2 & 66.75 / 84.00 & 10.00 / 23.33 & 6.67 / 20.00 & 27.81 / 42.44\\
\multirow{-3}{*}{Raw residual (w/o both)} & \multirow{-3}{*}{20} & S3 & 66.10 / 82.80 & 7.92 / 20.00 & 3.75 / 10.00 & 25.92 / 37.60\\
\bottomrule
\end{tabular*}
\endgroup

\end{table}

\subsection{Distillation combined with RL: sequential and joint}
\label{app:sequential}

\citet{li2026sequential} find that OPD followed by RL attains higher accuracy than the joint combinations they test. We run this schedule on Qwen3-1.7B-Base. REINFORCE++ continues for one epoch from the one-epoch checkpoint of full-vocabulary OPD, so the run processes the same 34,560 student responses as \RA{}+REINFORCE++, with every other setting as in Table~\ref{tab:setup}. Table~\ref{tab:sequential} compares it with its two stages and with \RA{}+REINFORCE++. The sequential run improves on both stages, reaching $28.39$ macro Avg@8 and $43.61$ Pass@8, and stays $0.86$ and $1.59$ points below \RA{}+REINFORCE++, which spends the same budget in one stage. It is ahead of \RA{} on AIME25 Avg@8 and behind on the other five benchmark scores, and every \RA{} seed exceeds every sequential seed in macro Avg@8, whereas the Pass@8 seeds overlap.

The joint schedule adds the full-vocabulary reverse KL of OPD, weighted by $\beta$, to the REINFORCE++ loss at every visited prefix, with every other setting as in Table~\ref{tab:setup}. At $\beta=0.05$, $0.2$, and $0.6$ it reaches $25.94$, $26.14$, and $25.82$ macro Avg@8 and $39.71$, $39.94$, and $39.53$ Pass@8. Table~\ref{tab:sequential} reports $\beta=0.2$, the weight selected on the validation split (Appendix~\ref{app:validation}): $0.47$ and $0.50$ points above verifier-only training, with every seed above the same verifier-only seed in macro Avg@8, but $3.11$ and $5.26$ points below \RA{}+REINFORCE++ and below the sequential schedule, in line with \citet{li2026sequential}.

\begin{table}[!htbp]
\caption{\textbf{Sequential and joint combinations of OPD and REINFORCE++ against \RA{}+REINFORCE++ at a matched response budget.} Qwen3-1.7B-Base; three-seed means in \%, with sample SD for the macro scores and the per-seed scores of the two combinations below; the reverse-KL weight is $\beta=0.2$. The other rows repeat Table~\ref{tab:main}; bold marks the best mean per column.}
\label{tab:sequential}
\centering
\begingroup
\small
\setlength{\tabcolsep}{2.5pt}
\renewcommand{\arraystretch}{1.0}
\begin{tabular}{@{}lcccccccc@{}}
\toprule
\tabletwoheadshade{tableheadcol}
\tabletwoheadlabel{Run} & \multicolumn{2}{c}{MATH-500} & \multicolumn{2}{c}{AIME24} & \multicolumn{2}{c}{AIME25} & \multicolumn{2}{c}{Macro}\\
\cmidrule(lr){2-3}\cmidrule(lr){4-5}\cmidrule(lr){6-7}\cmidrule(l){8-9}
 & Avg@8 & Pass@8 & Avg@8 & Pass@8 & Avg@8 & Pass@8 & Avg@8 & Pass@8\\
\midrule
OPD (two epochs) & 67.78 & 85.00 & 8.61 & 23.33 & 5.97 & 15.56 & 27.45\sd{0.25} & 41.30\sd{0.53}\\
REINFORCE++ & 65.76 & 83.87 & 7.78 & 20.00 & 3.47 & 14.44 & 25.67\sd{0.66} & 39.44\sd{1.65}\\
OPD, then REINFORCE++ & 68.65 & 85.27 & 9.72 & 26.67 & \textbf{6.81} & 18.89 & 28.39\sd{0.57} & 43.61\sd{1.88}\\
REINFORCE++ + reverse KL & 66.21 & 84.27 & 8.33 & 21.11 & 3.89 & 14.44 & 26.14\sd{0.97} & 39.94\sd{1.97}\\
\tablerowshade{reinforcecol}\RA{}+REINFORCE++ & \textbf{69.69} & \textbf{85.60} & \textbf{13.06} & \textbf{30.00} & 5.00 & \textbf{20.00} & \textbf{29.25}\sd{0.25} & \textbf{45.20}\sd{1.81}\\
\midrule
\multicolumn{9}{@{}l}{\textit{OPD, then REINFORCE++, by seed}}\\
\quad S1 & 67.93 & 84.80 & 8.33 & 23.33 & 7.08 & 16.67 & 27.78 & 41.60\\
\quad S2 & 69.25 & 86.00 & 11.25 & 30.00 & 5.00 & 20.00 & 28.50 & 45.33\\
\quad S3 & 68.78 & 85.00 & 9.58 & 26.67 & 8.33 & 20.00 & 28.90 & 43.89\\
\midrule
\multicolumn{9}{@{}l}{\textit{REINFORCE++ + reverse KL ($\beta{=}0.2$), by seed}}\\
\quad S1 & 66.50 & 84.60 & 9.17 & 23.33 & 4.17 & 16.67 & 26.61 & 41.53\\
\quad S2 & 65.08 & 83.20 & 7.08 & 16.67 & 2.92 & 13.33 & 25.03 & 37.73\\
\quad S3 & 67.05 & 85.00 & 8.75 & 23.33 & 4.58 & 13.33 & 26.79 & 40.56\\
\bottomrule
\end{tabular}
\endgroup

\end{table}

\subsection{Full-vocabulary and sampled OPD}
\label{app:sampled-opd}

The teacher-only OPD runs of Table~\ref{tab:main} minimize the reverse KL divergence over the full vocabulary at every visited prefix (Appendix~\ref{app:guidance}), which uses the same teacher information as \RA{}. \citet{lu2025opd} instead use the log ratio at the sampled token as a policy-gradient label. Table~\ref{tab:sampled-opd} runs this sampled variant on Qwen3-1.7B-Base with every other setting unchanged. In expectation over the sampled token the two give the same update at each prefix (Appendix~\ref{app:probability-path}), so they differ mainly in the variance of the estimate. The sampled variant reaches $24.80$ macro Avg@8 and $38.60$ Pass@8, $2.65$ and $2.70$ points below full-vocabulary OPD and $0.87$ and $0.84$ points below verifier-only REINFORCE++, and it is lower than full-vocabulary OPD on every benchmark score. A single sampled update can point against the expected one (Table~\ref{tab:app-sampled-case}), and in this setting the variance of the sampled estimate costs $2.65$ macro Avg@8 points.

\begin{table}[!htbp]
\caption{\textbf{Full-vocabulary and sampled OPD.} Qwen3-1.7B-Base; three-seed means in \%, with sample SD for the macro scores and the per-seed scores of the sampled run below. The other rows repeat Table~\ref{tab:main}.}
\label{tab:sampled-opd}
\centering
\begingroup
\small
\setlength{\tabcolsep}{3pt}
\renewcommand{\arraystretch}{1.0}
\begin{tabular}{@{}lcccccccc@{}}
\toprule
\tabletwoheadshade{tableheadcol}
\tabletwoheadlabel{Run} & \multicolumn{2}{c}{MATH-500} & \multicolumn{2}{c}{AIME24} & \multicolumn{2}{c}{AIME25} & \multicolumn{2}{c}{Macro}\\
\cmidrule(lr){2-3}\cmidrule(lr){4-5}\cmidrule(lr){6-7}\cmidrule(l){8-9}
 & Avg@8 & Pass@8 & Avg@8 & Pass@8 & Avg@8 & Pass@8 & Avg@8 & Pass@8\\
\midrule
OPD, full vocabulary & 67.78 & 85.00 & 8.61 & 23.33 & 5.97 & 15.56 & 27.45\sd{0.25} & 41.30\sd{0.53}\\
OPD, sampled token & 62.31 & 82.47 & 7.64 & 20.00 & 4.44 & 13.33 & 24.80\sd{0.79} & 38.60\sd{2.42}\\
REINFORCE++ & 65.76 & 83.87 & 7.78 & 20.00 & 3.47 & 14.44 & 25.67\sd{0.66} & 39.44\sd{1.65}\\
\tablerowshade{reinforcecol}\RA{}+REINFORCE++ & 69.69 & 85.60 & 13.06 & 30.00 & 5.00 & 20.00 & 29.25\sd{0.25} & 45.20\sd{1.81}\\
\midrule
\multicolumn{9}{@{}l}{\textit{OPD, sampled token, by seed}}\\
\quad S1 & 62.58 & 82.40 & 8.33 & 20.00 & 5.42 & 16.67 & 25.44 & 39.69\\
\quad S2 & 61.33 & 80.80 & 6.67 & 16.67 & 3.75 & 10.00 & 23.91 & 35.82\\
\quad S3 & 63.03 & 84.20 & 7.92 & 23.33 & 4.17 & 13.33 & 25.04 & 40.29\\
\bottomrule
\end{tabular}
\endgroup

\end{table}

\subsection{Pass@K up to 512}
\label{app:passk}

\begin{figure}[!htbp]
\centering
\includegraphics[width=0.55\linewidth]{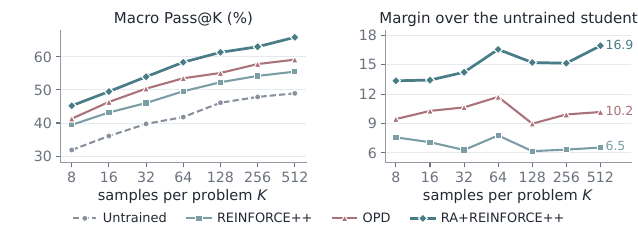}
\caption{\textbf{The margin over the untrained student does not close up to Pass@512.} Macro Pass@K of Table~\ref{tab:passk-full} and each trained run's margin over the untrained student, in points.}
\label{fig:passk}
\end{figure}

Table~\ref{tab:passk-full} and Figure~\ref{fig:passk} give the macro Pass@K of the 1.7B REINFORCE++ runs for $K\in\{8,16,\dots,512\}$ completions per problem, sampled under the evaluation settings of Table~\ref{tab:main}; the $K=8$ column reproduces Table~\ref{tab:main}. Pass@K is the share of problems solved at least once among the $K$ completions, averaged over the three benchmarks and then over three seeds. Every run gains with $K$ and the increments shrink, to $1.1$, $1.3$, and $1.3$ points at the last doubling for the untrained student, REINFORCE++, and OPD, against $2.9$ for \RA{}; the margins over the untrained student at $K=8$ and $K=512$ are $7.6$ and $6.6$ points for REINFORCE++, $9.5$ and $10.2$ for OPD, and $13.4$ and $16.9$ for \RA{}.

\begin{table}[!htbp]
\caption{\textbf{Macro Pass@K up to $K=512$, Qwen3-1.7B-Base with REINFORCE++.} Three-seed means in \%, with the increment from the previous $K$ in gray.}
\label{tab:passk-full}
\centering
\begingroup
\footnotesize
\setlength{\tabcolsep}{4pt}
\renewcommand{\arraystretch}{1.0}
\begin{tabular}{@{}lccccccc@{}}
\toprule
\tablerowshade{tableheadcol}
Method & $K{=}8$ & $K{=}16$ & $K{=}32$ & $K{=}64$ & $K{=}128$ & $K{=}256$ & $K{=}512$\\
\midrule
Qwen3-1.7B-Base (untrained) & 31.85 & 36.07 & 39.73 & 41.79 & 46.12 & 47.87 & 48.95\\
\quad\textcolor{black!60}{increment per doubling} & \textcolor{black!60}{--} & \textcolor{black!60}{+4.22} & \textcolor{black!60}{+3.66} & \textcolor{black!60}{+2.06} & \textcolor{black!60}{+4.33} & \textcolor{black!60}{+1.75} & \textcolor{black!60}{+1.08}\\
\addlinespace[2pt]
REINFORCE++ & 39.44 & 43.16 & 46.03 & 49.57 & 52.30 & 54.21 & 55.50\\
\quad\textcolor{black!60}{increment per doubling} & \textcolor{black!60}{--} & \textcolor{black!60}{+3.72} & \textcolor{black!60}{+2.87} & \textcolor{black!60}{+3.54} & \textcolor{black!60}{+2.73} & \textcolor{black!60}{+1.91} & \textcolor{black!60}{+1.29}\\
\addlinespace[2pt]
OPD (teacher only) & 41.30 & 46.35 & 50.37 & 53.50 & 55.10 & 57.79 & 59.12\\
\quad\textcolor{black!60}{increment per doubling} & \textcolor{black!60}{--} & \textcolor{black!60}{+5.05} & \textcolor{black!60}{+4.02} & \textcolor{black!60}{+3.13} & \textcolor{black!60}{+1.60} & \textcolor{black!60}{+2.69} & \textcolor{black!60}{+1.33}\\
\addlinespace[2pt]
\tablerowshade{reinforcecol}\RA{}+REINFORCE++ & 45.20 & 49.49 & 53.95 & 58.34 & 61.33 & 63.01 & 65.87\\
\quad\textcolor{black!60}{increment per doubling} & \textcolor{black!60}{--} & \textcolor{black!60}{+4.29} & \textcolor{black!60}{+4.46} & \textcolor{black!60}{+4.39} & \textcolor{black!60}{+2.99} & \textcolor{black!60}{+1.68} & \textcolor{black!60}{+2.86}\\
\bottomrule
\end{tabular}
\endgroup

\end{table}

\subsection{Untrained students and the teacher}
\label{app:reference-points}

Table~\ref{tab:reference-points} evaluates the two untrained students and the teacher under the evaluation protocol of Appendix~\ref{app:protocol}: the same prompt template, eight completions per problem at temperature $0.8$, and the 12,000-token cap, with the teacher in non-thinking mode. Here $\pm$ is the sample SD across three evaluation runs with different sampling seeds, not across training seeds. Verifier-only training lifts macro Avg@8 by 12 to 15 points over the untrained students; \RA{} adds 1.7 to 3.6 points and \CoRA{} 2.8 to 5.0 over the corresponding estimator. The most accurate 4B run, \CoRA{}+GRPO, ends 5.4 macro Avg@8 points below the teacher, 1.6 on MATH-500 and about 7 on each AIME set, and teacher-only OPD reaches 62\% and 79\% of the teacher's macro Avg@8 at the two sizes. Every trained run exceeds the untrained student's Pass@8 on every benchmark, and Appendix~\ref{app:passk} follows the 1.7B REINFORCE++ runs to Pass@512.

\begin{table}[!htbp]
\caption{\textbf{Untrained students and the teacher under the evaluation protocol.} Qwen3-1.7B-Base and Qwen3-4B-Base before any training, and the Qwen3-8B teacher in non-thinking mode. Scores in \%; $\pm$ is the sample SD across three evaluation runs.}
\label{tab:reference-points}
\centering
\begingroup
\footnotesize
\setlength{\tabcolsep}{1.5pt}
\renewcommand{\arraystretch}{1.0}
\begin{tabular}{@{}lcccccccc@{}}
\toprule
\tabletwoheadshade{tableheadcol}
\tabletwoheadlabel{Model} & \multicolumn{2}{c}{MATH-500} & \multicolumn{2}{c}{AIME24} & \multicolumn{2}{c}{AIME25} & \multicolumn{2}{c}{Macro}\\
\cmidrule(lr){2-3}\cmidrule(lr){4-5}\cmidrule(lr){6-7}\cmidrule(l){8-9}
 & Avg@8 & Pass@8 & Avg@8 & Pass@8 & Avg@8 & Pass@8 & Avg@8 & Pass@8\\
\midrule
1.7B-Base & 36.83\sd{0.17} & 76.67\sd{1.10} & 2.08\sd{1.10} & 10.00\sd{3.33} & 1.53\sd{0.64} & 8.89\sd{1.92} & 13.48\sd{0.56} & 31.85\sd{1.00}\\
4B-Base & 51.79\sd{0.30} & 89.60\sd{0.40} & 7.22\sd{0.64} & 22.22\sd{1.92} & 4.72\sd{0.24} & 20.00\sd{3.33} & 21.25\sd{0.13} & 43.94\sd{1.17}\\
8B teacher & 84.23\sd{0.33} & 95.53\sd{0.70} & 26.25\sd{1.25} & 43.33\sd{0.00} & 21.39\sd{2.29} & 38.89\sd{6.94} & 43.95\sd{1.23} & 59.25\sd{2.47}\\
\bottomrule
\end{tabular}
\endgroup
\end{table}

\subsection{The adapted teacher under the evaluation protocol}
\label{app:adapted-teacher}

Table~\ref{tab:adapted-teacher} evaluates the teacher LoRA at the end of each \CoRA{} run, applied to the frozen Qwen3-8B teacher in non-thinking mode, under the protocol of Table~\ref{tab:reference-points}; each adapted row averages the three adapters of the run's three training seeds, evaluated once each, and $\pm$ is the sample SD across them. Every adapted teacher scores below the frozen teacher, by $0.8$--$2.3$ macro Avg@8 and $0.7$--$3.5$ Pass@8 points, and the drop is larger for the adapters trained with the 1.7B student than for those trained with the 4B student. The adapted teacher is a better guide for the student that trained it (Table~\ref{tab:main}) and a worse solver on its own, in line with the distinction between a strong solver and a good teacher drawn in Section~1. Table~\ref{tab:adapted-teacher-seeds} lists the macro scores of each adapter.

\begin{table}[!htbp]
\caption{\textbf{The adapted teacher is a worse solver than the frozen teacher.} Qwen3-8B in non-thinking mode with the \CoRA{} adapter of each setting, under the evaluation protocol; scores in \%. The frozen row repeats Table~\ref{tab:reference-points}, where $\pm$ is the SD across three evaluation runs; in the adapted rows $\pm$ is the sample SD across the three seeds' adapters.}
\label{tab:adapted-teacher}
\centering
\begingroup
\footnotesize
\setlength{\tabcolsep}{1.8pt}
\renewcommand{\arraystretch}{1.0}
\begin{tabular}{@{}lcccccccc@{}}
\toprule
\tabletwoheadshade{tableheadcol}
\tabletwoheadlabel{Teacher} & \multicolumn{2}{c}{MATH-500} & \multicolumn{2}{c}{AIME24} & \multicolumn{2}{c}{AIME25} & \multicolumn{2}{c}{Macro}\\
\cmidrule(lr){2-3}\cmidrule(lr){4-5}\cmidrule(lr){6-7}\cmidrule(l){8-9}
 & Avg@8 & Pass@8 & Avg@8 & Pass@8 & Avg@8 & Pass@8 & Avg@8 & Pass@8\\
\midrule
Frozen $q_0$ & 84.23 & 95.53 & 26.25 & 43.33 & 21.39 & 38.89 & 43.95\sd{1.23} & 59.25\sd{2.47}\\
\midrule
Adapter, 1.7B + REINFORCE++ & 82.75 & 94.27 & 24.03 & 41.11 & 19.31 & 35.56 & 42.03\sd{0.77} & 56.98\sd{2.39}\\
Adapter, 1.7B + GRPO & 82.56 & 93.93 & 23.61 & 38.89 & 18.75 & 34.44 & 41.64\sd{0.65} & 55.76\sd{2.08}\\
Adapter, 4B + REINFORCE++ & 82.97 & 94.53 & 26.39 & 44.44 & 20.14 & 36.67 & 43.16\sd{0.53} & 58.55\sd{2.46}\\
Adapter, 4B + GRPO & 82.90 & 94.40 & 24.44 & 40.00 & 21.53 & 40.00 & 42.96\sd{0.80} & 58.13\sd{2.32}\\
\bottomrule
\end{tabular}
\endgroup

\end{table}

\begin{table}[!htbp]
\caption{\textbf{Per-seed macro results of the adapted teachers.} Macro Avg@8 / Pass@8 in \% of the adapter from each \CoRA{} training seed.}
\label{tab:adapted-teacher-seeds}
\centering
\begingroup
\small
\setlength{\tabcolsep}{6pt}
\renewcommand{\arraystretch}{1.0}
\begin{tabular}{@{}lccc@{}}
\toprule
\tablerowshade{tableheadcol}
Adapter trained with & S1 & S2 & S3\\
\midrule
Qwen3-1.7B + REINFORCE++ & 42.29 / 57.09 & 41.16 / 54.53 & 42.63 / 59.31\\
Qwen3-1.7B + GRPO & 42.01 / 56.89 & 40.89 / 53.36 & 42.02 / 57.02\\
Qwen3-4B + REINFORCE++ & 43.44 / 59.38 & 42.55 / 55.78 & 43.50 / 60.49\\
Qwen3-4B + GRPO & 43.08 / 58.20 & 42.10 / 55.78 & 43.69 / 60.42\\
\bottomrule
\end{tabular}
\endgroup

\end{table}


\begin{thebibliography}{40}
\providecommand{\natexlab}[1]{#1}
\providecommand{\url}[1]{\texttt{#1}}
\expandafter\ifx\csname urlstyle\endcsname\relax
  \providecommand{\doi}[1]{doi: #1}\else
  \providecommand{\doi}{doi: \begingroup \urlstyle{rm}\Url}\fi

\bibitem[Agarwal et~al.(2024)Agarwal, Vieillard, Zhou, Stanczyk, Ramos, Geist,
  and Bachem]{gkd}
Rishabh Agarwal, Nino Vieillard, Yongchao Zhou, Piotr Stanczyk, Sabela Ramos,
  Matthieu Geist, and Olivier Bachem.
\newblock On-policy distillation of language models: Learning from
  self-generated mistakes.
\newblock In \emph{International Conference on Learning Representations}, 2024.
\newblock URL \url{https://openreview.net/forum?id=3zKtaqxLhW}.

\bibitem[Ahmadian et~al.(2024)Ahmadian, Cremer, Gall{\'e}, Fadaee, Kreutzer,
  Pietquin, {\"U}st{\"u}n, and Hooker]{ahmadian2024rloo}
Arash Ahmadian, Chris Cremer, Matthias Gall{\'e}, Marzieh Fadaee, Julia
  Kreutzer, Olivier Pietquin, Ahmet {\"U}st{\"u}n, and Sara Hooker.
\newblock Back to basics: Revisiting {REINFORCE}-style optimization for
  learning from human feedback in {LLM}s.
\newblock In \emph{Proceedings of the 62nd Annual Meeting of the Association for
  Computational Linguistics (Volume 1: Long Papers)}, pp.\  12248--12267, 2024.

\bibitem[Basu et~al.(1998)Basu, Harris, Hjort, and Jones]{basu1998robust}
Ayanendranath Basu, Ian~R. Harris, Nils~L. Hjort, and M.~C. Jones.
\newblock Robust and efficient estimation by minimising a density power
  divergence.
\newblock \emph{Biometrika}, 85\penalty0 (3):\penalty0 549--559, 1998.

\bibitem[Cui et~al.(2025)Cui, Yuan, Wang, Wang, Zhang, Chen, Li, He, Fan, Yu,
  Xu, et~al.]{cui2025prime}
Ganqu Cui, Lifan Yuan, Zefan Wang, Hanbin Wang, Yuchen Zhang, Jiacheng Chen,
  Wendi Li, Bingxiang He, Yuchen Fan, Tianyu Yu, Qixin Xu, et~al.
\newblock Process reinforcement through implicit rewards.
\newblock \emph{arXiv preprint arXiv:2502.01456}, 2025.

\bibitem[{DeepSeek-AI} et~al.(2025)]{deepseekr1}
{DeepSeek-AI} et~al.
\newblock {DeepSeek-R1}: Incentivizing reasoning capability in {LLM}s via
  reinforcement learning.
\newblock \emph{arXiv preprint arXiv:2501.12948}, 2025.
\newblock URL \url{https://arxiv.org/abs/2501.12948}.

\bibitem[Duchi et~al.(2008)Duchi, Shalev-Shwartz, Singer, and Chandra]{duchi2008}
John Duchi, Shai Shalev-Shwartz, Yoram Singer, and Tushar Chandra.
\newblock Efficient projections onto the $\ell_1$-ball for learning in high
  dimensions.
\newblock In \emph{Proceedings of the 25th International Conference on Machine
  Learning}, pp.\  272--279, 2008.

\bibitem[Foerster et~al.(2018)Foerster, Farquhar, Afouras, Nardelli, and
  Whiteson]{foerster2018coma}
Jakob Foerster, Gregory Farquhar, Triantafyllos Afouras, Nantas Nardelli, and
  Shimon Whiteson.
\newblock Counterfactual multi-agent policy gradients.
\newblock In \emph{Proceedings of the AAAI Conference on Artificial
  Intelligence}, volume~32, 2018.

\bibitem[Gao et~al.(2025)Gao, Wan, Guo, Quan, and Wang]{ICLR2025_2f891d02}
Shiping Gao, Fanqi Wan, Jiajian Guo, Xiaojun Quan, and Qifan Wang.
\newblock Advantage-guided distillation for preference alignment in small
  language models.
\newblock In \emph{International Conference on Learning Representations}, 2025.
\newblock URL \url{https://openreview.net/forum?id=xsx3Fpo3UD}.

\bibitem[Greensmith et~al.(2004)Greensmith, Bartlett, and
  Baxter]{greensmith2004variance}
Evan Greensmith, Peter~L. Bartlett, and Jonathan Baxter.
\newblock Variance reduction techniques for gradient estimates in reinforcement
  learning.
\newblock \emph{Journal of Machine Learning Research}, 5:\penalty0 1471--1530,
  2004.

\bibitem[Gu et~al.(2024)Gu, Dong, Wei, and Huang]{minillm}
Yuxian Gu, Li~Dong, Furu Wei, and Minlie Huang.
\newblock {MiniLLM}: Knowledge distillation of large language models.
\newblock In \emph{International Conference on Learning Representations}, 2024.
\newblock URL \url{https://openreview.net/forum?id=5h0qf7IBZZ}.

\bibitem[Hampel(1974)]{hampel1974influence}
Frank~R. Hampel.
\newblock The influence curve and its role in robust estimation.
\newblock \emph{Journal of the American Statistical Association}, 69\penalty0
  (346):\penalty0 383--393, 1974.

\bibitem[Hu et~al.(2022)Hu, Shen, Wallis, Allen-Zhu, Li, Wang, Wang, and
  Chen]{hu2022lora}
Edward~J. Hu, Yelong Shen, Phillip Wallis, Zeyuan Allen-Zhu, Yuanzhi Li, Shean
  Wang, Lu Wang, and Weizhu Chen.
\newblock {LoRA}: Low-rank adaptation of large language models.
\newblock In \emph{International Conference on Learning Representations}, 2022.
\newblock URL \url{https://openreview.net/forum?id=nZeVKeeFYf9}.

\bibitem[Hu(2025)]{hu2025reinforcepp}
Jian Hu.
\newblock {REINFORCE++}: A simple and efficient approach for aligning large
  language models, 2025.
\newblock URL \url{https://arxiv.org/abs/2501.03262v1}.
\newblock Version 1, 4 January 2025.

\bibitem[Huang et~al.(2025)Huang, Block, Foster, Rohatgi, Zhang, Simchowitz,
  Ash, and Krishnamurthy]{huang2024sharpening}
Audrey Huang, Adam Block, Dylan~J. Foster, Dhruv Rohatgi, Cyril Zhang, Max
  Simchowitz, Jordan~T. Ash, and Akshay Krishnamurthy.
\newblock Self-improvement in language models: The sharpening mechanism.
\newblock In \emph{International Conference on Learning Representations}, 2025.
\newblock URL \url{https://openreview.net/forum?id=yNN1xYMgxG}.

\bibitem[Kazemnejad et~al.(2025)Kazemnejad, Aghajohari, Portelance, Sordoni,
  Reddy, Courville, and Le~Roux]{kazemnejad2024vineppo}
Amirhossein Kazemnejad, Milad Aghajohari, Eva Portelance, Alessandro Sordoni,
  Siva Reddy, Aaron Courville, and Nicolas Le~Roux.
\newblock {VinePPO}: Refining credit assignment in {RL} training of {LLMs}.
\newblock In \emph{Proceedings of the 42nd International Conference on Machine
  Learning}, volume~267, pp.\  29557--29590, 2025.

\bibitem[Li et~al.(2026{\natexlab{a}})Li, Chen, Yang, Nie, Zhao, and Ye]{li2026sequential}
Boyan Li, Bingsen Chen, Chenghao Yang, Ping Nie, Chen Zhao, and Xi Ye.
\newblock Sequential beats joint: On the interplay between on-policy
  distillation and {RLVR}, 2026{\natexlab{a}}.
\newblock URL \url{https://arxiv.org/abs/2609.04108v2}.
\newblock Version 2, 4 September 2026.

\bibitem[Li et~al.(2026{\natexlab{b}})Li, Nijkamp, Yavuz, and Joty]{li2026vpd}
Yang Li, Erik Nijkamp, Semih Yavuz, and Shafiq Rayhan Joty.
\newblock Learning from language feedback via variational policy distillation,
  2026{\natexlab{b}}.
\newblock URL \url{https://arxiv.org/abs/2605.15113v1}.
\newblock Version 1, 14 May 2026.

\bibitem[Li et~al.(2025)Li, Yue, Xu, Jiang, Niu, Lin, Ramasubramanian, and
  Poovendran]{li2025smallmodels}
Yuetai Li, Xiang Yue, Zhangchen Xu, Fengqing Jiang, Luyao Niu, Bill~Yuchen Lin,
  Bhaskar Ramasubramanian, and Radha Poovendran.
\newblock Small models struggle to learn from strong reasoners.
\newblock In \emph{Findings of the Association for Computational Linguistics: ACL
  2025}, pp.\  25366--25394, 2025.

\bibitem[Li et~al.(2024)Li, Xu, Zhang, Lin, Yu, Sun, and Luo]{li2023remax}
Ziniu Li, Tian Xu, Yushun Zhang, Zhihang Lin, Yang Yu, Ruoyu Sun, and Zhi-Quan
  Luo.
\newblock {ReMax}: A simple, effective, and efficient reinforcement learning
  method for aligning large language models.
\newblock In \emph{Proceedings of the 41st International Conference on Machine
  Learning}, volume~235, pp.\  29128--29163, 2024.

\bibitem[Lightman et~al.(2024)Lightman, Kosaraju, Burda, Edwards, Baker, Lee,
  Leike, Schulman, Sutskever, and Cobbe]{lightman2023lets}
Hunter Lightman, Vineet Kosaraju, Yura Burda, Harri Edwards, Bowen Baker, Teddy
  Lee, Jan Leike, John Schulman, Ilya Sutskever, and Karl Cobbe.
\newblock Let's verify step by step.
\newblock In \emph{International Conference on Learning Representations}, 2024.
\newblock URL \url{https://openreview.net/forum?id=v8L0pN6EOi}.

\bibitem[Lin et~al.(2026)Lin, Zhao, Jiang, Rao, Li, Wang, He, and Huang]{opdvr2026}
Wenze Lin, Jiale Zhao, Xitai Jiang, Songde Rao, Yining Li, Shenzhi Wang,
  Bingxiang He, and Gao Huang.
\newblock On-policy distillation with verifiable reward, 2026.
\newblock URL \url{https://arxiv.org/abs/2608.24696v1}.
\newblock Version 1, 25 August 2026.

\bibitem[Lu \& {Thinking Machines Lab}(2025)Lu and {Thinking Machines
  Lab}]{lu2025opd}
Kevin Lu and {Thinking Machines Lab}.
\newblock On-policy distillation.
\newblock Thinking Machines Lab: Connectionism,
  \url{https://thinkingmachines.ai/blog/on-policy-distillation/}, 2025.

\bibitem[Martins \& Astudillo(2016)Martins and Astudillo]{martins2016sparsemax}
Andr{\'e} F.~T. Martins and Ram{\'o}n Fernandez Astudillo.
\newblock From softmax to sparsemax: A sparse model of attention and
  multi-label classification.
\newblock In \emph{Proceedings of the 33rd International Conference on Machine
  Learning}, volume~48, pp.\  1614--1623, 2016.

\bibitem[Michelot(1986)]{michelot1986}
Christian Michelot.
\newblock A finite algorithm for finding the projection of a point onto the
  canonical simplex of $\mathbb{R}^n$.
\newblock \emph{Journal of Optimization Theory and Applications}, 50\penalty0
  (1):\penalty0 195--200, 1986.

\bibitem[Peng et~al.(2019)Peng, Kumar, Zhang, and Levine]{peng2019awr}
Xue~Bin Peng, Aviral Kumar, Grace Zhang, and Sergey Levine.
\newblock Advantage-weighted regression: Simple and scalable off-policy
  reinforcement learning.
\newblock \emph{arXiv preprint arXiv:1910.00177}, 2019.

\bibitem[Rafailov et~al.(2023)Rafailov, Sharma, Mitchell, Manning, Ermon, and
  Finn]{rafailov2023dpo}
Rafael Rafailov, Archit Sharma, Eric Mitchell, Christopher~D. Manning, Stefano
  Ermon, and Chelsea Finn.
\newblock Direct preference optimization: Your language model is secretly a
  reward model.
\newblock In \emph{Advances in Neural Information Processing Systems},
  volume~36, pp.\  53728--53741. Curran Associates, Inc., 2023.
\newblock \doi{10.52202/075280-2338}.
\newblock URL
  \url{https://proceedings.neurips.cc/paper_files/paper/2023/file/a85b405ed65c6477a4fe8302b5e06ce7-Paper-Conference.pdf}.

\bibitem[Rafailov et~al.(2024)Rafailov, Hejna, Park, and
  Finn]{rafailov2024rtoq}
Rafael Rafailov, Joey Hejna, Ryan Park, and Chelsea Finn.
\newblock From $r$ to {$Q^*$}: Your language model is secretly a {Q}-function.
\newblock In \emph{First Conference on Language Modeling}, 2024.
\newblock URL \url{https://arxiv.org/abs/2404.12358}.

\bibitem[Schulman et~al.(2017)Schulman, Wolski, Dhariwal, Radford, and
  Klimov]{schulman2017ppo}
John Schulman, Filip Wolski, Prafulla Dhariwal, Alec Radford, and Oleg Klimov.
\newblock Proximal policy optimization algorithms.
\newblock \emph{arXiv preprint arXiv:1707.06347}, 2017.
\newblock URL \url{https://arxiv.org/abs/1707.06347}.

\bibitem[Shao et~al.(2024)Shao, Wang, Zhu, Xu, Song, Bi, Zhang, Zhang, Li, Wu,
  and Guo]{deepseekmath}
Zhihong Shao, Peiyi Wang, Qihao Zhu, Runxin Xu, Junxiao Song, Xiao Bi, Haowei
  Zhang, Mingchuan Zhang, Y.~K. Li, Y.~Wu, and Daya Guo.
\newblock {DeepSeekMath}: Pushing the limits of mathematical reasoning in open
  language models.
\newblock \emph{arXiv preprint arXiv:2402.03300}, 2024.
\newblock URL \url{https://arxiv.org/abs/2402.03300}.

\bibitem[Sutton et~al.(2000)Sutton, McAllester, Singh, and
  Mansour]{sutton2000pg}
Richard~S. Sutton, David McAllester, Satinder Singh, and Yishay Mansour.
\newblock Policy gradient methods for reinforcement learning with function
  approximation.
\newblock In \emph{Advances in Neural Information Processing Systems},
  volume~12, pp.\  1057--1063, 2000.

\bibitem[Tiapkin et~al.(2025)Tiapkin, Calandriello, Ferret, Perrin, Vieillard,
  Ram{\'e}, and Blondel]{tiapkin2025teacherhacking}
Daniil Tiapkin, Daniele Calandriello, Johan Ferret, Sarah Perrin, Nino
  Vieillard, Alexandre Ram{\'e}, and Mathieu Blondel.
\newblock On teacher hacking in language model distillation.
\newblock In \emph{Proceedings of the 42nd International Conference on Machine
  Learning}, volume~267, pp.\  59552--59569, 2025.

\bibitem[Wang et~al.(2025)Wang, Yu, Gao, Zheng, Liu, Lu, Dang, Chen, Yang,
  Zhang, Liu, Yang, Zhao, Yue, Song, Yu, Huang, and Lin]{wang2025entropy}
Shenzhi Wang, Le Yu, Chang Gao, Chujie Zheng, Shixuan Liu, Rui Lu, Kai Dang,
  Xionghui Chen, Jianxin Yang, Zhenru Zhang, Yuqiong Liu, An Yang, Andrew Zhao,
  Yang Yue, Shiji Song, Bowen Yu, Gao Huang, and Junyang Lin.
\newblock Beyond the 80/20 rule: High-entropy minority tokens drive effective
  reinforcement learning for {LLM} reasoning.
\newblock \emph{arXiv preprint arXiv:2506.01939}, 2025.

\bibitem[Williams(1992)]{williams1992simple}
Ronald~J. Williams.
\newblock Simple statistical gradient-following algorithms for connectionist
  reinforcement learning.
\newblock \emph{Machine Learning}, 8\penalty0 (3--4):\penalty0 229--256, 1992.
\newblock \doi{10.1007/BF00992696}.

\bibitem[Yan et~al.(2025)Yan, Li, Hu, Wang, Cui, Qu, Cheng, and
  Zhang]{yan2025luffy}
Jianhao Yan, Yafu Li, Zican Hu, Zhi Wang, Ganqu Cui, Xiaoye Qu, Yu~Cheng, and
  Yue Zhang.
\newblock Learning to reason under off-policy guidance.
\newblock In \emph{Advances in Neural Information Processing Systems},
  volume~38, 2025.

\bibitem[Yang et~al.(2025)Yang, Li, Yang, Zhang, Hui, Zheng, Yu, Gao, Huang,
  Lv, et~al.]{qwen3}
An~Yang, Anfeng Li, Baosong Yang, Beichen Zhang, Binyuan Hui, Bo~Zheng, Bowen
  Yu, Chang Gao, Chengen Huang, Chenxu Lv, et~al.
\newblock {Qwen3} technical report.
\newblock \emph{arXiv preprint arXiv:2505.09388}, 2025.
\newblock URL \url{https://arxiv.org/abs/2505.09388}.

\bibitem[Yu et~al.(2025)Yu, Zhang, Zhu, Yuan, Zuo, Yue, Dai, Fan, Liu, Liu,
  et~al.]{dapo}
Qiying Yu, Zheng Zhang, Ruofei Zhu, Yufeng Yuan, Xiaochen Zuo, Yu~Yue, Weinan
  Dai, Tiantian Fan, Gaohong Liu, Lingjun Liu, et~al.
\newblock {DAPO}: An open-source {LLM} reinforcement learning system at scale.
\newblock In \emph{Advances in Neural Information Processing Systems},
  volume~38, 2025.
\newblock URL \url{https://arxiv.org/abs/2503.14476}.

\bibitem[Yuan et~al.(2025)Yuan, Li, Chen, Cui, Ding, Zhang, Zhou, Liu, and
  Peng]{yuan2024implicitprm}
Lifan Yuan, Wendi Li, Huayu Chen, Ganqu Cui, Ning Ding, Kaiyan Zhang, Bowen
  Zhou, Zhiyuan Liu, and Hao Peng.
\newblock Free process rewards without process labels.
\newblock In \emph{Proceedings of the 42nd International Conference on Machine
  Learning}, volume~267, pp.\  73511--73525, 2025.

\bibitem[Yue et~al.(2025)Yue, Chen, Lu, Zhao, Wang, Yue, Song, and
  Huang]{yue2025rlvr}
Yang Yue, Zhiqi Chen, Rui Lu, Andrew Zhao, Zhaokai Wang, Yang Yue, Shiji Song,
  and Gao Huang.
\newblock Does reinforcement learning really incentivize reasoning capacity in
  {LLM}s beyond the base model?
\newblock In \emph{Advances in Neural Information Processing Systems},
  volume~38, 2025.

\bibitem[Zhao et~al.(2026)Zhao, Chen, Lin, Winata, Yao, and Tang]{opdplus2026}
Hanyang Zhao, Haoxian Chen, Han Lin, Genta~Indra Winata, David Yao, and Wenpin
  Tang.
\newblock {OPD+}: Rethinking the advantage design for on-policy distillation,
  2026.
\newblock URL \url{https://arxiv.org/abs/2606.01039v1}.
\newblock Version 1, 31 May 2026.

\end{thebibliography}
\end{document}